\documentclass[12pt]{amsart} 

\usepackage{graphicx,amssymb,colordvi,textcomp,latexsym} 

\usepackage{color}
\usepackage{soul}

\newcommand{\rend}{\hfill $\maltese$}

\newcommand{\wbxi}{{\widetilde{\boldsymbol{\xi}}}}
\newcommand{\wbxip}{{\widetilde{\boldsymbol{\xi'_0}}}}
\newcommand{\bxi}{{\boldsymbol{\xi}}}
\newcommand{\wxi}{{\widetilde{\xi}}}
\newcommand{\trho}{{\bar{\rho}}}

\newcommand{\ip}[2]{{\langle #1 , #2\rangle}}

\newcommand{\grad}[1]{{\text{\rm grad}({#1})}}
\newcommand{\is}[1]{{\mathbf{#1}}}
\newcommand{\isk}[1]{{\mathbf{\mathfrak{#1}}}}

\newcommand{\dd}{\mbox{$\;|\;$}}

\newcommand{\oxi}{{\overline{\xi}}}

\newcommand{\Refb}[1]{(\ref{#1})}

\newcommand{\eps}{\mbox{$\varepsilon$}}
\newcommand{\real}{{\mathbb{R}}}

\newcommand{\intg}{{\mathbb{Z}}}
\newcommand{\xx}{{\mathbf{x}}}

\newcommand{\xxd}{\dot{\is{x}}}

\newcommand{\cal}[1]{{\mathcal{#1}}}

\newcommand{\On}[1]{{\text{\rm O}({#1})}}
\newcommand{\defoo}{\stackrel{\mathrm{def}}{=}}

\newcommand{\yy}{{\is{y}}}
\newcommand{\vv}{{\is{v}}}

\newcommand{\VV}{{\is{V}}}
\newcommand{\ww}{{\is{w}}}
\newcommand{\WW}{{\is{W}}}

\newcommand{\GG}{{\Gamma}}
\newcommand{\ee}{{\is{e}}}

\newcommand{\pint}{\mbox{$\mathbb{N}$}}

\newcommand{\arr}{{\rightarrow}}

\newcommand{\WXi}{{\widehat{\Xi}^\Sigma}}

\newcommand{\loss}{{\mathcal{L}}}

\newtheorem{lemma}{Lemma}[section]

\newtheorem{prop}[lemma]{Proposition}
\newtheorem{thm}[lemma]{Theorem}
\newtheorem{cor}[lemma]{Corollary}
\theoremstyle{definition}
\newtheorem{Def}[lemma]{Definition}
\newtheorem{exam}[lemma]{Example}
\newtheorem{exams}[lemma]{Examples}

\theoremstyle{remark}
\newtheorem{rem}[lemma]{Remark}
\newtheorem{rems}[lemma]{Remarks}

\newcommand{\vareps}{\varepsilon}

\title[Symmetry \& a model ReLU network]{Symmetry \& critical points for a model shallow neural network}
\author{Yossi Arjevani and Michael Field}
\address{Yossi Arjevani, Center for Data Science, NYU, New York, NY, 10011}
\email{yossi.arjevani@gmail.com}
\address{Michael Field, Department of Mechanical Engineering, UCSB, 
Santa Barbara, CA 93106}
\email{mikefield@gmail.com}

\date{\today}
\begin{document}
\begin{abstract}
Using methods based on the analysis of real analytic functions, symmetry and equivariant
bifurcation theory, we obtain sharp results on families of critical points of spurious minima that 
occur in optimization problems associated with fitting two-layer ReLU networks with $k$ hidden neurons. 
The main mathematical result proved is to obtain power series representations of families of critical points 
of spurious minima in terms of $1/\sqrt{k}$ (coefficients independent of $k$).  
We also give a path based formulation that naturally connects the critical points with critical points of
an associated linear, but highly singular, optimization problem. These critical points 
closely approximate the critical points in the original problem.

The mathematical  theory is used to derive results on the original problem in neural nets. For example,
precise estimates for several quantities that show that not all spurious minima are alike.
In particular, we show that while the loss function at certain types of spurious
minima decays to zero like $k^{-1}$, in other cases the loss converges to a strictly positive constant.
\end{abstract}

\maketitle

\section{Introduction}

The great empirical success of artificial neural networks over the
past decade has challenged the foundations of our understanding of
statistical learning processes. From the optimization point of view, a
particularly puzzling, and often observed phenomenon, 
is that---although highly non-convex---optimization landscapes
induced by \emph{natural} distributions allow simple gradient-based methods,
such as stochastic gradient descent (SGD), to find good minima efficiently~\cite{Dauphin2014,IG2014,LeCun1990}.

In an effort to find more tractable ways of investigating this phenomenon, a
large body of recent works has considered $2$-layer networks which differ
by their choice of, for example, activation function, underlying data
distribution, the number and width of the hidden layers with respect to the number of samples,
and numerical solvers~\cite{BrutzkusGloberson2018,Li2018,Soltanolkotabi2018,Xie2017,Zhong2017,Panigrahy2018,Du2018,Janzamin2015}. Much of this work has focused
on Gaussian inputs~\cite{Zhang2017,Du2017,Feizi2017,Li2017,Tian2017,BrutzkusGloberson2017,Ge2018}.
Recently, Safran \& Shamir~\cite{SafranShamir2018} considered a well-studied family of $2$-layer ReLU networks (details appear later in the introduction) and
showed that the expected squared loss with respect to a target network with identity weight matrix,
possessed a large number of spurious local minima which can cause gradient-based methods to fail. 

In this work we present a detailed analysis of the family of critical points determining
the spurious minima described by Safran \& Shamir~\emph{op.~cit} and two new families of spurious
minima that were not detected by SGD in their work. 
The families all exhibit symmetry related to that of the target model (see~\cite{ArjevaniField2019c} and below);
elsewhere~\cite{ArjevaniField2020b}, we show that the families define 
spurious minima if $k$, the number of neurons, is at least $6$ or $7$. One of the families (type A)
has the same symmetry as the solution giving the global minimum; the two other families have less symmetry.
Our emphasis is on understanding, in some depth, the structure of this deceptively simple model and so
we do not discuss issues associated with deep neural nets (see the survey 
article~\cite{Schmidhuber2015} and text~\cite{IG2017}). Thus, 
we formalize the symmetry properties of a class of student-teacher shallow ReLU neural networks and show 
their use in studying several families of
critical points. More specifically, 
\begin{itemize}
\item[---] We show that the optimization landscape has rich symmetry structure coming from a
natural action of the group $\Gamma = S_k\times S_d$ on the parameter space ($k \times d$-matrices). 
Our approach for addressing the intricate structure of the critical points uses this symmetry in essential ways, notably
by making use of the fixed point spaces of isotropy groups of critical points.

\item[---] We present the relevant facts about $\Gamma$-spaces and $\Gamma$-invariance needed for our approach. 

\item[---] We show that two families of critical points found by SGD in data sets of~\cite{SafranShamir2018} 
exhibit maximal isotropy reminiscent of many situations in Physics (spontaneous symmetry breaking) and
Mathematics (bifurcation theory) and confirm the empirical observation~\cite{ArjevaniField2019c} that SGD detects
highly symmetric minima.  

\item[---] The assumption of symmetry allows us to reduce much of the analysis to low dimensional fixed point spaces. 
Focusing on classes of critical points  with maximal isotropy, we develop novel approaches for constructing solutions and
obtain series in $1/\sqrt{k}$ for the critical points  when $d \ge k$. These series allow us to prove, for example,
that the spurious minima found by Safran \& Shamir~\cite{SafranShamir2018}
decay like $(\frac{1}{2} -\frac{2}{\pi^2})k^{-1}$.
Part of our analysis shows that we can find solutions of a simpler problem in fewer variables (what we call the \emph{consistency equations})
that give (quantifiably) extremely good approximations to the critical points defining spurious minima. 
We also describe three other families of spurious minima, with different symmetry patterns. Only one of these families
appears in the data sets of~\cite{SafranShamir2018}. 

\item[---] Overall, our approach introduces new ideas from symmetry breaking, bifurcation, and algebraic geometry, notably Artin's
implicit function theorem, and makes a surprising use of the leaky ReLU activation
function. The notion of real analyticity plays a central role.
Many intriguing and challenging mathematical problems remain, notably that of achieving a
more complete understanding of the singularity set of the objective function, which is related to the isotropy
structure of the $\Gamma$-action, and mechanisms for the creation 
of spurious minima~\cite{ArjevaniField2020c}.
\end{itemize}

After a brief review of neural nets, the introduction continues with a description of the model studied and
the basic structures required from neural nets, in particular the \emph{Rectified Linear Unit}
(ReLU) activation function. We conclude with a description of the main results and 
outline of the structure of the paper.

\subsection{Neural nets, neurons and activation functions}
\begin{figure}[h]
\centering
\includegraphics[width=\textwidth]{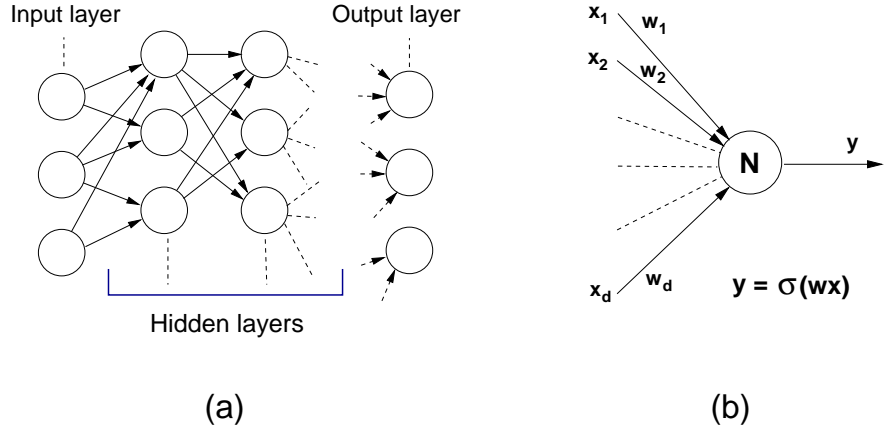}
\caption{(a) A feedforward neural net showing different layers. (b) Activation function for a neuron.}
\label{fig: ann}
\end{figure}

A typical neural net comprises an input layer, a number of hidden layers and an output layer. Each layer is
comprised of ``neurons'' which receive inputs from previous layers via weighted connections.  See Figure~\ref{fig: ann}(a). 

\begin{figure}[h]
\centering
\includegraphics[width=\textwidth]{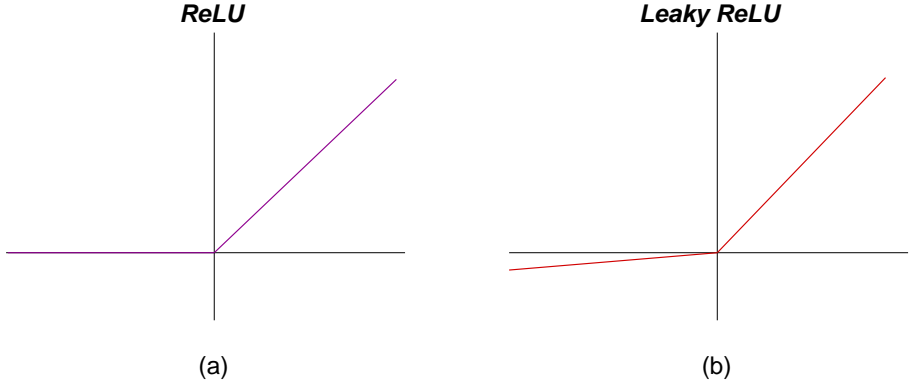}
\caption{(a) ReLU activation function $[\;]_+ = \sigma_1$. (b) Leaky ReLU activation function $\sigma_\lambda$, $\lambda \approx 0.9$.}
\label{fig: LeakyReLU}
\end{figure}

If neuron $N$ in a hidden layer
receives $d = d(N)$ inputs $x_{1},\ldots, x_{d}$ from neurons $N_{j_1},\ldots,N_{j_d}$ in the preceeding layer, and if the connection 
$N_{j_i} \arr N$ has weight $w_i$, then the output of $N$ is given by $\sigma(\ww\xx)$, where $\xx = (x_1,\ldots,x_d) \in \real^d$ is the vector of inputs to $N$ (a 
$d \times 1$-column matrix),  
$\ww = (w_1,\ldots,w_d)\in (\real^d)^\star$ is the \emph{parameter} or \emph{weight} vector (a linear functional on $\real^d$ or $1 \times d$-row matrix), 
$\ww\xx \in \real$ is matrix multiplication, and
$\sigma : \real\arr \real$ is the
\emph{activation function}.  See Figure~\ref{fig: ann}(b). 
Many different types of activation function have been proposed starting with the sign function used in the \emph{perceptron} model suggested 
by Rosenblatt~\cite{RO1958}. These activation functions often
possess the \emph{universal approximation property} (see Pinkus~\cite{Pinkus1999} for an overview current in 1999,
and \cite{SonodaMurata2017} for more recent results).
In this article, the focus is on the ReLU activation function $[\;]_+$ defined by
\[
\sigma(x) = [x]_+ \defoo \max (x,0), \; x \in \real.
\]
The ReLU activation function is commonly used in deep neural nets \cite[Chap 6]{IG2017},\cite{Ramachandran2017}, sometimes with a neuron dependent bias $b \in \real$ 
($\sigma(\ww\xx)$ is replaced by $\sigma(\ww\xx + b)$).
Advantages of ReLU include speed and the ease of applicability 
for back propagation and gradient descent used for training~\cite{MP1969}. 
A potential disadvantage of ReLU is `neuron death': if the input to a neuron is negative, there is no output and so no adaption of the input weights. 
One approach to this problem is the \emph{leaky ReLU} activation function defined for $\lambda \in [0,1]$ by
\[
\sigma_\lambda(x) = \max ((1-\lambda) x, x)
\]
($1-\lambda$ rather than the standard $\lambda$ is used for reasons that will become clear later).
Typically $\lambda $ is chosen close to 1, say $\lambda = 0.99$ (see Figure~\ref{fig: LeakyReLU}). 
The curve  $\{\sigma_\lambda \dd \lambda \in [0,1]\}$ of activation functions connects the ReLU activation $\sigma_1 = \sigma$ to $\sigma_0$
which is a linear activation function. The neural net defined by
$\sigma_0$ is tractable but not interesting for applications (the universal approximation property fails) though, as we shall see, $\sigma_0$  
plays an unexpected role in our approach: the associated neural net encodes important information about the neural net associated to $\sigma$.

\subsection{Student-Teacher model} 
In this work, we focus on an optimization problem originating from the training of a neural network (student) using  a planted model (teacher).  
This is also referred to as the
\emph{realizable}  setting where the labels of the samples in the underlying distribution are generated by the (teacher) neural network.
We use the simplest model here---inputs lie in $\real^d$, there are $k$ neurons and $d \ge  k$. Most of our analysis assumes $d= k$. This
is no loss of generality as our results extend naturally to $d \ge k$~\cite[\S 4.2]{SafranShamir2018}, \cite[\S E]{ArjevaniField2020b}.
This model is frequently used in theoretical investigations (for example, \cite{BrutzkusGloberson2017,Du2017,Li2017,Tian2017,Panigrahy2018}).

In more detail, assume $d \ge k$. Suppose that $\xx \in \real^d$ (input variable), and $\ww^1,\ldots, \ww^k$ are linear functionals on $\real^d$ (the \emph{parameters} or \emph{weights} and viewed as row vectors). 
Let $\cal{V} = \{\vv^1,\ldots,\vv^s\}$ be a fixed set of non-zero parameters with $s \le k$. We refer to $\cal{V}$ as the set of target weights (or just the target) 
used in the training of the neural net (student).  The term \emph{ground truth} is sometimes used for $\cal{V}$.
If $s < k$, the network is \emph{over-specified} 
(\emph{over-parametrized} in~\cite{SafranShamir2018}, where
$k$ signifies the number of inputs, $n$ the number of neurons).   
In this article, we focus on the case $s = d =k$.  For the present, however, assume that $s \le k \le d$

Let $M(k,d)$ denote the space of real $k \times d$ matrices. If $\is{W} \in M(k,d)$, denote the $i$th row of $\is{W}$ by $\is{w}^i$, $i \in \is{k}$; 
conversely let $\WW \in M(k,d)$ denote the matrix in $M(k,d)$ determined by the parameters (rows) $\ww^1,\cdots,\ww^k$. 
If $s = k$, $\VV \in M(k,d)$ is determined by $\cal{V}$ (the order of rows in $\VV$ is immaterial, but it is convenient to assume row $i$ of $\VV$ is $\vv^i$). 
If $s < k$, add zero rows 
$\vv^{s+1},\cdots, \vv^k$ to $\cal{V}$ so as to define $\VV \in M(k,d)$.  More generally, if $s < k$, $\cal{V}$ defines $\VV^s \in M(s,s)$ and we
extend $\VV^s$ to $\VV \in M(k,d)$ by  
appending $d-s$ zeros to each row of $\VV^s$ and then adding $k-s$ zero rows. 
\begin{rem}
In view of our use of (matrix) representation theory, 
we prefer to represent $\WW$ as a matrix rather than as a vector (element of $\real^{k \times d}$). In turn, this implies a strict adherence to 
viewing parameters as linear functionals (elements of the dual space $(\real^d)^\star$) 
and so row vectors---$1 \times d$ matrices.  In the literature,
$\ww\xx$ is often written as $\ww^T\xx$. In our context, this is confusing as $\ww$ is being treated both as a column (for $\ww^T\xx$) and as a row (in the matrix $\WW$). 
See also Section~\ref{sec: prelim}.
\rend
\end{rem}

If $\cal{V}$ contains $s \le k$ parameters, we define the loss function by 
\begin{equation}
\label{eq: sqloss}
\mathcal{L}(\WW,\VV) = \frac{1}{2}\mathbb{E}_{\xx \sim \cal{N}(0,I_d)}\left(\sum_{i=1}^k \sigma(\ww^i\xx) - \sum_{i=1}^s \sigma(\vv^i\xx)\right)^2
\end{equation}
The expectation gives the average as a function of $\WW,\VV$ assuming the inputs $\xx$ are distributed according to the zero mean, unit variance
Gaussian distribution on $\real^d$ (other distributions may be used, see~\cite{ArjevaniField2019c} and also Remarks~\ref{rem: dist}(1) at the end of Section~\ref{sec: ReLU}).  

Fixing $\VV$, define the \emph{objective} function
$\cal{F}:M(k,d)\arr \real$ by $\cal{F}(\WW) = \mathcal{L}(\WW,\VV)$. Thus $\cal{F}$ is a statistical average over the inputs of a $k$-neuron 2-layer neural net with ReLU activation.

Various initialization schemes are used. For example, initial weights $\ww^i$ can be sampled \emph{iid} from the normal distribution 
on $\real^d$ with zero mean and covariance matrix $k^{-1} I_d$ (\emph{Xavier initialization}~\cite{Xa}) and
stochastic gradient descent (SGD) applied to find a minimum value of $\mathcal{F}$.  Empirically, it appears that under gradient descent
there is convergence, with probability 1, to a local minimum value of $\mathcal{F}$. This is easy to prove if maps are $C^2$, 
proper\footnote{For a proper map level sets are compact.}, bounded below, and all critical points are
non-degenerate (non-singular Hessian). However, $\cal{F}$ is not $C^1$ on $M(k,d)$ and might have degenerate saddles
($0$ is not a local minimum of the proper map $f(x)=x^4/4+x^3/3$ but every trajectory $x(t)$ of $x'=-\grad{f}$
converges to $0$ as $t \arr +\infty$ if $x(0)> 0$).

Since $\cal{L} \ge 0$ and $\cal{L}(\VV,\VV) = 0$, $\mathcal{F}(\WW)$ has global minimum value zero which is attained when $\WW = \VV$.   
If a local minimum of $\cal{F}$ is not zero, it is called \emph{spurious}. In general, minima obtained by gradient descent may be 
spurious (see \cite[\S 3]{Swirszcz2017} for examples with just one neuron in the hidden layer).
Nevertheless, for the optimization problem considered here, there was the possibility that if strong conditions were imposed on $\VV$---for example, if $d = k=s$
and the rows of $\VV$ determine an orthonormal basis of $\real^k$---then convergence would 
be to the global minimum of $\mathcal{F}$.
However, Safran \& Shamir showed, using analytic estimates and numerical methods based on variable precision arithmetic, 
that if $6 \le k \le 20$, then spurious local minima are common even with these strong assumptions on $\VV$~\cite{SafranShamir2018}.
Their work suggested that (a) as $k$ increased, convergence to a spurious local minima was the default rather than the exception, and (b)
over-specification (choosing more neurons than parameters in the target set $\cal{V}$---$s  = d < k$), made it less likely that 
convergence would be to a spurious minimum.  It was also noted that the spurious minima had some symmetry. 
The symmetry of the parameter values determining spurious minima is, in part, a reflection of the symmetry of $\VV$~\cite{ArjevaniField2019c}.  

Although $\mathcal{L}$ is easily seen to be continuous, it is not everywhere differentiable as a function of $(\WW,\VV)$. 
However, explicit analytic formulas can be given for $\cal{L}$, $\mathcal{F}$ and $\grad{\mathcal{F}}$~\cite{ChooSaul2009,BrutzkusGloberson2017,Tian2017} and
from these it follows that $\mathcal{F}$ will be  \emph{real analytic}
on a full measure open and dense subset of the parameter space $M(k,d)$ that can be described precisely---the domain of analyticity domain depends strongly on 
the geometry determined by $\VV$.
In the case where $d=k=s$ and (say) $\VV = I_k$, 
real analyticity makes it possible to obtain precise quantitative results about the critical points of $\mathcal{F}$ for arbitrarily large $k$
as well as the asymptotics of key invariants, such as the value of the 
objective function at critical points of spurious minima, in terms of $1/\sqrt{k}$ or $1/k$.  

Although this model is relatively simple, 
the critical point structure is complex and mysterious. Symmetry based methods offer ways to 
illuminate the underlying structures and understand how they may change through symmetry breaking. 

\subsection{Outline of paper and main results}
The paper divides naturally into three main parts. Sections \ref{sec: prelim}---\ref{sec: isot_struct} are elementary and cover 
the required background and foundational material on symmetry and the student teacher model.
Sections 6 and 7 focus on the indirect method for finding critical points and the consistency equations. Section~\ref{sec: asym} is devoted to infinite series representations
of critical points and contains the main results and applications. Prerequisites for Section~\ref{sec: asym} include 
Sections \ref{sec: prelim}---\ref{sec: isot_struct} but only certain subsections of Sections 6 and 7 (indicated at appropriate points).
A discussion of contributions to machine learning and dynamics is in Section~\ref{sec: comm}.



\subsubsection{Sections~\ref{sec: prelim}---\ref{sec: isot_struct}}
Preliminaries on notational conventions and real analyticity are given in Section~\ref{sec: prelim}. Section~\ref{sec: foundations} is
devoted to groups, group actions and orthogonal representations. Modulo a familiarity with basic definitions from group
theory, the section is self-contained and includes the definition of the \emph{isotropy group} and \emph{fixed point space} of an action---both notions
play a central role in the paper. A short proof of the isotypic decomposition for orthogonal representations is included.
The focus throughout the section is on the natural orthogonal action of $S_k \times S_d$ on $M(k,d)$ (the first factor permutes rows, the second 
columns), and we describe the  associated isotypic decomposition of $M(k,d)$. We review basic definitions of equivariant maps (maps commuting with a group action), 
and invariant functions and verify that the gradient vector field of a $G$-invariant function is $G$-equivariant. The section concludes with brief comments about
critical points, symmetry breaking and maximal isotropy subgroup conjectures.

In Section~\ref{sec: ReLU} we give analytic expressions for the loss and objective functions of ReLU and leaky ReLU nets when the 
loss function is given as the expectation over an orthogonally invariant distribution and establish symmetry and regularity properties
of the loss and objective function.  Assuming here for simplicity that $d = k$, the objective function is $S_k \times S_k$-invariant if $\VV = I_k$.
The objective functions $\cal{F}_\lambda:M(k,k) \arr \real$, $\lambda \in (0,1]$, are real analytic on the complement $\Omega_a$ of a thin  $\lambda$-independent algebraic subset of $M(k,k)$
($\Omega_a$ will be open and dense in $M(k,k)$).
After giving explicit formulas for $\grad{\cal{F}_\lambda}$, we conclude with examples of critical points of $\cal{F}$ and a proof that
if $\VV = I_k$ and $\cal{F}_\lambda(\WW) = 0$ (the global minimum of $\cal{F}$), then $\WW$ lies on the $S_k\times S_k$-orbit of $\VV$.

In Section~\ref{sec: isot_struct}, we give results on the isotropy groups that occur for the action of $S_k \times S_k$ on $M(k,k)$ with an 
emphasis on isotropy conjugate to maximal proper subgroups of the diagonal group $\Delta S_k = \{(g,g) \dd g \in S_k\}$. These groups 
are known to appear as isotropy groups of critical points of $\cal{F}$ giving spurious minima. A much used fact
is that if $\WW \in M(k,k)$ has isotropy group $H\subset S_k \times S_k$, then $\WW$ is a critical point of $\cal{F}$ (or $\cal{F}_\lambda$) if and only if
$\WW$ is a critical point of $\cal{F}|M(k,k)^H$, where $M(k,k)^H$ is the fixed point space for the action of $H$ on $M(k,k)$.  The
dimension of $M(k,k)^H$ is often small compared with $k^2 = \text{dim}(M(k,k)$ and may be `independent' of $k$. For example,
if $H = \Delta S_{k-1}$, then $\text{dim}(M(k,k)^H = 5$, for all $k \ge 3$. Finally, we give parametrizations for
the fixed point spaces of $\Delta S_{k}, \Delta S_{k-1}$ and $\Delta (S_{k-p} \times S_p)$ and derive the 
equations for critical points used in Section~\ref{sec: asym}. 

\subsubsection{Sections~\ref{main-section} and \ref{sec: comp}}
In Section~\ref{main-section} we start work on obtaining precise analytic results about critical points of $\cal{F}$ lying in $\Omega_a$; in particular, certain natural families
of critical points that are associated with spurious minima. We adopt two approaches. One direct, covered under Section~\ref{sec: asym} below, 
and an indirect method---the main topic of Sections~\ref{main-section} and \ref{sec: comp}. Suppose that $\mathfrak{c}\in M(k,k)$ is a critical point of
$\cal{F}_1 = \cal{F}$. In many interesting cases we can construct a real analytic path $\{\bxi(\lambda)\dd \lambda\in [0,1]\}$ in $M(k,k)$ such that
$\bxi(1) = \mathfrak{c}$ and $\bxi(\lambda)$ is a critical point of $\cal{F}_\lambda$, for all $\lambda \in [0,1]$. Surprisingly, it turns out to
be straightforward to give equations for $\bxi(0)$; equations that are simpler than the defining equation $\grad{\cal{F}} = 0$ for $\bxi(1) = \mathfrak{c}$. 
This despite the critical points of $\cal{F}_0$ being highly degenerate:  they form a codimension $k$ hyperplane in $M(k,k)$.
Moreover, if we know $\bxi(0)$, we can then construct the real analytic path $\bxi(\lambda)$ without knowing $\bxi(1) = \mathfrak{c}$ \emph{a priori}. We
refer to the equations that determine $\bxi(0)$ as the \emph{consistency equations}. Not only are they closely related to the
equations defining  $\mathfrak{c}$, but $\bxi(0)$ is typically a good approximation to $\mathfrak{c}$.  The path based approach is most useful when critical points have 
isotropy conjugate to a subgroup of $\Delta S_k$ and the isotropy is natural---the dimension of $M(k,k)^H$ is independent of $k$. For example,
$\text{dim}(M(k,k)^{\Delta S_k}) = 2$ and $\text{dim}(M(k,k)^{\Delta S_{k-1}}) = 5$, for all $k \ge 3$.  These families are of special interest because for $k \ge 6$,
there is one  spurious minimum with isotropy $\Delta S_k$, referred to as type A, and two additional 
spurious minima with isotropy $\Delta S_{k-1}$, referred to as types I and II (for type I to be spurious, we require $k \ge 7$). The type II minima appears in~\cite[Example 1]{SafranShamir2018} but  
not the minima of types I and A. The $S_k \times S_k$-orbits of any of these critical points define spurious minima and so, for example, there are 
$k^2 (k-1)!$ critical points of type II. We have found 4 critical points in $M(k,k)^{\Delta S_{k-1}}$ that define local minima: $\VV$ (global), types A, I \& II. 
In Section~\ref{main-section} we outline general methods for obtaining the consistency equations and constructing the
paths. In Section~\ref{sec: comp}, we carry out the main steps of the construction for critical points of type A and type II. 
We include 
numerics to illustrate results and conclude with a brief discussion of another family of proper maximal isotropy groups $\Delta (S_{k-2} \times S_2)$ that
supports spurious minima for $k \ge 9$~\cite{ArjevaniField2020c}. 

\subsubsection{Section \ref{sec: asym}}
We obtain convergent power series in $1/\sqrt{k}$ for the critical points associated to families of natural isotropy subgroups of $\Delta S_k$. For example,
suppose $\mathfrak{c} \in M(k,k)^{\Delta S_{k-1}}$ is a critical point of type II. Since  $M(k,k)^{\Delta S_{k-1}} \approx \real^5$, we
may specify $\mathfrak{c}$ using 5 real parameters, $\xi_1,\cdots,\xi_5$ ($\xi_1$, resp.~$\xi_5$,  corresponds to the diagonal entries $\mathfrak{c}_{ii}, i < k$, resp.~$\mathfrak{c}_{kk}$---see
Section~\ref{ex: deltak-1ex} for details). We have\\
\noindent {\bf Theorem} (Theorem~\ref{thm: initial_terms})\\
If $\mathfrak{c}=(\xi_1,\cdots,\xi_5)$ is of type II, then 
\[
\xi_1 = 1 + \sum_{n=4}^\infty c_n k^{-\frac{n}{2}},\quad
\xi_2 = \sum_{n=4}^\infty e_n k^{-\frac{n}{2}},\quad \xi_5 = -1 + \sum_{n=2}^\infty d_n k^{-\frac{n}{2}} \]
\[
\xi_3 = \sum_{n=2}^\infty f_n k^{-\frac{n}{2}},\quad
\xi_4 = \sum_{n=2}^\infty g_n k^{-\frac{n}{2}}
\]
(The first two non-constant coefficients are given in Section~\ref{sec: asym}.)

Similar results are given for critical points of types I and A. In all cases, the series converge for sufficiently large values of $k$---we suspect 
convergence holds for all $k$ in the range of interest (that is, $k \ge 3$ for types A, I and II) and also that 
the constant terms (notably the sign), and initial exponents of the non-constant terms, uniquely determine
the critical point. Similar results hold for the solutions of the consistency equations and the initial terms of the series match those for the critical points
(the first two non-constant terms in the case of type II points).

These results allow 
computation of the decay of the objective function at critical points defining spurious minima. For example, we show that for type II critical points $\mathfrak{c}_k$
$\mathcal{F}(\is{c}_k)  = (\frac{1}{2}-\frac{2}{\pi^2})k^{-1}+O(k^{-\frac{3}{2}})$.
Critical values associated to critical points of types I and A, converge to a strictly positive constant as $k \arr\infty$ (Section~\ref{sec: slodecay}).
Using these power series representations, we can give precise estimates on the spectrum of the Hessian---and so verify that critical points of types I, II and A do indeed
define spurious minima for all $k \ge 6$~\cite{ArjevaniField2020b}.

A further consequence of these results is that as $k \arr \infty$,  a type II critical point converges to the matrix defined by 
the parameter set
$\{\vv^1,\vv^2,\ldots,\vv^{k-1},-\vv^k\}$. In a sense the spurious 
minima arise from a ``glitch'' in the optimization algorithm that allows convergence of $\ww^k$ to $-\vv^k$. The decay $O(k^{-1})$
of $\mathcal{F}(\is{c}_k)$ appears of because of cancellations involving differing rates of convergence of
$\ww^i$ to $\vv^i$, $i < k$ (fast) and $\ww^k$ to $-\vv^k$ (slow). Types I and A spurious minima show a similar pattern of convergence, 
but now with all (resp.~$(k-1)$) parameters 
converging to $-\vv^i$ for type A (resp.~type I).  

More is said about future directions and other results in the discussion section~\ref{sec: comm}.

\section{Preliminaries}\label{sec: prelim}
\subsection{Notation \& Conventions}\label{sec: notation}
Let $\pint$ denote the natural numbers---the strictly positive integers---and $\intg$ denote the set of all integers.
Given $k \in \pint$, define  $\is{k}=\{1,\ldots, k\}$ and
let $S_k$ denote the \emph{symmetric group} of permutations of $\is{k}$.
The symbols $\is{k}, \is{m}, \is{n},\is{p}, \is{q}$ are reserved for indexing. For example,
\[
\sum_{i=1}^m\sum_{j=1}^n a_{ij} = \sum_{(i,j) \in \is{n}\times \is{m}} a_{ij},
\]
otherwise boldface lower case is used to denote vectors. 

Let $\langle\xx,\yy\rangle$ denotes the Euclidean inner product of $\xx,\yy\in \real^n$ and $\|\xx\| = \langle\xx,\xx\rangle^{\frac{1}{2}}$
denotes the associated Euclidean norm. 
We always assume $\real^n$ is equipped with the Euclidean inner product and norm and the standard orthonormal basis (denoted by $\{\vv_j\}_{j\in\is{n}}$),
and that every vector subspace of $\real^n$ has the inner product induced from
that on $\real^n$. Let $\On{n}$ denote the \emph{orthogonal group} of $\real^n$---we often identify $\On{n}$ with the group of $n \times n$ orthogonal matrices.
If $\xx_0 \in \real^n$ and $r > 0$, then $D_r(\xx_0) = \{\xx \dd \|\xx-\xx_0\| < r\}$ (resp.~$\overline{D}_r(\xx_0)$) denotes the 
open (resp.~closed) Euclidean $r$-disk, centre $\xx_0$.  

For $k,d \in \pint$, $M(k,d)$ denotes the vector space of real $k \times d$ matrices (parameter vectors).  Matrices in $M(k,d)$ are usually denoted 
by boldfaced capitals. If $\WW \in M(k,d)$,
then $\is{W} = [w_{ij}]$, where $w_{ij}\in\real$, $(i,j) \in \is{k} \times \is{d}$. Let $\ww^i$ denote row $i$ of $\WW$, $i \in \is{k}$, and write
$\ww^i$ in coordinates as $(w_{ij})$. 
If $\xx \in \real^d$, $\ww^i\xx$, $\WW\xx$ are defined by \emph{matrix multiplication}---there is no use of the transpose. 
Given $\WW\in M(k,d)$, $\WW^\Sigma$ is the row vector obtained by summing the \emph{rows} (not columns) of $\WW$: 
$\WW^\Sigma = \sum_{i\in\is{k}} \ww^i\in M(1,d)$ and $\WW^\Sigma = (\sum_{i\in\is{k}}w_{ij})$.

On occasions, $M(k,d)$ is identified with $\real^{k \times d}$. For this we concatenate the rows of $\WW$ and map $\WW$ to $\ww^1\ww^2\cdots\ww^k$. 
The inner product on
$M(k,d)$ is induced from the Euclidean inner product on $\real^{k \times d}$ and
\[
\|\WW\| = \|\ww^1\ww^2\cdots\ww^k\| =\|(\ww^1,\cdots,\ww^k)\| = \sqrt{\sum_{i\in\is{k}} \|\ww^i\|^2}.
\]
All vector subspaces of $M(k,d)$ inherit this inner product.

For $a \in \{0,1\}$, let $\mathbf{a}_{p,q} \in M(p,q)$ be the matrix with all entries equal to $a$. If $a = 1$, let
$\mathbf{I}_{p,q}=\{t\mathbf{1}_{p,q} \dd t\in \real\}$ denote the line in $M(p,q)$ through $\mathbf{1}_{p,q}$. The subscripts $p,q$ may be omitted if 
clear from the context.
The identity $k\times k$-matrix $I_k$ plays a special
role and is denoted by $\VV$ when used as the target or ground truth.

Real analytic maps and the real analytic implicit function theorem play an important role and we recall that
if $\Omega \subset \real^n$ is a non-empty open set, then $f: \Omega \arr \real^m$ is \emph{real analytic} if
\begin{enumerate}
\item $f$ is  smooth ($C^\infty$) on $\Omega$.
\item For every $\xx_0 \in \Omega$, there exists $r > 0$ such that the Taylor series of $f$ at $\xx_0$ converges to
$f(\xx)$ for all $\xx \in D_r(\xx_0)\cap \Omega$.
\end{enumerate}
The basic theory of real analytic functions, using methods of \emph{real} analysis, is given in Krantz \& Parks~\cite{Krantz1992}.
However, it is often easier to complexify and use complex analytic results.

Finally, we use the abbreviation `iff' for `if and only if'.

\section{Groups, actions and symmetry}\label{sec: foundations}
After a review of group actions and representations, we  
give required definitions and results on equivariant maps. 
The section concludes with comments on symmetry breaking. 
\subsection{Groups and group actions}
\label{sec:symm}
Elementary properties of groups, subgroups and  group homomorphisms are assumed known.
The identity element of a group $G$ will be denoted by $e_G$ or $e$ and composition will be multiplicative.

\begin{exam}[Permutation matrices]\label{ex:ortho_Sn}
The symmetric group $S_n$ of permutations of $\is{n}$ is naturally isomorphic
to the subgroup ${P}_n$ of $\On{n}$ consisting of \emph{permutation matrices}: 
if $\eta \in S_n$, $[\eta] \in {P}_n$ is the matrix of the
orthogonal linear transformation $\eta(x_1,\cdots,x_n) = (x_{\eta^{-1}(1)},\cdots,x_{\eta^{-1}(n)})$.
\end{exam}
\begin{Def}
Let $G$ be a group and $X$ be a set. An \emph{action} of $G$ on $X$ consists of a map 
$ G \times X \arr X;\,(g,x) \mapsto g x $
such that
\begin{enumerate}
\item For fixed $g \in G$, $x\mapsto gx$ is a bijection of $X$.
\item $e x = x$, for all $x \in X$. 
\item $(gh)x = g(hx)$ for all $g,h \in G$, $x \in X$ (associativity).
\end{enumerate}
We call $X$ a \emph{$G$-set} (or \emph{$G$-space} if $X,G$ are topological spaces and the action is continuous).
\end{Def}
\begin{rem}
In what follows we always assume the action is \emph{effective}: $gx = x$ for all $x \in X$ iff $g = e$.\rend
\end{rem}

\begin{exam}[The $\Gamma_{k,d}$ space $M(k,d)$]\label{ex: Mkn}
Let $k,d \in \pint$ and set $\Gamma_{k,d} = S_k \times S_d$. Then $M(k,d)$ has the structure of a $\Gamma_{k,d}$-space with action defined by 
\begin{equation}\label{eq: Gamma-action}
(\rho,\eta)[w_{ij}] = [w_{\rho^{-1}(i),\eta^{-1}(j)}],\; \rho \in S_k, \eta \in S_d,\; [w_{ij}] \in M(k,d).
\end{equation}
Elements of $S_k$ (resp.~$S_d$) permute the \emph{rows} (resp.~\emph{columns}) of $[w_{ij}]$. The action is natural on columns and rows
in the sense that if $\rho \in S_k$ and $\rho(i) = i'$ then $\rho$ moves row $i$ to row $i'$; similarly for the action on columns. 
Identifying $S_k, S_d$ with the corresponding groups of permutation matrices, the action of $\Gamma_{k,d}$ on $M(k,d)$ may be written 
\[
(\rho,\eta) \WW = [\rho] \WW [\eta]^{-1},\; (\rho,\eta) \in \Gamma_{k,d}.
\]
\end{exam}
\subsubsection{Notational conventions}
We reserve
the symbols $\Gamma_{k,d},\, \Gamma$ for the group $S_k \times S_d$, 
with associated action on $M(k,d)$ given by \Refb{eq: Gamma-action}.
Subscripts $k,d$  are omitted from $\Gamma_{k,d}$
if clear from the context.   

Let $S_k^r$ and $S_d^c$ respectively denote the subgroups $S_k \times \{e\}$ and $\{e\}\times S_d$ of $\Gamma$
and note that $S_k^r$ (resp.~$S_d^c$) permutes rows (resp.~columns).
\subsubsection{Geometry of $G$-actions}
Given a $G$-set $X$ and $x \in X$, define 
\begin{enumerate}
\item $Gx = \{gx \dd g \in G\}$ to be the \emph{$G$-orbit} of $x$.
\item $G_x = \{g \in G \dd gx = x\}$ to be the \emph{isotropy} subgroup of $G$ at $x$.
\end{enumerate}
\begin{rem}
The isotropy subgroup of $x$ measures the `symmetry' of the point $x$, relative to the $G$-action: the more symmetric the point
$x$, the larger is the isotropy group. 
Subgroups $H, H'$ of $G$ are \emph{conjugate} if
there exists $g \in G$ such that $gHg^{-1} = H'$. Points $x, x' \in X$ have the same \emph{isotropy type} (or symmetry) if
$G_x, G_{x'}$ are conjugate subgroups of $G$. Since $G_{gx} = gG_x g^{-1}$, points on the same $G$-orbit have the same isotropy type.
\rend
\end{rem}
\begin{Def}
The action of $G$ on $X$ is \emph{transitive} if for some (any) $x \in X$, $X = Gx$. The action is \emph{doubly transitive} if
for any $x \in X$, $G_x$ acts transitively on $X \smallsetminus \{x\}$.
\end{Def}
\begin{rem}
If the action on $X$ is transitive, all points of $X$ have the same isotropy type. 
The action is \emph{doubly transitive} iff for all
$x,x',y,y' \in G$, $x \ne x'$, $y \ne y'$, there exists $g \in G$ such that $gx = y$, $gx' = y'$.  \rend
\end{rem}
\begin{exams}
(1) The action of  $\Gamma_{k,d}$ on $\is{k} \times \is{d}$ defined by
\[
(\rho,\eta)(i,j) = (\rho^{-1}(i),\eta^{-1}(j)),\; \rho \in S_k, \eta \in S_d,\; (i,j) \in \is{k} \times \is{d},
\]
is transitive but not doubly transitive if $k,d \ge 2$. \\
(2) Set $\Delta S_k = \{(\eta,\eta) \dd \eta \in S_k\}\subset S_k^2$---the \emph{diagonal} subgroup of
$S_k^2$. If $k > 1$, the action of $\Delta S_k$ on $\is{k}^2$ is not transitive: there are two group orbits, the diagonal $\Delta \is{k}=\{(j,j) \dd j \in \is{k}\}$
and the set of all non-diagonal elements.
\end{exams}
Given a $G$-set $X$ and subgroup $H$ of $G$, let 
\[
X^H = \{y \in X \dd hy =y, \forall h\in H\}
\]
denote the fixed point space for the action of $H$ on $X$.
\begin{rem}\label{rem:fixedptspace}
Note that $x \in X^H$ iff $G_x \supset H$. Consequently, if $H = G_{x_0}$ for some $x_0 \in X$, then $G_x \supseteq G_{x_0}$
for all $x \in X^H$.
\rend
\end{rem}

\subsection{Orthogonal representations}
We give a short review of representation theory that suffices for our
applications (for more detail and generality see~\cite{Brocker1985,Thomas2004}).

Let $(V,G)$ be a $G$-space. If $V$ is a finite dimensional inner product space
and $g: V \arr V$  is orthogonal for all $g \in G$, then $(V,G)$ is called an \emph{orthogonal $G$-representation}.
\begin{rems}
(1) Aside from the standard action of $\On{n}$ on $\real^n$, our focus will be on representations of \emph{finite}
groups $G$ where the continuity of the action is trivial. Since we assume actions are effective, we may regard
$G$ as a subgroup of $\On{n}$, rather than as a homomorphic image of $G$ in $\On{n}$. Of course, any finite group
embeds in $S_n$ (Cayley's theorem), and so in $\On{n}$ (Example~\ref{ex:ortho_Sn}), for all large enough $n$. \\
(2) All the representations we consider will be orthogonal and we usually omit the qualifier `orthogonal'. \rend
\end{rems}

\begin{exams}\label{ex: dual}
(1) $(\real^n,\On{n})$ is an $\On{n}$-representation. It may be shown that $\On{n}$ is a real analytic submanifold of $M(n,n)$ and that
the action is real analytic~\cite{Brocker1985}.\\
(2) The group $\Gamma_{k,d}$ is naturally a subgroup of $\On{kd}$, via the identification 
of $M(k,d)$ with $\real^{k\times d}$, and so $(M(k,d),\Gamma_{k,d})$ is a
$\Gamma_{k,d}$-representation with linear maps acting orthogonally on $M(k,d)$. \\
(3) Suppose  $(V,G)$ is an orthogonal representation and let $V^*$ denote the \emph{dual space} of $V$---that is $V^\star$ is the space of
linear functionals $\phi:V\arr \real$. Define the \emph{dual representation} $(V^\star,G)$ by $g \phi = \phi \circ g^{-1}$, $\phi\in V^*$, $g \in G$
(the use of $g^{-1}$, rather than $g$, assures associativity of the action).
Right multiplication 
by permutation matrices on $M(k,d)$ described in Example~\ref{ex: Mkn}, is an action (of $S_d$) 
and the rows of $\WW\in M(k,d)$ transform like linear functionals:  
$\ww^i \mapsto g\ww^i = \ww^i \circ g^{-1}$.
\end{exams}

\subsubsection{Isotropy structure for representations by a finite group.}
If $G$ is a finite subgroup of $\On{n}$ there are only 
finitely many different isotropy groups for the action of
$G$ on $\real^n$. If $H$ is an isotropy group for the action of $G$, define $F_{(H)} = \{y \in \real^n \dd G_y = H\}$ and note that
$F_{(H)} \subset (\real^n)^H$.
\begin{lemma}\label{lem:struct}
If $G \subset \On{n}$ is finite, then
\begin{enumerate}
\item $\{F_{(H)} \dd H\;\text{is an isotropy group}\}$ is a partition of $\real^n$.
\item $\overline{F_{(H)}} = (\real^n)^H$, for all isotropy groups $H$.
\end{enumerate}
\end{lemma}
\proof (1) is immediate; for (2), see~\cite[Chapter 2, \S 9]{Field2007}. \qed 
\begin{rem}\label{rems: isot_struct}
If $\gamma: [0,1]\arr (\real^n)^H$ is a continuous curve and $G_{\gamma(t)} = H$
for $ t < 1$, then $G_{\gamma(1)} \supseteq H$. The inclusion may be strict. \rend
\end{rem}
\subsubsection{Irreducible representations}
Suppose that $(V,G)$ is an orthogonal $G$-representation. 
A vector subspace $W$ of $V$ is
\emph{$G$-invariant} if $g(W) = W$, for all $g \in G$.
The representation $(V,G)$ is \emph{irreducible} if the only $G$-invariant subspaces of $V$ are
$V$ and $\{0\}$.

\begin{lemma}\label{lem: rep}
(Notations and assumptions as above.) If $(V,G)$ is not irreducible, 
then $V$ may be written as an orthogonal direct sum $\bigoplus V_i$ of irreducible $G$-representations $(V_i,G)$. 
\end{lemma}
\proof The orthogonal complement of an invariant subspace is invariant. The lemma follows easily by induction on $m=\text{dim}(V)$. \qed

\begin{Def}\label{def:repeq}
Let  $(V,G)$, $(W,G)$ be representations. A linear map $A:V \arr W$ is a \emph{$G$-map} if
$ A(g\vv) = gA(\vv)$, for all $g \in G,\,\vv \in V $.\\
The representations $(V,G)$, $(W,G)$ are \emph{($G$)-equivalent} or \emph{isomorphic} if there exists a $G$-map $A: V \arr W$ which is a linear isomorphism.
\end{Def}
\begin{rem}\label{rem: schur}
If $(V,G)$, $(W,G)$ are irreducible and inequivalent, every $G$-map $A: V \arr G$ is zero ($\text{Ker}(A)$ and $\text{Im}(A)$
are $G$-invariant subspaces of $V$ and $W$ respectively). If $(V,G)$, $(W,G)$ are irreducible and equivalent, then every non-zero $G$-map $A: V \arr W$ is an 
isomorphism. \rend
\end{rem}

\begin{thm}\label{thm:iso}
(Notations and assumptions as above.)
If $(V,G)$ is a $G$-representation, then
there exist $k \in \pint$, $p_i \in \pint$, $i \in\is{k}$, and $G$-invariant subspaces $V_{ij}\subset V$, $i \in \is{k}$, $j \in \is{p}_i$, such that
\begin{enumerate}
\item
$ V$ is isomorphic to $\bigoplus_{i\in\is{k}}( \oplus_{j\in \is{p_i}}V_{ij})\, \;(\text{orthogonal direct sum})$.
\item $k$ and $p_i$, $i \in k$, are uniquely determined by $(V,G)$.
\item The representations $(V_{ij},G)$ are all irreducible and $(V_{ij},G)$ is isomorphic to $(V_{i'j'},G)$ iff $i = i'$.
\item The subspaces $V_i = \oplus_{j\in \is{p_i}}V_{ij}$ are uniquely determined by $(V,G)$; the representations $(V_{ij},G)$
are uniquely determined up to isomorphism.
\end{enumerate}
\end{thm}
\proof A straightforward argument based on Lemma~\ref{lem: rep} and Remark~\ref{rem: schur} (see~\cite{Brocker1985,Thomas2004} for greater generality). \qed
\begin{rems}
(1) The decomposition of $(V,G)$ given by Theorem~\ref{thm:iso} is known as the \emph{isotypic} decomposition of 
$(V,G)$. If we let $\mathfrak{v}_i$ denote the isomorphism class of the representation $(V_{ij},G)$, $i \in \is{k}$, then the isomorphism class
of $(V,G)$ may be written uniquely (up to order) in the form $\oplus_{i \in \is{k}} p_i \mathfrak{v}_i$, where $p_i$ is the \emph{multiplicity} of the representation $\mathfrak{v}_i$ in $(V,G)$
(and $(V_i,G)$).  Note that although we assume $G$ is finite, the proof works for any closed subgroup $G$ of $\On{n}$.\\
(2) The subspaces $V_{ij}$ 
are not uniquely determined, unless $p_i = 1$. \\
(3) For a description of the space of $G$-maps of an irreducible $G$-representation and the proof that 
a finite group has only \emph{finitely} many inequivalent and irreducible $G$-representations, we refer to texts on the representation theory 
of finite groups (for example, \cite{Thomas2004}). \rend
\end{rems}
\subsubsection{Isotypic decomposition of $(M(k,d),\Gamma)$}\label{exam: isotypic}
We describe the isotypic decomposition of $(M(k,d),\Gamma)$. To avoid discussion of trivial cases, 
assume that $k,d > 1$. 
Define linear subspaces of $M(k,d)$ by
\begin{eqnarray*}
\mathbf{C} & = & \{\WW\in M(k,d) \dd \sum_{i\in\is{k}} w_{ij} = 0,\; j \in \is{d}\},\; (\text{column sums zero})\\
\mathbf{R}  & = & \{\WW \in M(k,d) \dd \sum_{j\in\is{d}} w_{ij} = 0,\; i \in \is{k}\},\; (\text{row sums zero})\\
\mathbf{A}  & = & \is{C} \cap \is{R},\qquad \mathbf{I}   =  \mathbf{I}_{k,d}=\real \mathbf{1}_{k,d}.
\end{eqnarray*}
Observe that $\is{C},\is{R},\is{A}$ and $\is{I}$ are all proper $\GG$-invariant subspaces of $M(k,d)$ and
$M(k,d) = \is{C}+\is{R}+\is{A}+\is{I}$. Since $\is{C},\is{R} \supsetneq \is{A}$, the representations  $\is{C}, \is{R}$ cannot be irreducible.
Let $\is{C}_1$ be the orthogonal complement of $\is{A}$ in $\is{C}$ and $\is{R}_1$ be the orthogonal
complement of $\is{A}$ in $\is{R}$. It is easy to check that the subspaces $\is{C}_1$, $\is{R}_1$, $\is{A}$ and $\is{I}$ are mutually orthogonal. Moreover,
the rows of $\is{R}_1$ (resp.~columns of $\is{C}_1$) are identical and given by the solutions of $r_1+\ldots+ r_d = 0$ (resp.~$c_1+\ldots+ c_c = 0$). 
Since it is well-known (and easy to verify) that the natural action of $S_p$ on the hyperplane $H_{p-1}\subset \real^p$: $x_1 + \cdots + x_p = 0$ is irreducible, the
representations $(\is{R}_1,S_k \times S_d)$ and $(\is{C}_1,S_k \times S_d)$ are irreducible. Finally, the representation $(\is{A},S_k \times S_d)$ is also
irreducible since it is isomorphic to the (exterior) tensor product of the irreducible representations $(H_{k-1},S_k)$ and $(H_{d-1},S_d)$. Summing up, 
\begin{enumerate}
\item $M(k,d) = \is{I} \oplus \is{C}_1 \oplus \is{R}_1 \oplus \is{A}$ is the unique decomposition of $(M(k,d),\GG)$ into an orthogonal direct sum of irreducible representations. 
In particular,
$\is{C}_1, \is{R}_1,\is{A},\mathbf{I}$ are irreducible and inequivalent $\GG$-representations.
\item $\text{dim}(\is{A}) = (k-1)(d-1)$, $\text{dim}(\is{C}_1) = k-1$,  $\text{dim}(\is{R}_1) = d-1$.
\end{enumerate}
\begin{rem}
The isotypic decomposition of $(M(k,k),\Gamma)$ is simple to obtain. However, an analysis of the eigenvalue structure of the Hessian of $\cal{F}$
requires the isotypic decomposition of $M(k,k)$, viewed as an $H$-representation, where $H \subseteq \Delta S_k$; this is less trivial~\cite{ArjevaniField2020b}.
\rend
\end{rem} 
\subsection{Invariant and equivariant maps}\label{sec:equivmaps}
We review the definition and properties of invariant and equivariant maps. For more details, see \emph{Dynamics and Symmetry}~\cite[Chapters 1, 2]{Field2007}.

The action of $G$ on $X$ is \emph{trivial} if 
$gx = x,\; \text{for all } g\in G, x \in X$.
\begin{Def}
A map $f: X \arr Y$ between $G$-spaces is \emph{$G$-equivariant} (or \emph{equivariant}) if
$ f(gx) = g f(x),\; x\in X,\; g \in G$. \\
If the $G$-action on $Y$ is trivial, $f$ is 
($G$-)\emph{invariant}. That is, 
\[
f(gx) = f(x),\; x \in X, \; g \in G
\]
\end{Def} 
\begin{exams}
(1) $G$-maps are $G$-equivariant (Definition~\ref{def:repeq}).\\
(2) The norm function $\|\;\|$ on $\real^n$ is $G$-invariant for all $G \subset \On{n}$. 
\end{exams}
\begin{prop}\label{prop: equiv}
If $f:X \arr Y$ is an equivariant map between $G$-spaces $X, Y$, then
\begin{enumerate}
\item $G_{f(x)} \supset G_x$ for all $x \in X$. 
\item If $f$ is a bijection, then $f^{-1}$ is equivariant and $G_x = G_{f(x)}$ for all $x \in X$.
\item For all subgroups $H$ of $G$, $f^H\defoo f|X^H: X^H \arr Y^H$ and if $f$ is bijective, so is $f^H$.
\end{enumerate}
\end{prop}
\proof An easy application of the definitions.  For example, (3) follows since if $x \in X^H$, then $f(x) = f(hx) = hf(x)$, for all
$h \in H$. 
\qed

\subsection{Gradient vector fields} 
\begin{prop}\label{prop: grad}
If $G$ is a closed subgroup of $\On{m}$,  $\Omega$ is an open $G$-invariant subset of $\real^m$
and $f: \Omega \arr \real$ is $G$-invariant and $C^r$, $r \ge 1$, (resp.~analytic), then
the gradient vector field of $f$, $\grad{f}:\Omega \arr \real^m$, is $C^{r-1}$ (resp.~analytic) and $G$-equivariant.
\end{prop}
\proof For completeness, a proof is given of equivariance. Let $Df: \Omega \arr L(\real^m,\real); \xx \mapsto Df_\xx$, denote the derivative map of 
$f$ ($L(\real^m,\real) $ is the vector space of
linear functionals from $\real^m$ to $\real$). Since
$Df_\xx(\ee) = \lim_{t\arr 0} \frac{f(\xx + t\ee) - f(\ee)}{t}$, the invariance of $f$ implies that 
$Df_{g\xx}(g\ee) = Df_\xx(\ee)$, for all $\xx\in\Omega$, $\ee \in \real^m,\, g \in G$.
By definition,
$
\langle\grad{f}(\xx),\ee\rangle = Df_\xx(\ee)$, for all $\ee \in \real^m$.
Therefore, 
\begin{eqnarray*}
\langle\grad{f}(g\xx),\ee\rangle&=& Df_{g\xx}(\ee) = Df_\xx(g^{-1}(\ee))\\
& = & \langle\grad{f}(\xx),g^{-1}\ee\rangle = \langle g\,\grad{f}(\xx),\ee\rangle,
\end{eqnarray*}
where the last equality follows by the invariance of the inner product under the diagonal action of $G$. Since the final equality holds for all $\ee \in \real^m$,
$\grad{f}(g\xx) = g\,\grad{f}(\xx)$ for all $g \in G$, $\xx \in \Omega$.  \qed

\begin{lemma}\label{lem: symm}
(Assumptions and notation of Proposition~\ref{prop: grad}.) 
If $H\subset G$, then
\[
\grad{f|\Omega^H} = \grad{f} | \Omega^H,
\]
and $ \grad{f}| \Omega^H$ is everywhere tangent to $(\real^m)^H$. If $\isk{c} \in \Omega^H$ is a critical point of
$f|\Omega^H$, then
\begin{enumerate}
\item  $\isk{c}$ is a critical point of $f$ (and conversely).
\item Eigenvalues of the Hessian of $f|\Omega^H$ at $\isk{c}$ determine the subset of eigenvalues of the Hessian of $f$ at  $\isk{c}$
associated to directions tangent to $(\real^m)^H$.
\end{enumerate}  
\end{lemma}
\proof Follows by the equivariance of $\grad{f}$ and Proposition~\ref{prop: equiv}.\qed
\begin{rems}
(1) If $\isk{c}$ is a critical point of $f|\Omega^H$,
then $G\isk{c}$ is  group orbit of critical points of $f$ all with the same critical value $f(\is{c})$.
The eigenvalues of the Hessian at critical points are constant along $G$-orbits (the Hessians 
are all similar). If $G$ is not finite and $\text{dim}(G\isk{c}) > 0$, there will be zero eigenvalues corresponding to directions
along the $G$-orbit~\cite[Chapter 9]{Field2007}.\\
(2) For large $m$ it may be hard to
find local minima of $f$ (for example, using SGD). However,
the dimension of fixed point spaces $(\real^m)^H$ may be small and Lemma~\ref{lem: symm} 
offers a computationally efficient way of finding critical points of $f$ that lie in fixed point spaces. 
\rend
\end{rems}

\subsection{Critical point sets and Maximal isotropy conjectures.}
Let $f:\real^m \arr \real$ be $C^r$, $r \ge 2$.
Analysis of $f$ typically focuses on the set $\Sigma_f$ of critical points of $\grad{f}$ and their stability (given by the Hessian).   If
$f$ is $G$-equivariant, $\grad{f}$ restricts to
a \emph{gradient} vector field on every fixed point space $(\real^m)^H$ (Lemma~\ref{lem: symm}).  
If $\exists R > 0$ such that $(\grad{f}(\xx), \xx) < 0$ for $\xx\notin D_R(\is{0})$, then every forward
trajectory $\xx(t)$ of $\xxd = \grad{f}(\xx)$ satisfies $\xx(t)\in\overline{D}_R(\is{0})$ for sufficiently large $t$ and so $\Sigma_f\subset \overline{D}_R(0)$.
Since $(\real^m)^G \ne \emptyset$, there 
exists $\isk{c}\in\Sigma_f$ with isotropy $G$. Necessarily $\isk{c}\in(\real^m)^H$ for all $H\subset G$ and so 
if $\isk{c}$ is not a local minimum for $f|(\real^m)^H$, $f$ must have at least two critical points in $\overline{D}_R(0)^H$.  
Morse theory and other topological methods can often be used to prove the existence of
additional fixed points (see~\cite{Field1989} for examples and references).

In the Higgs-Landau theory from physics and equivariant bifurcation theory from dynamics, conjectures have been
made about the symmetry of critical points and equilibria in equivariant problems. Thus Michel~\cite{Michel1980} proposed
that symmetry breaking of global minima with isotropy $G$ in families of $G$-equivariant gradient polynomial vector fields 
would always be to minima of maximal isotropy type.
Similarly, in bifurcation theory, Golubitsky~\cite{Golubitsky1983} conjectured that for generic bifurcations, symmetry breaking
would be to branches of equilibria with maximal isotropy type. By maximal, we mean here that if the original branch of equilibria
had isotropy $H$ then the branch of equilibria generated by the bifurcation would have isotropy $H' \subsetneq H$, where $H'$ was
maximal amongst isotropy subgroups contained in $H$. While these conjectures turn out to be false, they 
have proved instructive in our understanding of symmetry breaking. We refer to~\cite{FieldRichardson1990} and \cite[Chapter 3]{Field2007}
for more details and references. Later we discuss symmetry breaking for the objective function defined using ReLU activation.

\section{ReLU and leaky ReLU neural nets}\label{sec: ReLU}
We describe symmetry and regularity properties of the loss and objective functions with ReLU  
activation: $\sigma(t) = [t]_+ = \max\{0,t\}$, $t \in \real$.  
Following the introduction, assume input variables $\xx \in \real^d$, $k$ neurons and associated parameters $\ww^1,\cdots,\ww^k$,
where each parameter is regarded as a $1\times d$ row matrix (element of $(\real^d)^\star$). Let $s \le k,d$. We assume target parameters $\cal{V}$ 
given by $s$ fixed \emph{non-zero} parameters $\vv^1.\cdots,\vv^s$  (functionals on $\real^s\subset \real^d$) and represented by the matrix $\VV^s \in M(s,s)$. 
Extend $\VV^s$ to $\VV \in M(k,d)$ by first appending $d-s$ zeros to each row of $\VV^s$ and then adding $k-s$ zero rows to obtain the matrix
\begin{equation}\label{eq: gentarget}
\VV = \left[\begin{matrix}
\VV^s & \is{0}_{s,d-s} \\
\is{0}_{k-s,s} & \is{0}_{k-s,d-s}
\end{matrix}\right].
\end{equation}
The non-zero rows of $\VV$ define the associated set $\cal{V}$ of parameters. Our choice of $\VV$ making the first $s$ rows non-zero is for convenience. Any
row permutation of $\VV$ leads to the same results.

The \emph{loss function} is defined by
\begin{equation}\label{eq: loss}
\loss(\WW,\VV) = \frac{1}{2}\mathbb{E}_{\xx \sim \cal{D}} \left(\sum_{i\in\is{k}}\sigma(\ww^i\xx) - \sum_{i\in\is{s}}\sigma(\vv^i\xx)\right)^2,
\end{equation}
where $\mathbb{E}$ denotes the expectation over an orthogonally invariant distribution $\cal{D}$ of 
initializations $\xx\in\real^d$. Generally, we take $\cal{D}$ to be the standard Gaussian distribution $\cal{N}_d(0,1)=\cal{N}(0,I_d)$. However,
any orthogonally invariant distribution $\cal{D}$ may be used provided that (a) the support $C_\cal{D}$ of the associated measure $\mu_{\cal{D}}$ has non-zero Lebesgue measure
and (b) $\mu_{\cal{D}}$ is equivalent to Lebesgue measure on $C_\cal{D}$.
If $\cal{D}=\cal{N}_d(0,1)$, then $\mu_{\cal{N}_d(0,1)}$ is equivalent to Lebesgue measure on $\real^d$; in particular, 
$\mu_\cal{D}(U) > 0$, for all non-empty open subsets $U$ of $\real^k$. We always assume conditions (a,b) hold if $\cal{D}$ is not the standard Gaussian distribution.

Set $\cal{F}(\WW) = \loss(\WW,\VV)$ and refer to $\cal{F}$ as the \emph{objective function}.
\subsection{Explicit representation of $\mathcal{F}$}\label{sec: objective}
We have
\begin{equation}\label{eq: obj}
\mathcal{F}(\WW) = \frac{1}{2}\sum_{i,j \in \is{k}} f(\ww^i,\ww^j) - \sum_{i\in \is{k},j \in \is{s}} f(\ww^i,\vv^j) +  \frac{1}{2}\sum_{i,j \in \is{s}}f(\vv^i,\vv^j),
\end{equation}
where $f(\ww,\vv)=\mathbb{E}_{\xx \sim \cal{N}_d(0,1)}\big(\sigma(\ww\xx)\sigma(\vv\xx)\big)$ and
\begin{enumerate} 
\item If $\vv,\ww \ne \is{0}$ and we set $\theta_{\ww,\vv} = \cos^{-1}\left(\frac{\ip{\ww}{\vv}}{\|\ww\|\|\vv\|}\right)$, then
\[ 
f(\ww,\vv)  = \frac{1}{2\pi} \|\ww\|\|\vv\|\big(\sin(\theta_{\ww,\vv}) + (\pi-\theta_{\ww,\vv})\cos(\theta_{\ww,\vv})\big)
\]
\item $f(\ww,\vv) = 0$ iff either $\vv=\is{0}$ or $\ww = \is{0}$ or $\theta_{\ww,\vv} = \pi$.
\end{enumerate}
See Cho \& Saul~\cite[\S 2]{ChooSaul2009},
and Proposition~\ref{prop:leaky-relu} below, for the proof.
\begin{rem}
Zero parameters ($\vv$ or $\ww$) do not contribute to $\cal{F}(\WW)$. \rend
\end{rem}

\subsection{Leaky ReLU nets}\label{sec: LeakyReLU}
Recall the leaky ReLU activation function is defined for $\alpha \in [0,1]$ by
$ \sigma_\alpha(t) = \max\{t, (1-\alpha)t\}\; t \in \real  $,
and that $\sigma_0(t) = t$, $\sigma_1(t) = \sigma(t)$, $t \in \real$ (choosing $\alpha$ rather than $\lambda$ is deliberate here).  The loss function corresponding to $\sigma_\alpha$ 
is defined by
\[
\mathcal{L}_\alpha(\WW,\VV) = 
\frac{1}{2}\mathbb{E}_{\xx \sim \mathcal{D}}\left(\sum_{i\in\is{k}}\sigma_\alpha(\ww^i\xx) - \sum_{i \in \is{s}}\sigma_\alpha(\vv^i\xx)\right)^2,
\]
where $\cal{D}$ is orthogonally invariant.
For $\alpha \in [0,1]$, define
\[
f_\alpha(\ww,\vv) = \mathbb{E}_{\xx \sim \mathcal{D}} \big(\sigma_\alpha(\ww\xx)\sigma_\alpha(\vv\xx)\big).
\]
The natural orthogonal action of $\On{d}$ on $\real^d$ induces an orthogonal action on $M(k,d)$ (matrix multiplication on the right)
and on parameter vectors via the action on the dual space $(\real^d)^\star$ (Examples~\ref{ex: dual}(3)). 
If $\ww \in (\real^d)^\star$ and $\xx \in \real^d$, then $(g\ww)\xx =  \ww g^{-1} \xx$ , for all $g \in \On{d}$.
\begin{lemma}\label{lem: poshom}
(Notation and assumptions as above.)
\begin{enumerate}
\item $f_1 = f$.
\item For all $\alpha\in [0,1]$, $f_\alpha$ is positively homogeneous
\begin{equation}
f_\alpha(\nu\ww,\mu\vv) =\nu\mu f_\alpha(\ww,\vv), \;\nu\mu \ge 0. 
\end{equation}
\item $f_\alpha$ is $\On{d}$-invariant
\[
f_\alpha(g\ww,g\vv) = f_\alpha(\ww,\vv),\; \ww,\vv \in \real^d, g \in \On{d}
\]
\end{enumerate}
\end{lemma}
\proof For (3), use $g\ww\xx = \ww (g^{-1} \xx)$ and the $\On{d}$-invariance of $\cal{D}$. \qed
\begin{prop}[{cf.~\cite[\S 2]{ChooSaul2009}}] \label{prop:leaky-relu}
If $\mathcal{D}$ is $\On{d}$-invariant, then
\[
f_\alpha(\ww,\vv) = \frac{c_\mathcal{D}\|\ww\|\|\vv\|}{2\pi}\left[\alpha^2(\sin(\theta) - \theta\cos(\theta)) + (2+ \alpha^2-2\alpha)\pi \cos(\theta)\right],
\]
where $c_\mathcal{D}$ is a constant depending on $\mathcal{D}$ and $\theta$ is the angle between $\ww, \vv$. If
$\mathcal{D}=\cal{N}_d(0,1)$, then $c_{\mathcal{D}} = 1$.
\end{prop}
\proof Step 1. Let $\alpha = 1$. By Lemma~\ref{lem: poshom}(2,3), 
we may assume $\|\ww\| = \|\vv\| = 1$, $\vv = (1,0,\ldots,0)$, $\ww = (\cos \theta,\sin \theta,0,\ldots,0)$, where $\theta \in [0,\pi]$
(if not, reflect $\ww$ in the $x_1$-axis). Thereby we reduce to a
2-dimensional problem. Denote the probability density on $\real^2$ by $p_{\mathcal{D}}$. We have
\begin{eqnarray*}
f(\ww,\vv) & = & \int_{\real^2} \sigma(\ww\xx) \sigma(\vv\xx) p_\mathcal{D}(\xx)\, d\xx\\
& =&\int_{\ww\xx, \vv\xx\ge 0} \ww\xx \times \vv\xx\, p_\mathcal{D}(\xx)\, d\xx \\
&=& \int_{x_1 \cos\theta + x_2 \sin \theta, x_1\ge 0} \hspace*{-0.25in}(x_1^2 \cos \theta + x_1 x_2 \sin \theta) p_\mathcal{D}(x_1,x_2)\, dx_1 dx_2
\end{eqnarray*}
Transforming the last integral using polar coordinates $x_1 = r\cos \phi, x_2 = r\sin \phi$ and writing $p_\mathcal{D}(x_1,x_2) = \frac{1}{2\pi}p(r)$, we have
\begin{eqnarray*}
f(\ww,\vv) & = & \left(\int_0^\infty r^3 p(r)\, dr\right)\left(\frac{1}{2\pi}\int_{\theta-\frac{\pi}{2}}^\frac{\pi}{2} \cos \theta \cos^2 \phi + \sin \theta \cos \phi \sin \phi \, d\phi\right) \\
& = & \left(\int_0^\infty r^3 p(r)\, dr\right) \left(\frac{1}{4\pi}\big((\pi-\theta) \cos(\theta) + \sin(\theta)\big)\right)
\end{eqnarray*}
If $\cal{D} = \cal{N}_d(0,1)$, then $p_\mathcal{D} = \frac{1}{2\pi} e^{-r^2/2}$ and so $\int_0^\infty r^3 p(r)\, dr = 2$.
Hence
\[
f(\ww,\vv) = \frac{\|\ww\|\|\vv\|}{2\pi} \big( \sin(\theta)+(\pi-\theta) \cos(\theta)\big),
\]
where $\theta = \theta_{\ww,\vv}$---the angle between $\ww$ and $\vv$.

Step 2. To complete the proof, use the identity $\sigma_\alpha(t) = \sigma(t) - \alpha\sigma(-t)$ in combination with the result of step 1.
This is a straightforward substitution and details are omitted. \qed

Write $\lambda = \frac{\alpha^2}{2+ \alpha^2-2\alpha}$ and observe that as $\alpha$ increases from $0$ to $1$, $\lambda$ increases from $0$ to $1$.
If $\cal{D} = \cal{N}_d(0,1)$, then 
\[
f_\alpha(\ww,\vv) = (2+ \alpha^2-2\alpha) \left[\frac{\lambda\|\ww\|\|\vv\|}{2\pi} \left(\sin(\theta_{\ww,\vv}) - \theta_{\ww,\vv}\cos(\theta_{\ww,\vv}) \right) + \frac{\langle \ww,\vv\rangle }{2}\right]
\]
Ignoring the factor $(2+ \alpha^2-2\alpha) \in [1,2]$, define
\[
f_\lambda(\ww,\vv) = \frac{\lambda\|\ww\|\|\vv\|}{2\pi} \left(\sin(\theta_{\ww,\vv}) - \theta_{\ww,\vv}\cos(\theta_{\ww,\vv}) \right) + \frac{\langle \ww,\vv\rangle }{2}, \; \lambda \in [0,1],
\]
and let $\{\mathcal{F}_\lambda\}_{\lambda \in [0,1]}$ denote the family of objective functions defined by
\begin{equation}\label{eq: grd_dl}
\mathcal{F}_\lambda(\WW) = \frac{1}{2}\sum_{i,j \in \is{k}} f_\lambda(\ww^i,\ww^j) - \sum_{i\in\is{k},j \in \is{s}} f_\lambda(\ww^i,\vv^j) +  \frac{1}{2}\sum_{i,j \in \is{s}}f_\lambda(\vv^i,\vv^j)
\end{equation}
Clearly, $\mathcal{F}_1 = \mathcal{F}$. When $\lambda  = 0$, $\mathcal{F}_0$ is the objective function of a trivial linear neural net with 
critical value set $\{0\}$.

\subsection{Symmetry properties of $\cal{L}_\lambda$ and $\cal{F}_\lambda$}\label{sec: symmsec}

By definition of the actions of $\On{d}$ on $M(k,d)$ and $(\real^d)^\star$,  it follows that
for all $i \in \is{k}$, $g \in \On{d}$, $(g\WW)^i = \ww^i g^{-1}$. 
Since $\cal{D}$ is assumed $\On{d}$-invariant, and $g\ww \xx = \ww g^{-1} \xx$, for all $\ww\in(\real^d)^\star$, $\xx\in\real^d$,  and $g \in \On{d}$, the function
$f(\ww,\vv)$ is $\On{d}$-invariant. Hence, by~\Refb{eq: grd_dl}, $\mathcal{L}_\lambda$ is $\On{d}$-invariant:
\begin{equation}
\label{eq: lossfninv}
\loss_\lambda(g\WW,g\VV) = \loss_\lambda(\WW,\VV), \; g \in \On{d}.
\end{equation}
\begin{lemma}\label{lem: cols}
\[
\loss(\rho\WW,\VV) = \mathcal{L}(\WW,\rho\VV) = \loss(\WW,\VV), \; \rho\in S^r_k\subset \Gamma.
\]
The same result holds for $\loss_\lambda$, $\lambda \in [0,1]$.
\end{lemma}
\proof Immediate since $\loss(\rho\WW,\VV),\, \loss(\WW,\rho\VV)$ are computed using the same terms as $\loss(\WW,\VV)$ but summed in a different order. \qed
\begin{prop}
The loss function $\loss$ is $S_k \times \On{d}$-invariant
\[
\loss(\gamma\WW,\gamma\VV) = \loss(\WW,\VV), \;\text{for all } \gamma = (\rho,g) \in S_k \times  \On{d}.
\]
The same result holds for $\loss_\lambda$, $\lambda \in [0,1]$.
\end{prop}
\proof Immediate from ~\Refb{eq: lossfninv} and Lemma~\ref{lem: cols}. \qed

Next we turn to invariance properties of $\mathcal{F}_\lambda$; these depend on $\VV^s$. 
Lemma~\ref{lem: cols} implies
that $\cal{F}_\lambda$  is $S^r_k$-invariant and so
\begin{equation}
\label{eq: objfninv1}
\cal{F}_\lambda(\rho\WW) = \cal{F}_\lambda(\WW), \; \rho \in S^r_k,\; \lambda \in [0,1].
\end{equation}
Suppose now that the rows of $\VV^s$ are linearly independent. 
If we let $\On{s}\subset \On{d}$ (resp.~$\On{d-s} \subset \On{d}$) act on the first $s$ (resp.~last $d-s$) columns of $M(k,d)$, then 
\[
g \VV = \VV,\; \text{for all } g \in \On{d-s}.
\]
On the other hand, since the rows of $\VV^s$ span $\real^s$, the only element of $\On{s}$ fixing $\VV$ is the identity $I_s$.
Define
\[
\Pi(\VV) = \{g \in \On{s} \dd \exists \pi(g) \in S^r_s \; \text{such that }g \VV = \pi(g) \VV\}  
\]
and note that $\Pi(\VV) \ne\emptyset$ since $I_s \in \Pi(\VV)$.
\begin{lemma}\label{lem: Pi}
(Notation and assumptions as above.)
$\Pi(\VV)$ is a finite subgroup of $\On{s}$ and the map $\pi: \Pi(\VV) \arr S_s \subset S^r_k; g \mapsto \pi(g)$ is 
well-defined and a group monomorphism. A necessary condition for $\Pi(\VV)$ to contain more than the identity element is that $\cal{V}$ consists of parameters with the same norm.
\end{lemma}
\proof Since the rows of $\VV^s$ are linearly independent, $\pi(g)$ is uniquely determined by $g$.
The remainder of the proof is routine. \qed
\begin{prop}\label{prop: loss_invariance}
(Notation and assumptions as above.) For $\lambda \in [0,1]$, $\mathcal{F}_\lambda$ is
$S_k \times (\Pi(\VV) \times \On{d-s})$-invariant
\end{prop}
\proof Follows by definition of $\Pi(\VV)$ and \Refb{eq: objfninv1}. \qed
\begin{exams}
(1) Suppose $\VV^s = I_s$. Then $\Pi(\VV) = S_s$, where $S_s \subset \On{s}$ acts by permuting columns, Since 
every column permutation of $I_s$ is induced by the same row permutation of $I_s$,
$\pi:  \Pi(\VV) \arr S_s \subset S^r_k$ is an isomorphism onto $S_s$.\\
(2) If $s=k \le d$ and $\VV^k = I_k$, then $\mathcal{F}_\lambda$ is $S_k \times (S_k \times \On{d-k})$-invariant.
In particular, if $d = k$, $\mathcal{F}_\lambda$ is $\Gamma$-invariant. If $d > k$, $\cal{F}_\lambda$ can be expected to have
$\On{d-k}$-orbits of critical points (note $\On{1} \approx \intg_2$). \\
(3) If $s = d < k$  and $\VV^s = I_d$, then $\cal{F}_\lambda$ is $S_k \times S_d$-invariant and there are no
continuous group symmetries. 
\end{exams}

\subsection{Differentiability and the gradient of $\mathcal{F}_\lambda$}\label{sec: diffF}
It follows from Section~\ref{sec: objective} that $\cal{F}_\lambda:M(k,d)\arr\real$ is continuous for all
$\lambda \in [0,1]$ (the maps $f_\lambda(\ww,\vv)$ are
obviously continuous, independently of the choice of $\VV$).

Let $s \le d,k$ and $\VV$ be the extension of $\VV^s$ to $M(k,d)$. 
Regard $f_\lambda(\ww,\vv)$ as a function of $\ww$ and set $f_1 = f$. Brutzkus \& Globerson~\cite[Supp.~mat.~A]{BrutzkusGloberson2017}  show
that $f(\ww,\vv)$ is $C^1$ provided that $\ww \ne \is{0}$ and give a formula for
the gradient of $f(\ww,\vv)$. Their result applies to $f_\lambda$ and gives
\begin{equation}
\label{eq: gradrelu}
\grad{f_\lambda}(\ww) = \frac{\lambda}{2\pi}\left(\frac{\|\vv\|\sin(\theta_{\ww,\vv})}{\|\ww\|}\ww  - \theta_{\ww,\vv}\vv\right) + \frac{\vv}{2}.
\end{equation}

Define subsets $\Omega_2,\Omega_v, \Omega_w, \Omega_a$ of $M(k,d)$ by
\begin{eqnarray*}
\Omega_2& = & \{\WW \dd \ww^i \ne \is{0},\; i \in \is{k}\} \\
\Omega_v&=&\{\WW \dd \ip{\ww^i}{\vv^j} \ne \pm \|\ww^i\|\|\vv^j\|,\; i \in \is{k},\;j\in \is{s}\}\\
\Omega_w&=&\{\WW \dd \ip{\ww^i}{\ww^j} \ne \pm \|\ww^i\|\|\ww^j\|,\; i,j \in \is{k}, i \ne j\}\\
\Omega_a&=&\Omega_v \cap \Omega_w
\end{eqnarray*}
\begin{lemma}\label{lem: reg}
\begin{enumerate}\item[]
\item $\Omega_2, \Omega_v, \Omega_w, \Omega_a$ are open and dense subsets of $M(k,d)$ and
\[ 
\Omega_a \subsetneq \Omega_v, \Omega_w \subsetneq \Omega_2
\]
\item $\cal{F}_\lambda$ is real analytic, as a function of $(\WW,\lambda)$, on $\Omega_a \times [0,1]$. 
\item For all
$\lambda \in (0,1]$, $\cal{F}_\lambda$ is $C^2$ on $\Omega_2$.  
\end{enumerate}
\end{lemma}
\proof We give the proof in the case of most interest here: $d \ge k$ and $\VV = I_k$ (the general case is similar). 
(1) $M(k,d) \smallsetminus \Omega_2$, $M(k,d) \smallsetminus \Omega_v$ are both finite unions of hyperplanes, each of codimension $d$. 
On the other hand, $M(k,d) \smallsetminus \Omega_w$ is a finite union of quartic hypersurfaces, each of codimension 1.  
Hence $\Omega_2, \Omega_v, \Omega_w, \Omega_a$ are  open and dense subsets of $M(k,d)$. It is easy to see that the inclusions are strict.\\
(2) This is immediate from~\Refb{eq: gradrelu}, the definition of $\Omega_a$, and the real analyticity of $\theta_{\ww,\vv}$ away from $\theta_{\ww,\vv} = 0,\pi$. \\
(3) Assume $\lambda = 1$ (the proof for $\lambda\in(0,1)$ is similar). The result of  Brutzkus \& Globerson cited above 
implies that $\cal{F}$ is $C^1$ on $\Omega_2$.
To show $\cal{F}$ is $C^2$ on $\Omega_2$, we use the Hessian computations of~\cite[\S 4.3.1]{SafranShamir2018}.  
Although $\theta_{\ww,\vv}$ is not differentiable if $\theta_{\ww,\vv}\in \{0,\pi\}$, 
$\grad{f}$ is $C^1$ at points $(\ww,\vv)$ where $\ww,\vv$ are parallel and non-zero: 
the contributions from the derivatives of 
$\sin(\theta_{\ww,\vv})$ and $-\theta_{\ww,\vv}$ cancel in the limit when $\ww$, $\vv$ are parallel. More formally, using the computations
of \cite[\S 4.3.1]{SafranShamir2018}, it is easy to
see that $\grad{f}$ is $C^1$ in $\ww$ at points where $\vv$ is parallel to $\ww$. 
The proof for the $\vv$-derivative is similar and 
based on~\cite{SafranShamir2018}. \qed

\begin{rems}
(1) Lemma~\ref{lem: reg} implies if $d \ge k$, then $\cal{F}$ is $C^2$ on a neighbourhood of the critical point $\VV$ giving the global minimum. If $d < k$,
$\cal{F}$ is not even $C^1$ at $\WW = \VV$ as $\VV \notin \Omega_2$ (see Remarks~\ref{rem: dist}(2)). \\
(2) $\cal{F}$ will typically not be $C^3$ at points in $\Omega_2\smallsetminus\Omega_a$. \\
(3) See~\cite{ArjevaniField2020b} for a formula for the Hessian when $d=k$, $\VV = I_k$. \rend
\end{rems}
For the remainder of the section assume $s = k \le d$ and $\VV^s = I_k$.

Set $\grad{\cal{F}_\lambda} = \Phi_\lambda: \Omega_2\arr M(k,d)$ and $\Sigma_\lambda = \{\WW\dd\Phi_\lambda(\WW) = \is{0}\}$. Recall $(\WW - \VV)^\Sigma$ is
the row sum $\sum_{j\in\is{k}} (\ww^j  - \vv^j)$ (Section~\ref{sec: prelim}). 
\begin{prop}\label{prop: weakrelu}
If $\WW \in \Omega_2$, then
$ \Phi_\lambda(\WW) = \is{G}_\lambda \in M(k,d)$, where 
$\is{G}_\lambda$ has rows $\is{g}_{\lambda}^1,\cdots,\is{g}_{\lambda}^k$ and, for $i\in\is{k}$,
\begin{align*}
\is{g}_{\lambda}^i =&\frac{\lambda}{2\pi}\sum_{j\in \is{k}} \left(\frac{\|\ww^j\|\sin(\theta_{\ww^i,\ww^j})}{\|\ww^i\|}\ww^i -\theta_{\ww^i,\ww^j}\ww^j\right)-  \\
&\frac{\lambda}{2\pi}\sum_{j\in \is{k}}\left(\frac{\sin(\theta_{\ww^i,\vv^j})}{\|\ww^i\|}\ww^i - \theta_{\ww^i,\vv^j}\vv^j\right) + \frac{1}{2}\big(\WW - \VV\big)^\Sigma.
\end{align*}
$\Phi_\lambda$ is real analytic on $\Omega_a \times [0,1]$.
\end{prop}
\proof Follows from Lemma~\ref{lem: reg} and \Refb{eq: gradrelu}. \qed

Recall that $\is{C} = \{\WW \in M(k,d) \dd \WW^\Sigma = \is{0}_{1,d}\}$ (Example~\ref{exam: isotypic}).
As an immediate (and trivial) consequence of Proposition~\ref{prop: weakrelu} we have the following result characterizing the
critical point set $\Sigma_0$ of $\mathcal{F}_0$. 
\begin{lemma}\label{lem: weakrelu}
(Notation as above.)
$\Phi_0(\WW) = \is{0}$ iff $\WW=\VV + \is{Z}$, for some  $\is{Z} \in \is{C}$. In particular, if $d = k$, 
$\WW\in\Sigma_0$ iff $\WW^\Sigma = \is{1}_{1,k}$. 
\end{lemma}
\proof It follows from Example~\ref{exam: isotypic} that $\is{Z} \in \is{C}$ iff $\is{Z}^\Sigma = \is{0}$. \qed

\subsection{Critical points and minima of $\mathcal{F}$}\label{sec: cp_comm}
We assume $\VV\in M(k,d)$ is the extension of $\VV^k = I_k$ to $M(k,d)$. 
For the moment assume $\lambda = 1$ and set $\cal{F}_1 = \cal{F}$.
If $d \ge k$, $\mathcal{F}$ has the minimum value of zero which is attained iff $\WW = \sigma\VV$ for some
$\sigma \in S_k \times S_k$. The `if' statement follows by $S_k \times (S_k \times \On{d-k})$-invariance and 
verification that $\mathcal{F}(\VV) = 0$. 
The proof of the converse is less trivial and deferred to the end of the section. 
Note that if $d \ge  k$, and $\sigma \in \Gamma_{k,d}$, then 
$\mathcal{F}$ is $C^2$ at $\WW = \sigma \VV$ since the rows are non-zero.

If $d = k$, $\cal{F}$ is $\Gamma_{k,k} = \Gamma$-invariant, the isotropy subgroup $\Gamma_\VV$ of $\VV$ 
is the diagonal subgroup $\Delta S_k\subset S_k \times S_k$ 
and $\cal{F}$ takes the minimum value of zero 
at any point of $\GG\VV$. From the perspective of symmetry breaking and bifurcation theory, one might expect bifurcation of
the global minima $\VV\in\Sigma_1$, as $k$ is increased, to \emph{spurious} minima---local minima which are not global minima---and that the spurious minima
should have isotropy which is conjugate to a \emph{proper} subgroup of $\Delta S_k$. However, this does not happen: $\VV$ is a global
minimum for all $k$ and the eigenvalues of the Hessian of $\cal{F}$ at $\VV$ are always strictly positive (see~\cite{ArjevaniField2020b} for
explicit computation of the eigenvalues). Moreover, for $k \ge 6$ there are spurious minima which have isotropy $\Delta S_k$ (we
refer to this class of minima as being of \emph{type A}). All the spurious minima for this problem that we are aware of have isotropy conjugate to a subgroup of
$\Delta S_k$.

Of special interest is the phenomenon that 
as we increase $k$, with $d = k$, we see increasing numbers of spurious minima 
of different isotropy type but always conjugate to a subgroup of
$\Delta S_k$~\cite{SafranShamir2018,ArjevaniField2020b}. 
The mechanisms underlying this behaviour can be explained using ideas from dynamical systems and equivariant bifurcation theory for the symmetric group $S_k$ and,
using results from this paper and~\cite{ArjevaniField2020b}, we address this in \cite{ArjevaniField2020c}. 
A feature of the analysis is that we regard $k$ as the bifurcation parameter making use of results in Section~\ref{sec: asym} where we obtain 
power series solutions in $1/\sqrt{k}$ for families of critical points of $\cal{F}$. 

Another approach to the analysis of the critical point structure of the objective function is to use a path-based approach and consider the family 
$\{\cal{F}_\lambda \dd \lambda \in [0,1]\}$ of $\Gamma$-invariant objective functions (assuming $k = d$ for simplicity). 
This family is highly singular at $\lambda =0$---$\Sigma_0=\is{C}$ is a codimension $k$ affine hyperplane of 
$M(k,k)$ (Lemma~\ref{lem: weakrelu}).  One standard approach to this type of problem is to attempt a desingularization of the family at 
$\lambda = 0$. That is, via a blowing-up procedure, define a modified family $\{\widetilde{\cal{F}}_\lambda\}_{\lambda \in [0,1]}$ on a new space 
where the critical point structure of $\cal{F}$ can be inferred from that of $\widetilde{\cal{F}}_0$. Although we do not know
how to implement such a desingularization, it turns out that specific classes of points in $\Sigma_0$ are naturally connected to critical points
of $\cal{F}$ via continuous paths in $\lambda$ of critical points for $\cal{F}_\lambda$. Indeed, this mechanism leads to methods for
obtaining good estimates for critical points of $\cal{F}$ and is suggestive of a deeper underlying structure (see 
Section~\ref{main-section} for more on this approach).

Although we conjecture that for our choice of $\VV$, the isotropy 
of spurious minima is conjugate to a subgroup of $\Delta S_k$, we emphasize that the isotropy of 
a general critical point of $\cal{F}$ is not always conjugate to a subgroup of $\Delta S_k$ and we give an example to illustrate this.  

\begin{exam}\label{ex: non-diag}
Set $\Phi | M(k,k)^\Gamma = \Psi^k$. Since $M(k,k)^\Gamma = \is{I}_{k,k}$, 
\begin{figure}[h]
\centering
\includegraphics[width=0.95\textwidth]{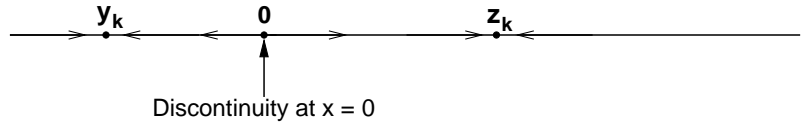}
\caption{Gradient descent: $x' = -\Psi^k(x)$ showing critical points $\is{y}_k,\is{z}_k$ as sinks and $-\Psi^k$ for $k\ge 2$.}
\label{fig: bif}
\end{figure}
we may regard $\Psi^k$ as defined on
$\real$ ($x\in\real$ is identified with $\is{x} \defoo x\is{1}_{k,k}$). Using the results of Section~\ref{sec: objective},  we find that
\[
\frac{2}{k^2}\Psi^k(x) = \begin{cases} 
&kx - 1 - \frac{1}{\pi}\left( \sqrt{k-1} - \cos^{-1}(\frac{1}{\sqrt{k}})\right),\; x > 0\\
&kx  + \frac{1}{\pi} \left(\sqrt{k-1} - \cos^{-1}(\frac{1}{\sqrt{k}})\right),\; x < 0
\end{cases}
\]
For $k \ge 1$, $\lim_{x\arr 0+}\Psi^k(x) < 0$, and $\lim_{x\arr 0-}\Psi^k(x) \ge 0$ (with equality only if $k=1$). For sufficiently large 
$|x|$, $\Psi^k(x) > 0$, if $x > 0$, and $\Psi^k(x) < 0$, if $x < 0$. It follows easily that for $k \ge 2$, there exist unique zeros $y_k < 0 < z_k$
for $\Psi^k$ (and so critical points $\is{z}_k,\is{y}_k$ of $\Phi$). We show the vector field for 
gradient descent  $x' = -\Psi^k(x)$ in Figure~\ref{fig: bif} and note that $\is{y}_k, \is{z}_k$
are critical points of $\cal{F}$ which define local minima of $\cal{F}| M(k,k)^\Gamma$ (though not of $\cal{F}$). We have
$\Gamma_{\is{y}_k} = \Gamma_{\is{z}_k} = \Gamma \not\subset \Delta S_k$. 
If instead we look for critical points with isotropy $\{e\} \times S_k$, it may be shown that
there exist $k-1$-dimensional linear \emph{simplices} of critical points with this isotropy for $\cal{F}$. Of course, this degeneracy results from working 
outside of the region $\Omega_a$ of real analyticity---in spite of $\cal{F}$ being $C^2$. Finally, if we take
\begin{align*}
x_k =&\frac{1}{(k-1)\pi}\left(\sqrt{k-1}+\pi - \cos^{-1}(\frac{1}{\sqrt{k}})\right) \\
y_k =&\frac{1}{\pi}\left(\sqrt{k-1}  - \cos^{-1}(\frac{1}{\sqrt{k}}\right)
\end{align*}
then $(-y_k\is{1}_{1,k}, x_k\is{1}_{1,k},\cdots, x_k\is{1}_{1,k})$ is a critical point with isotropy
$(S_{k-1} \times \{e\}) \times S_{k}$. Here the rows are parallel but the first row points in the reverse direction to the 
remaining rows:  all of these critical points lie in $\Omega_1 \smallsetminus \Omega_2$ and do not define not local minima of $\cal{F}$.
\end{exam}

\begin{prop}\label{prop: globalmin}
Let $s = k\le d$, $\VV\in M(k,d)$ be given by the standard orthonormal basis of $\real^k$, and regard $(S_k \times S_k)$ as a subgroup of $S_k \times (S_k \times \On{d-k})$ if $d > k$.
Assume $\cal{D}$ is $\On{d}$-invariant with $\mu_{\cal{D}}$ equivalent to Lebesgue measure on $\real^d$.
The objective function $\cal{F}(\WW)$ attains its global minimum value of zero iff $\WW \in (S_k \times S_k) \VV$.
The same result holds for the leaky objective function $\cal{F}_\lambda$, $\lambda \in (0,1)$.
\end{prop}
\proof As previously indicated, the proof that $\cal{F}(\WW) = 0$ if $\WW \in (S_k \times S_d) \VV$ is easy
since $\cal{F}(\VV) = 0$ and $\cal{F}$ is $\Gamma$-invariant. 

We now prove the converse for $\cal{F}$ leaving details for the leaky case to the reader. There are two main ingredients: the $\GG$-invariance of
$\cal{F}$ and the requirement that the $\On{d}$-invariant measure $\mu_{\cal{D}}$ associated to $\cal{D}$ is strictly positive on
non-empty open subsets of $\real^d$.  

Using the condition on $\mu_{\cal{D}}$ and the continuity of $\sigma$ and the matrix product, it follows from the defining equation~\Refb{eq: loss} that if there exists $\xx \in \real^k$ such that
\[
\sum_{i\in\is{k}} \sigma(\ww^i\xx) - \sum_{i\in\is{k}}\sigma(\vv^i\xx) \ne 0,
\]
then $\loss(\WW,\VV) > 0$ and $\WW$ cannot define a global minimum. Hence a necessary and sufficient condition for $\WW$ to define a global minimum is 
\begin{equation}\label{eq:globmin}
\sum_{i\in\is{k}} \sigma(\ww^i\xx) = \sum_{i\in\is{k}}\sigma(\vv^i\xx),\;\; \text{for all } \xx \in \real^d.
\end{equation}
Since $\sigma(\ww^i(-\xx)) = -\ww^i\xx$ if $\ww^i\xx < 0$, \Refb{eq:globmin} implies that for all $\xx\in \real^d$
\begin{eqnarray}\label{eq:globmin2+}
\sum_{i\,| \ww^i\xx > 0} \ww^i\xx&=&\sum_{i\,| \vv^i\xx > 0}\vv^i\xx, \\ 
\label{eq:globmin2-}
\sum_{i\,|  \ww^i\xx < 0} \ww^i\xx&=&\sum_{i\,| \vv^i\xx < 0}\vv^i\xx. 
\end{eqnarray}
and, in particular, that
\begin{equation}\label{eq:globmin3}
\sum_{i\in\is{k}} \ww^i\xx = \sum_{i\in\is{k}}\vv^i\xx,\;\; \text{for all } \xx \in \real^d.
\end{equation}
Taking $\xx = \vv_j$ in \Refb{eq:globmin3}, $j \in \is{k}$, it follows that $\cal{F}(\WW)=0$ only if (a) the column sums $\sum_{i\in\is{k}}w_{ij} = 1$, $j \in \is{k}$ (cf.~Lemma~\ref{lem: weakrelu}), and
(b) $w_{ij} = 0$, if $j > k$, $i\in \is{k}$. It follows from (b) that it is no loss of generality to assume $d = k$.
Taking $\xx = \vv_j$ in~(\ref{eq:globmin2+},\ref{eq:globmin2-}) implies that $w_{ij} \in [0,1]$, all $i,j \in \is{k}$. 
Hence, a necessary condition for $\cal{F}(\WW)=0$ is
\begin{equation}\label{eq:globmin4}
\WW^\Sigma = \is{1}_{1,k}.
\end{equation}
The proof now proceeds by induction on $k \ge 2$ (the case $k = 1$ is trivial). Suppose then that $\WW\in M(2,2)$ and $\cal{F}(\WW) = 0$. 
By~\Refb{eq:globmin4}, there exist
$\alpha_1, \alpha_2 \in (0,1]$ such that, after a permutation of rows and columns, 
\[
\WW = \left[\begin{matrix}
        \alpha_1 & 1-\alpha_2 \\                                   
        1 - \alpha_1 & \alpha_2                                   
        \end{matrix}\right] 
\]
Taking $\xx = \vv_1 - \mu\vv_2$, and substituting in~\Refb{eq:globmin2+}, gives
\[
[\alpha_1 - \mu (1-\alpha_2)]_+ + [(1-\alpha_1) - \mu \alpha_2]_+ = 1,\;\text{for all } \mu \ge 0.
\]
Noting that $\alpha_1, \alpha_2 > 0$, the only way this can hold is if $\alpha_1 = \alpha_2 = 1$, proving the case $k=2$.
Assuming the result has been proved for $2,\cdots,k-1$, it remains to show that the result holds for $k$. For this, start
by permuting rows and columns so that $w_{11} > 0$. Then, taking $\xx = \vv_1 - \mu \vv_j$, $j > 1$, follow the same recipe used for the case $k = 2$,
to show that $w_{11} = 1$ and $w_{1j} = 0$, $j > 0$. Since $w_{ij} = 0$, $j > 1$ by~\Refb{eq:globmin4}, this allows reduction to the 
matrix $\WW' \in M(k-1,k-1)$ defined by deleting the first row and column of $\WW$. Now apply the inductive hypothesis. \qed

\begin{rems}\label{rem: dist}
(1) The proof of Proposition~\ref{prop: globalmin} is simple because it does not use the analytic formula for $\cal{F}$. 
Versions of the proposition hold for truncated $\On{d}$-invariant distributions and other distributions 
provided that they are $\Gamma$-invariant and invariant under $-I_k$. For example,
the $k$-fold product of the uniform distribution on $[-1,1]$ (cf.~\cite{ArjevaniField2019c}). \\
(2) The case $k > d$ is harder. Setting $n = k-d$, it may be proved, along similar lines to Proposition~\ref{prop: globalmin}, that there is
a connected $n$-dimensional $\Gamma$-invariant simplicial complex $\Pi\subset M(k,d)$ such that (a) $\VV \in \Pi$, and (b) $\cal{F}(\WW) = 0$ iff
$\WW\in\Pi$. \rend
\end{rems}

\begin{cor}
(Assumptions and notation of Proposition~\ref{prop: globalmin}.)
The objective function $\cal{F}:M(k,k) \arr \real$ is a proper map. In particular, the level sets $\cal{F}^{-1}(c)$ are compact subsets of
$M(k,k)$ for all $ c \ge 0$.
\end{cor}
\proof Assume $d = k$ (the proof for $d > k$ is similar).
If $R > \|\VV\|$, then $\Gamma \VV \subset D_R(\is{0}) \subset M(k,k)$. By Proposition~\ref{prop: globalmin}, $\cal{F}(\WW) > 0$, if $\WW \in M(k,k)\smallsetminus D_R(\is{0}) $ and
so $\inf_{\WW \in  \partial D_R(\is{0})} \cal{F}(\WW) = C > 0$.
Write $\cal{F} = \sum_{i\in \is{3}} F_i$, where
\[
F_1(\WW) = \frac{1}{2}\sum_{i,j\in\is{k}}f(\ww^i,\ww^j),\;F_2(\WW) = \sum_{i,j\in\is{k}}f(\ww^i,\vv^j),\; F_3=\frac{1}{2}\sum_{i,j\in\is{k}}f(\vv^i,\vv^j)
\]
We have
\begin{enumerate}
\item $F_1(\nu\WW) = \nu^2 F_1(\WW)$, $F_2(\nu\WW) = \nu F_2(\WW)$, for all $\nu \ge 0$, $\WW \in M(k,k)$ (Lemma~\ref{lem: poshom}).
\item $F_1(\WW), F_2(\WW) > 0$, for all $\WW \ne \is{0}$ (statement (2) of Section~\ref{sec: objective}).
\end{enumerate}
It follows from (2) that there exist $\nu_0 \ge 1$, $D > 0$ such that $(\nu F_1 - F_2)(\WW) \ge \nu D$, for all $\nu \ge \nu_0$ and $\WW \in  \partial D_R(\is{0})$.
Using (1), we see easily that there exists $\alpha > 0$ such that $\cal{F}(\WW) \ge \alpha \|\WW\|$ for all $\WW \in M(k,k)\smallsetminus D_R(\is{0}) $.
Hence $\cal{F}$ is a proper map. \qed

\section{Isotropy and invariant space structure of $M(k,k)$}\label{sec: isot_struct}
We assume $d = k$ and consider the isotropy types that occur for the representation $(M(k,k),\GG)$, where $\GG=S_k \times S_k$. 
Results extend easily to $M(k,d)$, $d > k$. 
In line with previous comments on symmetry breaking, we focus 
on isotropy conjugate to a subgroup of $\Gamma_{\VV} = \Delta S_k$ 
rather than on general isotropy groups for the $\Gamma$-action. 
We provide few details of proofs which are all elementary.
\subsection{Isotropy conjugate to a product $H_r \times H_c \subset S_k \times S_k$} 
\begin{lemma}\label{lem: isostruct}
Let $\WW \in M(k,k)$ and suppose that $\Gamma_\WW = H^r \times H^c \subset S_k \times S_k$. Then
$\GG_\WW$ is conjugate to $(\prod_{\ell\in\is{p}} S_{r_\ell}) \times (\prod_{\ell\in\is{q}} S_{s_\ell})$,
where $r_\ell, s_\ell \ge 1$ and $k \ge p, q > 1$.
\end{lemma}
\proof Since $H^r$ acts on $\is{k}$, we may partition $\is{k}$ into $H^r$-orbits: 
$\is{k} = \{X_\ell \dd \ell \in \is{p}\}$, where $p \in \is{k}$ is the number of parts. For $\ell \in \is{p}$, let
$r_\ell\ge 1$ denote the cardinality of $X_\ell$.  It is no loss of generality to assume that
$X_1 = \{1,\cdots, r_1\},\cdots, X_p = \{r_{p-1}+1,\cdots, r_p\}$ since the relabelling gives a subgroup conjugate to $H^r$.  
Since $H^r$ acts transitively on each part $X^\ell$ and $H^r$ is an isotropy group we have $H^r = \prod_{\ell\in\is{p}} S_{r_\ell} \subset S_k$.
The argument is the same for $H^c$. \qed

Recall from Section~\ref{exam: isotypic} that $(M(k,k),\GG)$ has isotypic decomposition $\is{I} \oplus \is{C}_1 \oplus \is{R}_1 \oplus \is{A}$,
where $\is{R}_1$ (resp.~$\is{C}_1$) is the subspace of $M(k,k)$ consisting of matrices with identical rows (resp.~columns) and all row (resp.~column) sums 
equal to zero, $\is{A}$ is the space of matrices will all row and column sums equal to zero, and $\is{I}=\is{I}_{k,k}\subset M(k,k)$.

If $\WW \in \is{R}_1$ (resp.~$\is{C}_1$), then $\Gamma_\WW \supset S^r$ (resp.~$S^c$). In general,
if $\WW = \WW_I\oplus \WW_C\oplus \WW_R\oplus\WW_A \in \is{I} \oplus \is{C}_1 \oplus \is{R}_1 \oplus \is{A}$, then 
\begin{equation}
\GG_\WW = \GG_{\WW_I} \cap \GG_{\WW_C} \cap \GG_{\WW_R}\cap \GG_{\WW_A} =  \GG_{\WW_C} \cap \GG_{\WW_R}\cap \GG_{\WW_A},
\end{equation}
where the second equality follows since $\GG_{\WW_I} = \GG$. 
\begin{prop}\label{lem: CRrep}
Let $ \WW \in \is{I} \oplus \is{C}_1 \oplus \is{R}_1$. 
\begin{enumerate}
\item $\GG_\WW=\GG$ iff $\WW \in \is{I}$.
\item If $\WW \in \is{C}_1\smallsetminus \{\is{0}\}$, $\GG_\WW$ is conjugate to $(\prod_{\ell\in\is{p}} S_{r_\ell}) \times S_k$, where
$\sum_{\ell\in\is{p}} r_\ell = k$, $r_\ell \ge 1$ and $k \ge p > 1$ (if $p = k$, then $\GG_\WW = S^c_k$). 
\item If $\WW \in \is{R}_1\smallsetminus \{\is{0}\}$, $\GG_\WW$ is conjugate to $S_k \times (\prod_{\ell\in\is{q}} S_{s_\ell})$, where
$\sum_{\ell\in\is{q}} s_\ell = k$, $s_\ell \ge 1$ and $k \ge q > 1$ (if $q = k$, then $\GG_\WW = S^r_k$). 
\item If $\WW \notin \is{I}\oplus\is{C}_1 \cup \is{I}\oplus\is{R}_1$, then $\GG_\WW$ is conjugate to $(\prod_{\ell\in\is{p}} S_{r_\ell}) \times (\prod_{\ell\in\is{q}} S_{s_\ell})$
where $r_\ell, s_\ell \ge 1$ and $k \ge p, q > 1$.
\end{enumerate}
All the possibilities listed can occur for appropriate choices of $\WW \in \is{U}$.
\end{prop}
\proof Follows from Lemma~\ref{lem: isostruct}. \qed
\begin{rem}
If $\WW\in \is{A}$, then $\Gamma_\WW$ is conjugate to $H^r \times H^c$ iff $H^r = H^c$ (as subgroups of $S_k$). \rend
\end{rem}
\subsection{Isotropy of $\Gamma$-actions on $M(k,k)$}
Isotropy for the action of $\GG$ on $\is{A}$ is more complex than that given by Lemma~\ref{lem: isostruct}.
With a view to applications, we emphasize isotropy conjugate to a subgroup of $\Delta S_k$ rather than attempt 
a general classification. We start with a cautionary example.
\begin{exam}\label{ex: not_diag}
Suppose $k=4$, $a \ne b$, and
$
[\WW] = \left[\begin{matrix}
        a & b & b & a \\                                   
        a& a & b & b \\                                   
        b & a & a & b \\
        b & b & a & a                                    
        \end{matrix}\right] 
$.
Observe that $\Gamma_\WW$
contains the symmetries
$
\eta = ((1234)^r, (1234)^c)$, $\gamma = ((13)^r,(12)^c(34)^c)$
and $\eta^4 = \gamma^2 = (\eta\gamma)^2 = e$. It is well-known that these are the generating relations for $\mathbb{D}_4$---the dihedral group
of order $8$. Hence $|\Gamma_\WW| \ge 8$.  We leave it to the reader to verify $|\Gamma_\WW| = 8$, so that $\Gamma_\WW \approx \mathbb{D}_4$,
and $\Gamma_\WW$ is not a product of subgroups of $S_k$ or conjugate to a subgroup of $\Delta S_4$. 
\end{exam}
\subsubsection{Isotropy of diagonal type}
\begin{Def}
An isotropy group $J$ for the action of $\Gamma$ on $M(k,k)$ is of \emph{diagonal type} if there exists a subgroup $H$ of $S_k$ such that $J$ is conjugate to
$\Delta H = \{(h,h) \dd h \in H\}$. 
\end{Def}
\begin{lemma}
If $H$ is a transitive subgroup of $S_k$ and $\WW \in M(k,k)^{\Delta H}$ (so $\Gamma_\WW \supset \Delta H$), then the diagonal elements of $\WW$ are all equal.
Conversely, if 
the induced action of $\Gamma_\WW$ on $\is{k}^2$ has an orbit with $k$-elements meeting each row and column in $\is{k}^2$, then
$\Gamma_\WW $ is conjugate to $\Delta H$, where $H\subset S_k$ is transitive.
\end{lemma}
\proof 
The first statement follows since $H$ is transitive and so for $i \in \is{k}$,  there exists $\rho \in H$ such that $\rho(1) = i$. Hence
$(\rho,\rho)(1,1) = (i,i)$.  For the converse, note $\Gamma_\WW$ is conjugate to a subgroup $H$ of $S_k\times S_k$such that the $H(1,1) = \Delta \is{k}^2$.
If there exists $(g,h) \in H \smallsetminus \Delta S_k$, then the $H$-orbit of $(1,1)$ must contain more than $k$-elements. Hence $H \subset \Delta S_k$. \qed

\begin{rems}
(1) If $\WW\in M(k,k)$, then $\Gamma_\WW  = \Delta S_k$ iff there exist $a,b \in \real$, $a \ne b$, such that
$w_{ii} = a$, $i \in \is{k}$, and $w_{ij} = b$, $i,j \in \is{k}$, $i \ne j$. \\
(2) If $K$ a doubly transitive subgroup of $S_k$ then $\Delta K$ will be an isotropy group for the action of $\Gamma$ on $M(k,k)$
iff $K=S_k$. Indeed,
if $K \subsetneq S_k$ is a doubly transitive subgroup of $S_k$ (for example, the alternating
subgroup $A_k$ of $S_k$, $k > 3$), then the double transitivity
implies that if $\Gamma_\WW = \Delta K$ then all off-diagonal entries of $\WW$ are equal. Hence $\Gamma_\WW = \Delta S_k$ by (1). 
If $H$ is a subgroup of $S_k$ which does not act doubly transitively on any part of the transitivity partition of $H$, then
$\Delta H$ will be an isotropy group for the action of $\Delta H$ on $M(k,k)$. \rend
\end{rems}
The analysis of isotropy of diagonal type can largely be reduced to the study of the diagonal action of transitive subgroups of $S_p$, $2 \le p \le k$. 
We give two examples to illustrate the approach and then concentrate on describing maximal isotropy subgroups of $\Delta S_k$.
\begin{exams}
(1) Suppose $K_4\subset S_4$ is the Klein 4-group---the Abelian group of order $4$ generated by the involutions $(12)(34)$ and 
$(13)(24)$. Matrices with
isotropy $\Delta K_4$ are of the form
\begin{equation}\label{eq: blockform}
\WW = 
\left(\begin{matrix}
a & b & c & d \\
b & a & d & c \\
c & d & a & b \\
d & c & b & a 
\end{matrix} \right) \in \is{I}\oplus \is{A},
\end{equation}
where $a,b,c,d$ are distinct (else, the matrix has a bigger isotropy group). \\
(2) If $k = 8$ and $H = \Delta K_4 \times \Delta K_4 = \Delta (K_4 \times K_4)$, then matrices
with isotropy $H$ may be written in block matrix form as
$
\WW = 
\left(\begin{matrix}
A & B \\
C & D 
\end{matrix} \right), 
$
where $A,D$ have the structure given by~\Refb{eq: blockform}. 
Since $H$ is a product of groups of diagonal type, $B$ and $C$ are real
multiples of $\is{1}_{4,4}$ and so $\text{dim}(M(4,4)^{H}) = 10$.
We may vary this example to get $4$ copies of the basic block.
To this end, observe that if
$K \subset S_8$ is  generated by $(12)(34)(56)(78), (13)(24)(57)(68)$, then $K \approx K_4$. With
$H = \Delta K$, if $\GG_\WW = H$, then $\WW$ has the same
block decomposition as before but now every block has the structure given by~\Refb{eq: blockform} and 
$\text{dim}(M(4,4)^H) = 16$. Add the involution $(15)(26)(37)(48)$ to $K_4 \times K_4$  to generate $K'\subset S_8$. If
$\GG_\WW = \Delta K'$,  then $A=D$, $C = B = e \is{1}_{4,4}$, and 
$\text{dim}(M(4,4)^{\Delta H'}) = 5$.

\end{exams}

\subsection{Maximal isotropy subgroups of $\Delta S_k$}
Of special interest are maximal isotropy subgroups of $\Delta S_k = \Gamma_\VV$.
These subgroups are the diagonals of \emph{maximal
proper subgroups} of $S_k$, groups which have received attention from group theorists
because of connections with the classification of simple groups (see~\cite[Appendix 2]{AschbacherScott1985} for the O'Nan--Scott theorem which
describes the structure of maximal subgroups of $S_k$). 
Here we consider maximal subgroups of $S_k$ which are not transitive, and the class of imprimitive 
transitive subgroups of $S_k$ (for example, \cite[Prop. 2.1]{NewtonBenesh2006}). 
We do not discuss primitive transitive subgroups of $S_k$---see 
Liebeck \emph{et al.}~\cite{Liebecketal1987} and 
Dixon \& Mortimer~\cite[chap.~8]{DixonMortimer1996}. 

\begin{lemma}
\begin{enumerate}
\item If $p+q = k$, $p,q \ge 1$, $p\ne q$, then $S_p \times S_{q}$ is a maximal proper subgroup of $S_k$ (intransitive case).
\item If $k = pq$, $p, q > 1$, the wreath product $S_p \wr S_q$ is transitive and a maximal proper subgroup of $S_k$ with $|S_p \wr S_q| =(p!)^q q!$ 
\end{enumerate}
\end{lemma}
\proof (Sketch) (1) If $p = q = k/2$, we can add to $S_p\times S_p$ permutations which map $\is{p}$ to $\is{k} \smallsetminus \is{p}$ 
to obtain a larger proper subgroup of $S_k$. 
(2) The transitive partition breaks into $q$ blocks $(B_i)_{i\in \is{q}}$ each of size $p$. The wreath product~\cite[Chap.~7]{JJRotman} 
acts by permuting elements 
in each block and then permuting the blocks.  \qed
\begin{exams}\label{ex: sk-1-ex}
(1) Set $H = \Delta S_{k-1}$, $k \ge 3$. If  $\WW \in M(k,k)$ and $\GG_\WW$ is conjugate to $H$ then, 
after a permutation of rows and columns,
\[
\WW = 
\left(\begin{matrix}
a & b & b & \ldots & b & e\\
b & a & b & \ldots & b & e\\
\ldots & \ldots & \ldots & \ldots & \ldots & \ldots \\
b & b & b & \ldots & a & e \\
f & f & f & \ldots & f & g
\end{matrix} \right),
\]
where  $a,b,e,f,g \in \real$, $a \ne b$, and we do not have $a = g$ and $b=e=f$ (giving isotropy $\Delta S_k$).
Hence $\text{dim}(M(k,k)^H)=5$. Note that $\GG\VV \cap M(k,k)^H = \{\VV\}$.
If $H_p = \Delta S_{p} \times \Delta S_{k-p}$, $1 < p < k/2$, then
$\WW \in M(k,k)^{H_p}$ has block matrix structure $\left(\begin{matrix}
A & c\is{1}_{p,k-p} \\
d\is{1}_{k-p,p} & D 
\end{matrix} \right)$, where $A \in  M(p,p)^{\Delta S_{p}}$, $D \in M(k-p,k-p)^{\Delta S_{k-p}}$ and $c,d \in \real$.
It follows that $\text{dim}(M(k,k)^{H_p}) = 6$. Again we have $\GG\VV \cap M(k,k)^H = \{\VV\}$. \\
(2) If $k = pq$, $p,q > 1$, then $H = S_p \wr S_q$ is a maximal transitive subgroup of $S_k$ and so $\Delta H$ is a maximal subgroup of $\Delta S_k$.
If $\Gamma_\WW = \Delta H$, then we may write $\WW$ in block form as
\[
\WW = 
\left(\begin{matrix}
A & C & C & \cdots & C\\
C & A & C & \cdots & C\\
\cdots & \cdots & \cdots & \cdots &\cdots \\
C & C & C & \cdots & A
\end{matrix} \right) \in \is{I}\oplus \is{A},
\]
where 
$
A = \left(\begin{matrix} 
a & b & \cdots & b\\ 
b & a & \cdots & b\\ 
\cdots & \cdots & \cdots & \cdots\\ 
b & b & \cdots & a
\end{matrix} \right)$, $C = c \is{1}_{p,p}$, and $a,b,c \in \real$ with $a \ne b$.
We have $\text{dim}(M(k,k)^{\Delta H}) = 3$, independently of $k, p,q$. Unlike what happens in the previous example, 
$M(k,k)^{\Delta H}$ contains two points of $\GG \VV$ and matrices in $M(k,k)^{\Delta H}$ are all self-adjoint.
\end{exams}
\subsection{Parametrizing certain families of fixed point spaces}\label{ex: deltak-1ex}

The objective function $\mathcal{F}_\lambda: M(k,k) \arr \real$ is $\Gamma$-equivariant and so, by Lemma~\ref{lem: symm},  if
$H$ is any subgroup of $\Gamma$, then $\Sigma_\lambda^H = \Sigma_\lambda \cap M(k,k)^H$ is equal to the critical point set of $\mathcal{F}_\lambda | M(k,k)^H$. 
In order to find $\Sigma_\lambda^H$, it suffices to find the critical points of
$\mathcal{F}_\lambda | M(k,k)^H$. In order to do this we define a natural parametrization of $M(k,k)^H$.
We are especially interested in studying $k$-dependent families of fixed point spaces where the dimension of 
the fixed point space is \emph{independent} of $k$. We focus on the 
fixed point space of $M(k,k)$ defined by $\Delta (S_{k-p} \times S_p)$, where $0 \le p < k/2$.  
As in Examples~\ref{ex: sk-1-ex}(1), we find 
\[
\text{dim}(M(k,k)^{\Delta (S_{k-p}\times S_p)} = 2 + \min \{4,2p\}, \;\; k \ge p+2.
\]
Set $F(k-p,p)  =  M(k,k)^{\Delta (S_{k-p} \times S_p)}$. We start with the simplest case
when $p = 0$ and $\text{dim}(F(k,0)) =2$. For $k \ge 2$ define
the linear isomorphism  $\Xi: \real^2 = F(k,0)$  by
\begin{equation}
\label{eq:X0_def}
\Xi(\boldsymbol{\xi}) =
\left[\begin{matrix}
\xi_1 & \xi_2& \ldots & \xi_2 \\
\xi_2 & \xi_1& \ldots & \xi_2 \\
\ldots & \ldots&\ldots&\ldots \\
\xi_2 & \xi_2& \ldots & \xi_1 
\end{matrix}\right] \defoo 
A_{k,k}(\xi_1,\xi_2)
\end{equation}
The matrix $A_{k,k}(\xi_1,\xi_2)$ is defined for all $k \ge 2$ and lies in $M(k,k)^{\Delta S_{k}}$. We have $\Gamma_{A_{k,k}(\xi_1,\xi_2)} = \Delta S_{k}$ iff $\xi_1 \ne \xi_2$.
Now suppose $k/2 > p \ge 2$. Define $\Xi: \real^6 \arr F(k-p,p)$ by
\begin{equation}
\label{eq:X2_def}
\Xi(\boldsymbol{\xi}) =
\left[\begin{matrix}
A_{k-p,k-p}(\xi_1,\xi_2) & A_{k-p,p}(\xi_3)\\
A_{p,k-p}(\xi_4) & A_{p,p}(\xi_5,\xi_6)
\end{matrix}\right],\;
\end{equation}
where $A_{k-p,k-p}(\xi_1,\xi_2)$, $A_{p,p}(\xi_5,\xi_6)$ are as defined above and
\[
A_{k-p,p}(\xi_3) = \xi_3 \is{1}_{k-p,p},\quad A_{p,k-p}(\xi_4)=\xi_4 \is{1}_{p,k-p}.
\]
We have 
\[
A_{k-p,p}: \real^6\arr M(k-p,p)^{S_{k-p}\times S_p}, \quad A_{p,k-p}:\real^6\arr M(p,k-p)^{S_p \times S_{k-p}}.
\]
In case $p = 1$, $\Xi: \real^5\arr F(k-1,1)$ and
\begin{equation}
\label{eq:XI_def}
\Xi(\boldsymbol{\xi}) =
\left[\begin{matrix}
A_{k-1,k-1}(\xi_1,\xi_2) & A_{k-1,1}(\xi_3)\\
A_{1,k-1}(\xi_4) & A_{1,1}(\xi_5)
\end{matrix}\right].
\end{equation}
This parametrization is the same as that given in Examples~\ref{ex: sk-1-ex}(1).
\subsection{Critical point equations in the presence of symmetry}\label{sec: cp_speq}
We obtain symmetry optimized equations for critical points of $\cal{F}$ in $F(k-p,p) \cap \Omega_a$, $p=0,1$. The equations
are simple and used in Section~\ref{sec: asym}. 

Take $p=1$ and let $\WW \in F(k-1,1) \cap \Omega_a$. Set $\Delta (S_{k-1} \times S_1) = \Delta S_{k-1}$. Following the notational conventions of Proposition~\ref{prop: weakrelu},
set $\Theta = \theta_{\ww^1,\ww^2}$, and $\Lambda = \theta_{\ww^1,\ww^k}$. Since $\Delta S_{k-1} \subset\Gamma_\WW$,
we have
\begin{enumerate}
\item $ \theta_{\ww^i,\ww^j} = \Theta$, for all $i, j \le k-1$, $i \ne j$.
\item $\theta_{\ww^i,\ww^k} = \theta_{\ww^k,\ww^j} = \Lambda$, for all $i,j \le k-1$.
\end{enumerate}
(If $p = 0$, $\Lambda$ is not defined.) 
Define $\alpha_{ii} = \theta_{\ww^a,\vv^a}$, $\alpha_{ij} = \theta_{\ww^a,\vv^b}$, $\alpha_{ik} = \theta_{\ww^a,\vv^k}$,
$\alpha_{kk} = \theta_{\ww^k,\vv^k}$, and $\alpha_{k j} = \theta_{\ww^k,\vv^b}$
for $a,b \in \is{k-1}$, $a \ne b$.  Since $\Gamma_\WW\supset \Delta S_{k-1}$, the $\alpha$ angles are well defined.
If $p=0$, only $\alpha_{ii}$ and $\alpha_{ij}$ are defined and (later) we set $\alpha_{ij} = \alpha$, $\alpha_{ii} = \beta$.

Given $\bxi \in \real^5$, set $\widehat{\Xi}^\Sigma = \Xi(\bxi)^\Sigma-\mathbf{I}_{1,k}$.  We have $\widehat{\Xi}_1^\Sigma = \cdots = \widehat{\Xi}_{k-1}^\Sigma$. 
Define 
\begin{eqnarray*}
P&=&\sum_{j \in\is{k}} \left(\frac{\|\ww^j\|\sin (\theta_{\ww^i,\ww^j})}{\|\ww^i\|} - \frac{\sin(\theta_{\ww^i,\vv^j})}{\|\ww^i\|} \right),\; i < k. \\
Q&=&\sum_{j \in\is{k}} \left(\frac{\|\ww^j\|\sin (\theta_{\ww^k,\ww^j})}{\|\ww^k\|} - \frac{\sin(\theta_{\ww^k,\vv^j})}{\|\ww^k\|} \right)\\
\mathbf{A}&=&\Xi(\alpha_{ii},\alpha_{ij},\alpha_{ik},\alpha_{kj},\alpha_{kk}) \in M(k,k)^H \\
\mathbf{E}^i & = & (\pi - \Theta)\widehat{\Xi}^\Sigma + \Theta \ww^i +  (\Theta-\Lambda) \ww^k + \mathbf{A}^i - \Theta \is{1}_{1,k},\; i \in \is{k-1}\\
\mathbf{E}^k & = & (\pi - \Lambda)\widehat{\Xi}^\Sigma + \Lambda \ww^k + \mathbf{A}^k - \Lambda \is{1}_{1,k}.
\end{eqnarray*}
On account of the $\Delta S_{k-1}$-symmetry, $P$ does not depend on the choice of $i \in \is{k-1}$.  If $p = 0$, then $P = Q$.
\begin{prop}\label{prop: cp_eqns}
(Notation and assumptions as above.)
\begin{enumerate}
\item[(p=1)] Let $\bxi \in \real^5$. Then $\Xi(\bxi)=\WW \in \Sigma_1^{\Delta S_{k-1}}$ iff
\[
P\ww^i + \is{E}^i = Q \ww^k + \is{E}^k = \is{0},\;\text{for all } i < k.
\]
\item[(p=0)] Let $\bxi \in \real^2$. Then $\Xi(\bxi)=\WW \in \Sigma_1^{\Delta S_{k}}$ iff
\[
P\ww^i + \is{E}^i = 0\;\text{for all } i\in \is{k}.
\]
\end{enumerate}
\end{prop}
\proof Straightforward substitution. \qed
\subsection{Minimal set of critical point equations  for $p = 0,1$}\label{sec: cpequations}  \mbox{    }\\
Using $\Delta S_{k-1}$ (or $\Delta S_k$) symmetry, we derive a minimal set of equations determining the critical points in Proposition~\ref{prop: cp_eqns}.
\[
\text{\hspace*{-0.75in}$p=0$:\hspace*{0.75in}}\begin{cases}
&P \xi_1 + (\pi-\Theta)\WXi_1 + \Theta \xi_1 + \beta-\Theta = 0 \\
&P \xi_2 + (\pi-\Theta)\WXi_1 + \Theta \xi_2 + \alpha-\Theta = 0 
\end{cases}
\]
\[
\text{\hspace*{-0.2in}$p=1$:\hspace*{0.2in}}\begin{cases}
&P \xi_1 + (\pi-\Theta)\WXi_1 + \Theta \xi_1 +  (\Theta-\Lambda) \xi_4 + \alpha_{ii}-\Theta = 0 \\
&P \xi_2 + (\pi-\Theta)\WXi_1 + \Theta \xi_2 +  (\Theta-\Lambda) \xi_4+ \alpha_{ij}-\Theta = 0 \\
&P \xi_3 + (\pi-\Theta)\WXi_k + \Theta \xi_3 + (\Theta-\Lambda) \xi_5+ \alpha_{ik}-\Theta = 0 \\
&Q \xi_4 + (\pi-\Lambda)\WXi_1 + \Lambda \xi_4 + \alpha_{kj}-\Lambda = 0 \\
&Q \xi_5 + (\pi-\Lambda)\WXi_k + \Lambda \xi_5 +\alpha_{kk}  -\Lambda = 0
\end{cases}
\]

\subsection{A regularity constraint on critical points of $\cal{F}$}\label{sec: const}
Example~\ref{ex: non-diag} gives one case where the isotropy of a 
critical point $\isk{c}$ of $\cal{F}$ is not conjugate to a subgroup of $\Delta S_k$ and $\isk{c}\notin \Omega_a$. 
More generally, we have
\begin{prop}\label{prop: constrain}
If $\is{W} \in M(k,k)$ and $\Gamma_\WW$ contains a row permutation, then
$\is{W}\notin \Omega_a$.
\end{prop}
\proof The hypothesis implies $\WW$ has a pair of parallel rows. \qed
\begin{rem}
Proposition~\ref{prop: constrain} constrains the symmetry of critical points of $\cal{F}$ lying in $\Omega_a$ 
but says nothing about critical points with isotropy of the type described by Example~\ref{ex: not_diag} which is not conjugate to a subgroup
of $\Delta S_k$ or to a product subgroup $H \times K$. \rend
\end{rem}

\section{Results, Methods \& Conjectures}
\label{main-section}
\subsection{Introductory comments}
A primary aim of this paper is to obtain analytic results about the critical points of $\cal{F}$; in particular, the critical points of spurious minima.
While it is straightforward to find small sets of analytic equations for the critical points---at least if the critical points have
non-trivial isotropy---only exceptionally can one find explicit analytic solutions for these equations. 
However, it is often possible to find convergent power series in $1/\sqrt{k}$ for families of critical points and the initial terms of
these series can be computed. 
These series allow one to prove sharp results about the spectrum of the Hessian~\cite{ArjevaniField2020b} and the decay of spurious minima (Section~\ref{sec: asym}) as $k\arr\infty$.

We have two approaches to the construction of critical points and power series solutions: a direct approach 
and an indirect path based method using the family $\{\cal{F}_\lambda\}_{\lambda\in[0,1]}$. The direct method gives
exact power series solutions while the indirect method, discussed in the remainder of the section, assumes the column constraint $(\WW-\VV)^\Sigma = \is{1}_{1,d}$---an affine
linear condition on the components of the critical point---and gives a solution in $\Sigma_0$. 
We start with two examples where a complete description of critical points can be given and both methods apply.
\begin{exam}[Families of critical points for leaky ReLU nets]\label{ex: cpsexpl}
Let $\Phi_\lambda$ denote the gradient vector field of $\cal{F}_\lambda$ and $\Sigma_\lambda$ denote the set of critical points of $\cal{F}_\lambda$ 
($\Sigma_0$ is the codimension $k$ affine linear subspace of $M(k,k)$ defined by requiring that all columns sum to $1$). \\
(a) Substituting in the formula for $\Phi_\lambda$ (Proposition~\ref{prop: weakrelu}), we obtain the trivial family $\{\VV(\lambda)\}_{\lambda \in [0,1]}$ 
of critical points for $\cal{F}_\lambda$ defined by
\[
\VV(\lambda)  = \VV,\; \lambda \in [0,1].
\]
There is no non-trivial dependence on $\lambda$ but the solution curve uniquely determines the point $\VV\in\Sigma_0$. \\
(b) The critical points of $\Phi_1$ with maximal symmetry $\Gamma$ are described in Example~\ref{ex: non-diag}. 
In particular the critical point $\is{z}_k = z_k \is{1}_{k,k}$, where $z_k > 0$.  Using Proposition~\ref{prop: weakrelu}, 
the associated curve $\{\is{z}_k(\lambda)=z_k(\lambda)\is{1}_{k,k}\}_{\lambda \in [0,1]}$ of critical points for $\cal{F}_\lambda$ is given by
\[
z_k(\lambda) = \frac{1}{k} + \frac{\lambda}{\pi} \left[\sqrt{k-1} - \cos^{-1}\left(\frac{1}{\sqrt{k}}\right)\right],\; k \ge 1, \lambda \in [0,1].
\]
The dependence of $\is{z}_k(\lambda)$ on $\lambda$ is linear and $\is{z}_k(0) = \frac{1}{k} \is{1}_{k,k}\in \Sigma_0$.
Noting the Maclaurin series
\[
(1-x)^{\frac{1}{2}}=1 - \sum_{n=0}^\infty \frac{2^{-2n-1}}{n+1}\binom{2n}{n} x^{n+1},\;
\sin^{-1}(x) = \sum_{n=0}^\infty \frac{2^{-2n}}{2n+1}\binom{2n}{n} x^{2n+1},
\]
and the identity $\cos^{-1}(x) = \frac{\pi}{2} - \sin^{-1}(x)$, we obtain 
\begin{equation*}
z_k(\lambda) = \frac{\lambda}{\pi\sqrt{k}}+ (1-\frac{\lambda}{2})\frac{1}{k} + \frac{\lambda\sqrt{k}}{\pi}\left[\sum_{n=0}^\infty \frac{2^{-2n}}{(2n+1)(2n+2)} \binom{2n}{n}\frac{1}{k^{n+1}}\right]
\end{equation*}
Hence, for $\lambda > 0$, $|z_k| = 0(1/\sqrt{k})$ and $\is{z}_k(0)=k^{-1}\is{1}_{k,k}\in\Sigma_0$. Observe that
$z_k(1)\in\Sigma_1$ is a power series in $1/\sqrt{k}$, with initial term $\frac{3}{2\pi\sqrt{k}}$. Setting $s = 1/\sqrt{k}$, $z_k(\lambda)=z(s,\lambda)$ is a real analytic  function of 
$(s,\lambda)$
on $[0,1) \times [0,1]$ (that is, for $k > 1$, $\lambda \in [0,1]$). 
\end{exam}

In both examples we have explicit analytic expressions for critical points $\mathfrak{c} \notin \Omega_a$.
The simplicity of the examples is reflected in the geometry of $\VV$ and $\mathfrak{c}$ 
through the presence of parallel weights in $\mathfrak{c},\VV$.

The spurious minima found numerically in~\cite{SafranShamir2018} all lie in $\Omega_a$ and it is unlikely that simple
analytic expressions can be found for these minima. Indeed, no rows of $\mathfrak{c} \in \Omega_a$ are parallel to another row of $\mathfrak{c}$ or to a row of $\VV$
and there is no obvious way of using the geometry to find expressions for the critical points.  
However, the spurious minima described in~\cite{SafranShamir2018} all have isotropy conjugate to a subgroup of the diagonal group $\Delta S_k$. We focus on critical
points with isotropy of this type since the isotropy does not constrain the regularity of $\cal{F}$ (Proposition~\ref{prop: constrain}).

\begin{figure}[h]
\centering
\includegraphics[width=\textwidth]{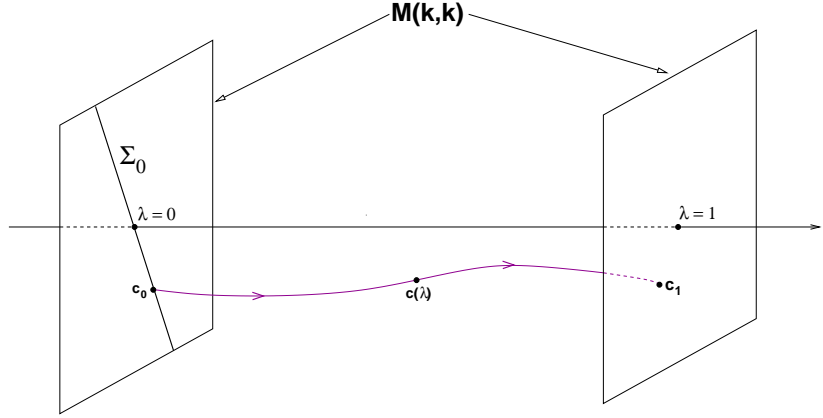}
\caption{Curve $\mathfrak{c}(\lambda)$ of critical points joining $\mathfrak{c}_0\in\Sigma_0$ to a critical point $\mathfrak{c}_1$ of $\cal{F}$.}
\label{fig: path}
\end{figure}

In the above examples we obtained curves joining critical points in $\Sigma_0$ and $\Sigma_1$ which were either constant or linear in $\lambda$
and appear to be of little interest. However, if instead we ask about critical points in $\Sigma_1$ with isotropy conjugate to a subgroup of $\Delta S_k$, there are some surprises.
Suppose $\mathfrak{c}_1 \in \Sigma_1$ and the isotropy of $\mathfrak{c}_1$ is conjugate to a subgroup of $\Delta S_k$. In many (we conjecture all) cases, it is possible
to construct a real analytic path $\{\mathfrak{c}(\lambda)\in\Sigma_\lambda\dd \lambda \in [0,1]\}$ from a 
(unique) point $\mathfrak{c}_0\in\Sigma_0$ to $\mathfrak{c}_1$---see Figure~\ref{fig: path}; the path is \emph{not} linear in $\lambda$. 
Moreover, 
\begin{enumerate}
\item The point $\mathfrak{c}_0$ is determined by a set of equations (the ``consistency equations'') that are simpler 
than the equations for $\mathfrak{c}_1$. 
\item The point $\mathfrak{c}_0$ gives a good approximation to $\mathfrak{c}_1$ that improves as $k$ increases. 
More precisely, we may construct power series in $1/\sqrt{k}$ for $\mathfrak{c}_0$ and the initial terms
are the same as those for $\mathfrak{c}_1$ (for more detail, see Section~\ref{sec: asym}).
\end{enumerate}

Our interest in the indirect method stems partly from a belief that the approach sheds light on the problem of desingularization (see the discussion in Section~\ref{sec: cp_comm}). 

We conclude this section with an outline of the construction of the path  $\{\mathfrak{c}(\lambda)\in\Sigma_\lambda\dd \lambda \in [0,1]\}$. Details are 
in Section~\ref{sec: comp}. Readers primarily interested
in the direct method and the results of Section~\ref{sec: asym}, should skim through the remainder of this section and the beginning of Section~\ref{sec: comp}, where we derive
the consistency equations if the isotropy is $\Delta S_k$ (type A). The critical point equations we use are in Section~\ref{sec: cpequations}. 

\subsection{Paths of critical points for $\{\cal{F}_\lambda\dd \lambda \in [0,1]\}$}\label{sec: paths}
Assume $d = k$ (the arguments extend easily to $d > k$~\cite{SafranShamir2018,ArjevaniField2020b}) and $\VV = I_k$. 
Define the $\Gamma$-invariant affine linear subspace $\mathbb{P}_{k,k}$ of $M(k,k)$ by
\[
\mathbb{P}_{k,k} = \VV + \is{C} = \{\WW\in M(k,k) \dd \WW^\Sigma = \is{1}_{1,k}\}
\]
and recall that $\mathbb{P}_{k,k}=\Sigma_0$---the set of critical points of $\mathcal{F}_0$. Set
$ \mathbb{A}_{k,k} = \Omega_a \cap \mathbb{P}_{k,k}$ and note that $\mathbb{A}_{k,k}$ is a $\GG$-invariant open subset of $\mathbb{P}_{k,k}$.
 
Recall that $\Phi(\WW,\lambda)=\grad{\mathcal{F}_\lambda}(\WW)$ is the real analytic family of gradient vector fields associated to $\{\cal{F}_\lambda\dd \lambda \in [0,1]\}$ (Lemma~\ref{lem: reg}).
It follows from Proposition~\ref{prop: weakrelu} that 
\begin{equation}\label{eq: grad}
\Phi(\WW,\lambda)^i = \lambda S^i(\WW) + \frac{1}{2} \big(\WW - \VV\big)^\Sigma,\; i \in \is{k},
\end{equation}
where $S^i: \Omega_a \arr (\real^d)^\star$ is real analytic, $i \in \is{k}$. 

Now suppose that $\mathfrak{c}:[0,1] \arr \Omega_a$ is a real analytic curve of critical points for the family $\{\cal{F}_\lambda\}$. That is,
\[
\Phi(\mathfrak{c}(\lambda) ,\lambda)= \is{0},\; \lambda \in [0,1].
\]
Substituting in~\Refb{eq: grad}, we have
\begin{equation}\label{eq: gradp}
\lambda S^i(\mathfrak{c}(\lambda)) + \frac{1}{2} \big(\mathfrak{c}(\lambda) - \VV\big)^\Sigma = 0,\; i \in \is{k}, \lambda \in [0,1].
\end{equation}
Taking $\lambda = 0$, we have $\left(\sum_{j \in \is{k}}(\mathfrak{c}^j(0) - \vv^j)\right) = 0$---since $\mathfrak{c}(0) \in \Sigma_0$. Hence 
$\frac{1}{2} \left(\sum_{j \in \is{k}}(\mathfrak{c}^j(\lambda) - \vv^j)\right) = \lambda R(\lambda)$ where $R:[0,1]\arr  (\real^d)^\star$ is real analytic
and $R(0) = \mathfrak{c}'(0)$.
After dividing by $\lambda$ in \Refb{eq: gradp} and taking $\lambda = 0$, we obtain
\begin{equation}\label{eq: gradc}
S^i(\mathfrak{c}(0)) + R(0) = 0,\; i \in \is{k}.
\end{equation}
Since $R(0)$ depends on the derivative of $\mathfrak{c}(\lambda)$ at $\lambda = 0$, attempting to solve $\Phi(\mathfrak{c}(\lambda) ,\lambda)= \is{0}$ directly using the implicit function
theorem looks problematic because of the loss of differentiability. However, the row vector $R(0)$ is common to the equations \Refb{eq: gradc} and so 
\begin{equation}\label{eq: gradc2}
S^i(\mathfrak{c}(0)) = S^j(\mathfrak{c}(0)), \; i,j \in \is{k}, \, i\ne j.
\end{equation}
Denoting the matrix with rows $S^i$ by $\is{S} = [s_{ij}]$, we have 
\begin{eqnarray}
\label{eq: gradcc}
s_{i\ell}(\mathfrak{c}(0))&=&s_{i' \ell}(\mathfrak{c}(0)), \; i,i',\ell \in \is{k}, \, i\ne i'.
\end{eqnarray}
We refer to~\Refb{eq: gradcc} as the \emph{consistency equations}. Together with the condition $\mathfrak{c}(0)^\Sigma = \is{1}_{1,k}$, these equations  uniquely determine 
$\mathfrak{c}(0)$. Moreover, in the specific problems we consider, it possible to find a unique formal power series solution for $\mathfrak{c}(\lambda)$ and
it then follows from Artin's approximation theorem~\cite{Artin1968} that this solution must be a real analytic solution  on an interval $[0,\lambda_0]$, where $\lambda_0 > 0$. 
We have described the hard work. Once we are away from the singularity at $\lambda = 0$, the extension of the solution to $[0,1]$ is routine (we indicate some of the details
in the following section). 

One interesting feature of the analysis is that for the classes of critical points we look at later, 
$\mathfrak{c}(0)$ gives a very good approximation to $\mathfrak{c}(1)$. In order to make quantitative sense of this statement, we
need to bring symmetry to the forefront of our problem.
\subsubsection{The role of symmetry}
Suppose that $H$ is an isotropy group for the $\Gamma$-action on $M(k,k)$ and that
$\mathfrak{c}:[0,1] \arr \Omega_a^H$ is a real analytic curve of critical points of isotropy $H$ for the family $\{\cal{F}_\lambda\}$.
In this case the equations (\ref{eq: gradc2},\ref{eq: gradcc}) are defined on $\Sigma_0^H$ and typically at most
$\text{dim}(\Sigma_0^H)$ independent scalar equations need to be chosen from the set~\Refb{eq: gradcc}. 

We restrict attention to isotropy groups which are subgroups of $\Delta S_k = \Gamma_\VV$ and consider
$k$-dependent families of isotropy groups $H$ for which there exists $k(H) \in\pint$ such that $\text{dim}(M(k,k)^H)$ is independent of $k \ge k(H)$.
We call isotropy groups of this type \emph{natural}.
The next example illustrates the formal structure required.
\begin{exam}\label{ex: natk-1}
Let $k \ge 3$. Consider the isotropy group $\Delta S_{k-1}$, where $S_{k-1}$ is the subgroup of $S_k$ fixing $k \in \is{k}$. Following
Section~\ref{ex: deltak-1ex}, we have a linear isomorphism $\Xi = \Xi(k): \real^5 \arr M(k,k)^{\Delta S_{k-1}}$ with
$\Xi(k)(\boldsymbol{\xi}) = [A_{ij}(\boldsymbol{\xi})]$, where $i,j \in \{1,k-1\}$.  Define the
projection $\pi_{k}: M(k+1,k+1) \arr M(k,k)$ for $k \ge 3$  by deleting row and column $k$ from $\WW \in M(k+1,k+1)$. Observe that
$\pi_{k}$ naturally induces maps on the block structure. For example,  $\pi_{k}$ maps $A_{k,k}$ to $A_{k-1,k-1}$ and $A_{k,1}$ to $A_{k-1,1}$.  
\end{exam}
Without spelling out the details, we give two other examples of natural isotropy groups.
\begin{exams}[Natural isotropy groups]
(1) If $H = \Delta S_k$, then $\text{dim}(M(k,k)^H) = 2$, for all $k \ge k_0 = 2$. \\
(2) Let $s \ge 2$ and fix $p_1, \cdots, p_{s-1} \ge 2$,  Set $q =\sum_i p_i$ and $p_s = k - q$. If $k \ge k_0 = q+2$ and $H = \prod_{i=1}^s \Delta S_{p_i} $, then $\text{dim}(M(k,k)^H) = s(s+1)$. 
If $s = 1$, then $H = \Delta (S_p \times S_{k-p})$ and $\text{dim}(M(k,k)^H) = 6$, $k \ge k_0 = 4$.
\end{exams}

\begin{rems}
(1) Let $\cal{H}=\{H_k \subset \Delta S_k, k \ge k_0\}$ be a family of natural isotropy groups and suppose that $\text{dim}(M(k,k)^{H_k}) = m$. For $k \ge k_0$, we can use $\Xi(k)$
to pull back $\grad{\cal{F}}$ to a gradient vector field $\grad{f}$ on $\real^m$. Moreover, $\grad{f}$ may now be viewed as a $k$-dependent family on $\real^m$ where
$k \ge k_0$ is a \emph{real} parameter. In practice, this means that if for some integer value of $\bar{k} \ge k_0$ we can find a critical point $\mathfrak{c}(\bar{k})\in M(\bar{k},\bar{k})^{H_{\bar{k}}}$ of 
$\grad{f}$ with isotropy in $\cal{H}$, then we can vary $k$ continuously and track the evolution of $\mathfrak{c}(k)$. This can be done forwards or backwards as long as $k \ge k_0$.
Values of $\mathfrak{c}(k)$ when $k$ is an integer give critical points of $\grad{f}$ and so of $\grad{\cal{F}}$. Similar remarks hold for solutions of the consistency equations which
have isotropy in $\cal{H}$. This fixed point space approach offers a fast and easy way to compute critical points numerically. We say more about the numerics in the
next section.
From the point of view
of bifurcation theory, we definitely do see bifurcation, at non-integral values
of $k$, in $\grad{f}$, viewed as a $k$-dependent family. Bifurcation is addressed further in~\cite{ArjevaniField2020c}. 
We have not observed bifurcation in the $\lambda$-dependent family $\{\cal{F}_\lambda\}$. \\
(2) We have emphasized natural isotropy groups. Similar methods apply for
imprimitive maximal isotropy groups where the fixed point space has dimension 3 (Examples~\ref{ex: sk-1-ex}(2)) but this seems of less interest. \rend
\end{rems} 

\subsection{Outline of the indirect method for the family $\Delta S_{k-1}$, $k\ge 3$}\label{sec:methods}

We illustrate the general method by discussing the family described in Example~\ref{ex: natk-1}.
Fix $k \ge 3$, set $K = \Delta S_{k-1}\subset \Gamma$. Following Section~\ref{ex: deltak-1ex}, let 
$ \Xi: \real^5 \arr M(k,k)^K$ be the linear isomorphism parametrizing points in $M(k,k)^K$.  For $\boldsymbol{\xi}\in\real^5$ recall
the column sums
\begin{eqnarray}
\label{eq: rowsums1}
\Xi^\Sigma_j = \Xi(\boldsymbol{\xi})^\Sigma_j&=&\xi_1 + (k-2) \xi_2 + \xi_4, \; j < k\\
\label{eq: rowsums2}
\Xi^\Sigma_k = \Xi(\boldsymbol{\xi})^\Sigma_k&=&(k-1)\xi_3 + \xi_5.
\end{eqnarray}

Let $\WW: [0,1] \arr M(k,k)^K\cap \Omega_a$ be real analytic and
$\Phi(\WW(\lambda),\lambda) = 0$, $\lambda \in [0,1]$. Defining $\bxi:[0,1]\arr \real^5$ by $\bxi(\lambda) = \Xi^{-1}(\WW(\lambda))$,
we have
\begin{equation}\label{eq: basic}
\bxi(\lambda) = \bxi_0 + \lambda\wbxi(\lambda),\;\lambda \in [0,1],
\end{equation}
where $\bxi_0 = \bxi(0) = \Xi^{-1}(\WW(0))$ and $\wbxi(\lambda) = \lambda^{-1}(\bxi(\lambda) - \bxi_0)$.

Taking $\lambda = 0$, $\Phi(\WW(0),0) = \is{0}$ and so $\WW(0)\in \mathbb{P}_{k,k}^K$.
Write $\boldsymbol{\xi}_0 $ in component form as
$(\xi_{01},\xi_{02},\ldots,\xi_{05})$. Since $\Xi(\bxi_0) \in \mathbb{P}_{k,k}^K$, we have 
\begin{equation}
\label{eq: lin_cond}
\xi_{01} + (k-2)\xi_{02} + \xi_{04} = 1,\;\; \xi_{03} + (k-1)\xi_{05} = 1
\end{equation}
Since $\text{dim}(\mathbb{P}_{k,k}^K) = 3$, there exists a unique $\mathfrak{t} = (\rho,\nu,\vareps) \in \real^3$ such that
\[
\WW(0) = 
\left[\begin{matrix}
1+ \rho    &  \vareps&\cdots&\vareps & -\frac{\nu}{k-1} \\
\vareps    &  1 + \rho& \cdots&\vareps & -\frac{\nu}{k-1}\\
\vareps    &  \vareps & \cdots&\vareps & -\frac{\nu}{k-1} \\
\cdots     & \cdots   & \cdots & \cdots & \cdots  \\
\vareps    & \vareps  & \cdots & 1+\rho & -\frac{\nu}{k-1} \\
-\rho - (k-2)\vareps &-\rho - (k-2)\vareps  & \cdots&-\rho - (k-2)\vareps & 1 + \nu
\end{matrix}\right]
\]
We have $\xi_{01} = 1+\rho$, $\xi_{02} = \vareps$, $\xi_{03} = -\frac{\nu}{k-1}$, $\xi_{04} = -\rho - (k-2)\vareps$, 
and $\xi_{05} = 1 + \nu$.
Henceforth, set
$\WW(0) = \WW^\mathfrak{t}$ and denote the $i$th row of $ \WW^\mathfrak{t}$ by $\ww^{\mathfrak{t},i}$, $i\in \is{k}$.
Note that $\WW^\mathfrak{t} \in \mathbb{A}_{k,k}^K$ iff 
$1+\rho \ne \eps$ or $1+\rho = \eps$ and $\nu \ne -1 +1/k$ (rows are not parallel) and 
$\mathfrak{t} \ne \is{0}$.

Since $\Phi(\WW(0),0) = 0$, and we assume analyticity, 
$\Phi(\WW(\lambda),\lambda)$ is divisible by $\lambda$. Substituting in the formula
for the components of $\Phi_\lambda$ given by Proposition~\ref{prop: weakrelu},  we have
\begin{eqnarray}\label{eq: phihat}
\Phi(\WW(\lambda),\lambda)&= &\lambda \widehat{\Phi}(\WW(\lambda),\lambda)
 =  \lambda \widehat{\is{G}}_\lambda(\bxi),
\end{eqnarray}
where $\WW(\lambda) = \Xi(\bxi)$ and
\begin{equation}\label{eq: Ghat}
\begin{split}
\widehat{\is{g}}_{\lambda}^i(\bxi) =\frac{1}{2\pi}\sum_{j\in \is{k}, j \ne i} \left(\frac{\|\ww^j\|\sin(\theta_{\ww^i,\ww^j})}{\|\ww^i\|}\ww^i -\theta_{\ww^i,\ww^j}\ww^j\right)-  \\\\
\frac{1}{2\pi}\sum_{j\in \is{k}}\left(\frac{\sin(\theta_{\ww^i,\vv^j})}{\|\ww^i\|}\ww^i - \theta_{\ww^i,\vv^j}\vv^j\right) + 
\frac{1}{2}\Xi(\wbxi(\lambda))^\Sigma. 
\end{split}
\end{equation}
\begin{rem}\label{rem: con_grad_eq}
Formally, $\Phi_1$ and $\widehat{\Phi}_1$ differ \emph{only} in their final terms $\frac{1}{2}\WXi(\bxi)$ and $\frac{1}{2}\Xi(\wbxi(\lambda))^\Sigma$
(note that $\WXi(\bxi) = (\Xi(\bxi)-\VV)^\Sigma$, see Section~\ref{sec: cp_speq}). \rend
\end{rem}
Pulling back to $\real^5 \times \real$, define 
$\Psi: \real^5 \times \real \arr \real^5$ by
\begin{equation}
\label{eq: onfix}
\Psi(\boldsymbol{\xi},\lambda) = \Xi^{-1}(\lambda^{-1}\Phi(\Xi(\boldsymbol{\xi}),\lambda))
 = \Xi^{-1}\widehat{\Phi}(\is{W}(\lambda),\lambda)
\end{equation}
and note that if
$\boldsymbol{\xi}:[0,1]\arr \real^5$, $\boldsymbol{\xi}_0$ satisfies~\Refb{eq: lin_cond}, and $\Psi(\boldsymbol{\xi}(\lambda),\lambda) = 0$,  then
$\WW(\lambda) = \Xi(\boldsymbol{\xi}(\lambda))$ will solve $\Phi(\WW(\lambda),\lambda) = 0$.

The expressions for the rows
$\widehat{\is{g}}^i_{\lambda}$ all include $\Xi(\wbxi(\lambda))^\Sigma$ which depends on 
a derivative of $\bxi$.  Our approach 
is to assume a formal power series solution $\bxi(\lambda)=\sum_{n=0}^\infty \bxi_n \lambda^n$ and then
verify the coefficients $\bxi_n $ exist and are uniquely determined. It then follows the analyticity of $\Phi$ and Artin's implicit function theorem~\cite{Artin1968} that
$\bxi(\lambda)$ is real analytic and the formal power series for $\bxi$ converges to a unique solution. 
For this approach to work we need to (a) Find $\bxi_0$ (starting the
induction), (b) show that each $\bxi_n$ is uniquely determined, $n > 0$.   

Suppose $\bxi(\lambda) = \sum_{n=0}^\infty \bxi_n \lambda^n$ (formal power series).  Write
$\bxi_n = (\xi_{n1},\cdots,\xi_{n5})\in \real^5$, $ n \ge 0$.
When $n = 1$, we  often write $\xi'_{0i}$ rather than $\xi_{1i}$. We use similar notational conventions
for  $\wbxi(\lambda) = \sum_{n=0}^\infty \wbxi_n \lambda^n$ ($\wbxi_n = \bxi_{n+1}$, $n \ge 0$).

First we find $\WW^\mathfrak{t}$ and hence $\boldsymbol{\xi}_0$. This step will also determine the column sums
$\xi'_{01} + (k-2) \xi'_{02} + \xi'_{04}$ and $(k-1)\xi'_{03} + \xi'_{05}$  which will not be zero.
Next we construct $\wbxi$. For this we use methods based on the implicit function theorem to express $(\wxi_2,\wxi_3,\wxi_4)$ as a real analytic function of $(\wxi_1,\wxi_5,\lambda)$. 
Using this representation, we find a unique formal power series solution for $\wbxi$ and then use 
Artin's implicit function theorem, and the analyticity of $\Phi$, to show that the formal solution is real analytic and unique on some $[0,\lambda_0]$, $\lambda_0 > 0$.
Since $\boldsymbol{\xi}_0$ is determined in the first step, we now have a real analytic solution
$\bxi(\lambda) = \bxi_0 + \lambda \wbxi(\lambda)$ on $[0,\lambda_0]$. Using results from Section~\ref{sec: asym}, we may
use standard continuation methods to show that $\boldsymbol{\xi}$ is real analytic on $[0,1]$. Finally, $\Xi(\boldsymbol{\xi}(1))\in M(k,k)^K$ is a 
critical point of $\Phi$.
\begin{rems}
(1) The term $\Xi(\wbxi)^\Sigma$ 
makes it difficult to extend our method to $C^r$ maps, $r < \infty$---at least without a loss of differentiability. See
Tougeron~\cite[Chapter 2]{Tougeron1968} for $C^\infty$ versions of Artin's theorem. \\
(2) For small values of $k$, the easiest way to find $\bxi(0)$ is numerically. For larger values of $k$ (likely all $k \ge k_0$), $\bxi(0)$ is given by a power
series in $1/\sqrt{k}$ and the initial terms of the series give a good approximation to $\bxi(0)$. 
Moreover, $\bxi(0)$ gives a quantifiably good approximation to the critical point $\bxi(1)$ (see Section~\ref{sec: asym}).
\rend
\end{rems}

\section{Solution curves for $\Phi_\lambda$ with isotropy $\Delta S_k$ or $\Delta S_{k-1}$.}\label{sec: comp}
We assume $d = k\ge 3$ and $\VV = I_k$; results extend 
to $d > k$~\cite{SafranShamir2018,ArjevaniField2020b}. We 
refer to Section~\ref{sec:methods} for the definition of $\widehat{\Phi}$ and $\widehat{\is{G}}_\lambda$ (see (\ref{eq: phihat},\ref{eq: Ghat})).

\subsection{Solutions of $\Phi_\lambda$ with isotropy $\Delta S_k$}\label{sec: dek}
If $\WW=[w_{ij}] \in M(k,k)$, then $\Gamma_\WW = \Delta S_k$ iff diagonal entries are equal and 
off-diagonal entries are equal but different from the diagonal entries. Since $\text{dim}(\mathbb{P}_{k,k}^{\Delta S_k}) = 1$, $\WW^\rho \in \mathbb{P}_{k,k}^{\Delta S_k}$
is uniquely specified by $\rho \in \real$ if we define
\[
w_{ii}=1+\rho,\; i \in \is{k}, \quad
w_{ij} =  -\rho/(k-1) \defoo \vareps, \; i,j \in \is{k},\,i \ne j.
\]
Provided $\rho \ne -1 + \frac{1}{k}$, $\Gamma_{\WW^\rho} = \Delta S_k$.
Since $\WW^\rho \in \Sigma_0$,
$\Phi(\WW^\rho,0) = \is{0}$ for all $\rho \in \real$.

We seek real analytic solutions $\WW:[0,1]\arr  M(k,k)^{\Delta S_k}$ to $\Phi_\lambda = \is{0}$. 
As shown in Section~\ref{sec: paths}, we may write $\WW(\lambda)$ in the form
$\WW(\lambda) =\Xi(\bxi_0) + \lambda \Xi(\wbxi(\lambda))$, where 
$\bxi: [0,\lambda] \arr \real^2$, $\Xi(\bxi_0) = \WW^\rho$, $\wbxi(\lambda) = \lambda^{-1}(\bxi(\lambda) - \bxi_0)$,
and $\Xi: \real^2 \arr M(k,k)^{\Delta S_k}$ is the linear isomorphism of Section~\ref{ex: deltak-1ex}. 
\subsubsection{Notational conventions}
Denote the $i$th row of $\WW(\lambda)$ by $\ww^i$ (implicit dependence on $\lambda$). It follows from the $\Delta S_k$ symmetry that 
$\|\ww^i\|$, $\ip{\ww^i}{\ww^j}$ and $\ip{\ww^i}{\vv^j}$ are independent of $i,j \in \is{k}$, $i \ne j$, and
$\ip{\ww^i}{\vv^i}$ is independent of $i \in \is{k}$.  Set $\tau = \|\ww^i\|$, let $\Theta$ (resp.~$\alpha$)
denote the angle between the rows $\ww^i, \ww^j$ (resp.~$\ww^i,\vv^j$), $i \ne j$,
and $\beta$ denote the angle between the rows $\ww^i$ and $\vv^i$. In case $\lambda = 0$, we add the subscript $0$ writing, for example, $\Theta_0$ rather than $\Theta(0)$.
The terms $\tau,\,\Theta,\,\alpha,$ and $\beta$ depend real analytically on $\lambda$ (and $\bxi, \wbxi \in \real^2$) provided
none of the rows $\ww^i, \vv^j$ are parallel (which is true if $\rho\notin \{0, -1+\frac{1}{k}\}$), and $|\lambda|$ is sufficiently small).

\subsubsection{Determination of $\bxi_0$ and the consistency equation}  
Noting Remark~\ref{rem: con_grad_eq}, we see from Section~\ref{sec: cpequations} (case $p = 0$, with $\xi_1 = 1+\rho,\, \xi_2 = -\frac{\rho}{k-1}$) 
that $\widehat{\Phi}(\WW^\rho,0) = \is{0}$ only if $\rho$ satisfies the \emph{consistency} equation
\begin{equation}\label{eq: consk}
(P_0 + \Theta_0) \left(1+\rho + \frac{\rho}{k-1}\right) + \beta_0 -\alpha_0 = 0,
\end{equation}
where
\[
P_0 = (k-1)\left(\sin(\Theta_0) - \frac{\sin(\alpha_0)}{\tau_0}\right) - \frac{\sin(\beta_0)}{\tau_0}.
\]
In terms of $\widehat{\Phi}$, if $\WW(\lambda) = \Xi(\bxi(\lambda))$, then $\widehat{\Phi}(\WW(\lambda),\lambda)= \widehat{\is{G}}_\lambda(\bxi(\lambda))$,
where 
\begin{align*}
\widehat{\is{g}}^i_{\lambda} =&\frac{1}{2\pi}\left[\sum_{j\in \is{k}, j \ne i} \left(\sin(\Theta)\ww^i -\Theta\ww^j\right)-  
\sum_{j\in \is{k}, j\ne i}\left(\frac{\sin(\alpha)}{\tau}\ww^i - \alpha\vv^j\right)\right] - \\
& \frac{1}{2\pi}\left(\frac{\sin(\beta)}{\tau}\ww^i - \beta\vv^i\right) + \frac{1}{2}\Xi(\wbxi(\lambda))^\Sigma,\; i  \in \is{k}.
\end{align*}
We may write $\Xi(\wbxi(\lambda))^\Sigma = a(\lambda)\is{1}_{1,k}$, where  
\begin{eqnarray}
\label{eq: derb}
a(\lambda) & = & (\xi'_1 + (k-1)\xi'_2)(0) + O(\lambda).
\end{eqnarray}
Pull back $\widehat{\Phi}$ by $\Xi$ to $\Psi: \real^2 \times \real \arr \real^2$, where
\begin{equation}
\label{eq: fixSk}
\Psi(\boldsymbol{\xi},\lambda) = (\psi_1,\psi_2)(\bxi,\lambda) = \Xi^{-1} \widehat{\Phi}(\WW(\lambda),\lambda),
\end{equation}
and $\psi_1 = \widehat{g}_{\lambda,11}$, $\psi_2 = \widehat{{g}}_{\lambda,12}$
($\widehat{{g}}_{\lambda,ii} = \psi_1$, $i \in \is{k}$,
and $\widehat{{g}}_{\lambda,ij} = \psi_2$, if $i\ne j$).

Now $\Psi(\boldsymbol{\xi},\lambda) = 0$ iff
$\psi_\ell(\bxi(\lambda),\lambda) = 0$, $\ell \in\is{2}$, $\lambda \in [0,1]$. Hence $\Psi(\bxi_0,0) = 0$ only if
$\psi_1(\bxi_0,0) = \psi_2(\bxi_0,0)$; the  same equation as~\Refb{eq: consk}, but using the variable $\bxi_0=(1+\rho,-\rho/(k-1)$. 
In particular, solutions of $\widehat{\Phi}(\Xi(\bxi_0),0) = \is{0}$ 
are given by solutions of $(P_0 + \Theta_0) (\xi_{01} -\xi_{02} ) + \beta_0 -\alpha_0 = 0$ satisfying the constraint
$\xi_{01} + (k-1)\xi_{02} = 1$.
\subsubsection{Solutions of the consistency equation}
One solution of~\Refb{eq: consk} is given by $\rho = 0$ (with $\Theta = \alpha = \pi/2$, $\tau = 1$ and $\beta = 0$). 
This is the known solution
$\WW = \VV$ of $\Phi_\lambda$, $\lambda \in [0,1]$.  Two additional solutions with isotropy $\Gamma$ are given by Example~\ref{ex: non-diag}. Neither give a
spurious minimum of $\cal{F}$. For $k \ge 3$, there is 
also a solution with isotropy $\Delta S_k$ which is not equal to $\VV$. These solutions, and the associated critical point 
$\Xi(\bxi(1))$, are referred to as being of \emph{type A}.  For $k \ge 6$, $\bxi(1)$ gives a spurious minimum of $\cal{F}$~\cite{ArjevaniField2020b}.
If $k = 6$, $\rho = -1.66064$\footnote{Numerical computation shown to 5 significant figures} and 
$\bxi_0 =  (-0.66064,0.33213)$. Although $\Xi(\bxi_0)$ is not a critical point of $\grad{\cal{F}}$, 
it gives a fair approximation to $\Xi(\bxi(1))\in\Sigma_1$ since $\bxi(1) = (-0.66340, 0.33071)$. The approximation 
improves rapidly with increasing $k$.

Assume the type A solution $\bxi_0$. Using~\Refb{eq: derb} and the formula for $\widehat{g}_{0,11}=\psi_1$, 
we see that the initial values $1 + \rho = \xi_{01}$, $\vareps = \xi_{02}$, determine the initial value $\xi'_{01} + (k-1)\xi'_{02}$ according to
\begin{equation}\label{eq: xi0_1}
\begin{split}
\xi'_{01} + (k-1)\xi'_{02}  = & \frac{1}{\pi}\left[(k-1)\left(\frac{\sin(\alpha_0)}{\tau_0} - \sin(\Theta_0)\right)\right]\xi_{01}\\
  &+ \frac{1}{\pi} \left((k-1)\xi_{02}\Theta_0 - \beta_0 + \frac{\sin(\beta_0)}{\tau_0}\xi_{01}\right).
\end{split}
\end{equation}
\subsubsection{Construction of the curve $\bxi(\lambda)$}
We emphasize the construction of the initial part of  $\bxi(\lambda)$ as this addresses the singularity at $\lambda = 0$. 
We omit details of the extension of $\bxi$ to all of $[0,1]$. This is a standard continuation that can be done rigorously using the results of 
Section~\ref{sec: asym}. The main technical problem is to show that $\Xi(\bxi([0,1]))\subset \Omega_a \cap M(k,k)^{\Delta S_k}$. 

We proceed by finding a formal power series solution for $\bxi(\lambda)$. 
The constant term $\bxi_0$ is known by the previous step and so we regard the variable as $\wbxi$. Set
$\xi'_{01} = \wxi_{01}$, $\xi'_{02} = \wxi_{02}$. We follow the notation previously given for norms and angles, using
the subscript ``0'' if evaluated at $\lambda = 0$.  Let $\rho$ denote the solution of \Refb{eq: consk}.
Define the constants
\begin{eqnarray*}
\trho&=&1+ \rho,\;\vareps = -\rho/(k-1),\;\eta=\trho+ (k-2)\vareps \\
A&=&2 \trho\vareps + (k-2)\vareps^2 = \langle \ww^{\rho,i},\ww^{\rho,j}\rangle,\; i \ne j,
\end{eqnarray*}
and note $\trho  = \xi_{01}$, $\vareps  = \xi_{02}$ and $\eta = 1 - \vareps$.
We estimate norm and angle terms ignoring terms of order $\lambda^2$ and treating $ \wxi_{1}, \wxi_{2}$ as variables. 
It is helpful to define some constants:
\begin{align*}
&J_1 = \vareps - \frac{A}{\tau_0^2}\trho,\; J_2 = \eta  - \frac{A}{\tau_0^2}(k-1)\vareps,\; 
&K_1 = \frac{\vareps\trho}{\tau_0^2},\;  K_2 = \frac{(k-1)\vareps^2}{\tau_0^2}-1\\
&L_1 = \sin^2(\alpha_0) - \frac{\vareps^2}{\tau_0},\; L_2 = \sin^2(\beta_0) -\frac{\trho^2}{\tau_0},\; &M_1 =1 - \frac{\trho^2}{\tau_0^2} ,\; M_2 =  - \frac{(k-1)\vareps\trho}{\tau_0^2}\\
& N_1 = \tau_0 + (k-1)\sin^2 (\alpha_0) - \frac{(k-1)\vareps^2}{\tau_0},\; &N_2 = \tau_0 + \sin^2(\beta_0) - \frac{\trho^2}{\tau_0}  \\
&P = (k-1)\left(\sin(\Theta_0)- \frac{\sin(\alpha_0)}{\tau_0}\right) - \frac{\sin(\beta_0)}{\tau_0} & 
\end{align*}
We find that
$\tau(\lambda)^{-1}  = \frac{1}{\tau_0} - \frac{\lambda }{\tau_0^3}\left(\trho \wxi_{1} + (k-1) \vareps \wxi_{2}\right)$ and
{\small
\begin{align*}
\Theta(\lambda) & = \Theta_0 - \frac{2 \lambda}{\tau_0^2 \sin(\Theta_0)}\left(J_1\wxi_{1} +J_2\wxi_{2}  \right), \;
\sin(\Theta(\lambda))  =\sin(\Theta_0) -  \frac{2A \lambda}{\tau_0^4 \sin(\Theta_0)}\left(J_1\wxi_{1} +J_2\wxi_{2} \right) \\
\alpha(\lambda) & = \alpha_0 + \frac{\lambda}{\tau_0\sin(\alpha_0)}\big(K_1\wxi_{1}+ K_2\wxi_{2} \big),\; 
\frac{\sin(\alpha(\lambda))}{\tau_0(\lambda)}  = \frac{\sin(\alpha_0)}{\tau_0} - \frac{\lambda\trho N_1\wxi_{1}}{\tau_0^3\sin(\alpha_0)} - \frac{\lambda\vareps N_2\wxi_{2}}{\tau_0^3\sin(\alpha_0)}  \\
\beta(\lambda) & = \beta_0 -  \frac{\lambda}{\tau_0\sin(\beta_0)}\big(M_1\wxi_{1}+M_2\wxi_{2}  \big),\; 
\frac{\sin(\beta(\lambda))}{\tau_0(\lambda)}  =   \frac{\sin(\beta_0)}{\tau_0} - \frac{\lambda \trho N_2\wxi_{1}}{\tau_0^3\sin(\beta_0)} -
\frac{\lambda(k-1)\vareps L_2 \wxi_{2}}{\tau_0^3\sin(\beta_0)}
\end{align*}
}
\normalsize

Since $\psi_1, \psi_2$ vanish at $(\bxi_0,0)$, we may define $h_i(\wbxi,\lambda) = \lambda^{-1} \psi_i(\bxi,\lambda)$, for $i \in\is{2}$. 
Substituting in the formula for
$\widehat{{g}}_{\lambda,1i}$, $i\in\is{2}$, we  find that
{\small
\begin{align*}
h_1(\wbxi,\lambda) =\; & P\wxi_{1}-
 \frac{2A(k-1)\trho}{\tau_0^4 \sin(\Theta_0)}\left(J_1\wxi_{1}+J_2\wxi_{2} \right)- 
(k-1)\Theta_0  \wxi_{2} + \frac{2(k-1)\vareps}{\tau_0^2 \sin(\Theta_0)}\left(J_1\wxi_{1} +J_2\wxi_{2}  \right)+\\
& \frac{(k-1)\trho}{\tau_0^3\sin(\alpha_0)}\left[\trho L_1 \wxi_{1} + 
\vareps N_1 \wxi_{2}\right] + 
\frac{\trho}{\tau_0^3\sin(\beta_0)}\left[\trho N_2\wxi_{1} + (k-1)\vareps L_2\wxi_{2}\right] - \\
&  \frac{1}{\tau_0\sin(\beta_0)}\left(M_1\wxi_{1} + 
M_2 \wxi_{2}\right)+\pi\Xi_1(\wbxip)^\Sigma  + O(\lambda)
\end{align*}
\begin{align*}
h_2(\wbxi,\lambda)\; & = P\wxi_{2}- 
\frac{2A(k-1)\eps}{\tau_0^4 \sin(\Theta_0)}\left(J_1\wxi_{1}+J_2\wxi_{2} \right)-
\Theta_0 (\wxi_{1} + (k-2) \wxi_{2}) + \\
&\frac{2\eta}{\tau_0^2 \sin(\Theta_0)}\left(J_1\wxi_{1} +J_2\wxi_{2}  \right)+
\frac{(k-1)\vareps}{\tau_0^3\sin(\alpha_0)}\left[\trho L_1\wxi_{1} + 
\vareps N_1 \wxi_{2}\right] + \\
& \frac{\vareps}{\tau_0^3\sin(\beta_0)}\left[\trho N_2 \wxi_{1} +
\vareps L_2 \wxi_{2}\right] + 
\frac{1}{\tau_0\sin(\alpha_0)}\left(K_1\wxi_{1}+ 
K_2 \wxi_{2} \right) +\pi\Xi_{2}(\wbxip)^\Sigma+ O(\lambda)
\end{align*}
}
\normalsize
Set $h_1 - h_2 = H^{12}$. We have $H^{12}(\wbxi_0,0) = A_1 \wxi_{01} + A_2 \wxi_{01}$, where
{\small
\begin{equation}\label{eq:  A1A2}
A_1 = \frac{\partial H^{12}}{\partial \wxi_1}(\wbxi_0,0),\quad A_2 = \frac{\partial H^{12}}{\partial \wxi_2}(\wbxi_0,0).
\end{equation}
and 
\begin{align*}
A_1  =\; & P  + \Theta_0 - 
  \left(\frac{2A(k-1)(1- k\vareps)}{\tau_0^4 \sin(\Theta_0)} + \frac{2(1-k\vareps)}{\tau_0^2 \sin(\Theta_0)}\right)J_1 + \\
&  \frac{(k-1)\trho(1-k\vareps)}{\tau_0^3\sin(\alpha_0)}L_1 + \frac{\trho(1-k\vareps)}{\tau_0^3\sin(\beta_0)}N_2 -\\
&  \frac{M_1}{\tau_0\sin(\beta_0)} - \frac{K_1}{\tau_0\sin(\alpha_0)}
\end{align*}
\begin{align*}
A_2 =\; & -P - \Theta_0- 
\left(\frac{2A(k-1)(1-k\eps)}{\tau_0^4 \sin(\Theta_0)} +\frac{2(1-k\vareps)}{\tau_0^2 \sin(\Theta_0)}\right)J_2+\\
&  \frac{(k-1)\vareps(1-k\vareps)}{\tau_0^3\sin(\alpha_0)}N_1 + 
\frac{(k-1)\vareps(1-k\vareps)}{\tau_0^3\sin(\beta_0)}L_2 - \\
& \frac{M_1}{\tau\sin(\alpha_0)} - \frac{M_2}{\tau_0\sin(\beta_0)}
\end{align*}
} \normalsize

\noindent Note that $A_1, A_2$ do not depend on $\wbxi_0$ and $\Xi_1(\wbxip)^\Sigma=\Xi_{2}(\wbxip)^\Sigma$. 
\begin{rem}\label{rem: A1_A2}
Numerics indicate that over the range $3 \le k \le 15000$, $A_1$ is strictly positive and increasing and
$A_2$ is strictly negative and decreasing. For $k=6$, $A_1 \approx4.9889$, $A_2\approx -9.7101$.
The dominant terms in the expressions for $A_1$ and $A_2$
are 
\[
\frac{(k-1)\trho(1-k\vareps)}{\tau_0^3\sin(\alpha_0)}\sin^2(\alpha_0),\;\text{and }
\frac{(k-1)\vareps(1-k\vareps)}{\tau_0^3\sin(\alpha_0)}(\tau_0 + (k-1)\sin^2 (\alpha_0)).
\]
An analysis of $A_1, A_2$, based on Section~\ref{sec: asym}, proves that
$\lim_{k \arr\infty} \frac{A_1}{k} = 1$, $\lim_{k \arr\infty} \frac{A_2}{k} = -2$.  These estimates are
consistent with the numerics. For example, if $k =10000$, $A_1 \approx 0.99986\times 10^4$ and $A_2 \approx -1.9997\times 10^4$.
In what follows we assume $A_1 > 0 > A_2$ for all $k \ge 3$.  
\rend
\end{rem}
\subsubsection{Computation of $ \wxi_{01},\wxi_{02}$}
If $H^{12}(\wbxi_0,0) = 0$, then $A_1 \xi'_{01} + A_2 \xi'_{02} = 0$ and so, with \Refb{eq: xi0_1},  
we have two linear equations for $ \xi'_{01}, \xi'_{02}$.
\begin{exam}
Taking $k =6$, and the values for $A_1, A_2$ given in Remark~\ref{rem: A1_A2}, we find that 
$ \xi'_{01}\approx -1.68903\times 10^{-3}$, $\xi'_{02} \approx -8.67792 \times 10^{-4}$. The small
values of the derivatives hint at the good approximation to $\bxi(1)$ given by $\bxi_0$. 
\end{exam}
 
We can compute $ \xi'_{01},\xi'_{02}$ for all $k \ge 3$ provided that $A_2/A_1 \ne k-1$. 
By Remark~\ref{rem: A1_A2}, $A_1, A_2$ are always of opposite sign and so $A_2/A_1 \ne k-1$. 
Hence the equations are consistent and solvable for all $k \ge 3$.

\subsubsection{Application of the implicit function theorem} 
Since $H_{12}(\wbxi_0,0) = 0$, and $A_1, A_2 \ne 0$, 
the implicit function theorem for real analytic 
maps 
applies to 
$H^{12}(\wxi_1, \wxi_2,\lambda)$ and so we may express $\wxi_1$ as an analytic function of $(\wxi_2,\lambda)$ on a 
neighbourhood of $(\wxi_{02},0)$ (using $A_1 \ne 0$), or $\wxi_2$ 
as an analytic function of $(\wxi_1,\lambda)$ on a neighbourhood of $(\wxi_{01},0)$ (using $A_2 \ne 0$). Choosing the first option, 
there exists an open  neighbourhood $U \times V$ of $(\wxi_{02},0) \in \real^2$ and analytic function 
$F: U \times V\subset \real^2 \arr \real$, such that
\[
H^{12}(F(\wxi_2,\lambda), \wxi_2,\lambda) = 0,\;\;\text{for all } (\wxi_2,\lambda) \in U \times V.
\]

Therefore, we may write
\begin{equation}\label{eq: xi12}
\wxi_1(\lambda) = \sum_{\substack{n=0 \\ m=1}}^\infty \alpha_{mn} \wxi_2^m \lambda^n,
\end{equation}
where $\wxi_1(\lambda) = \lambda^{-1}(\xi_1(\lambda) - \xi_{01})$ and $\alpha_{10} = -A_1 /A_2 \ne 0$. We now look 
for a unique formal power series solution 
$\wbxi(\lambda) = \sum_{p=0}^\infty \wbxi_p \lambda^p$ to 
$\Psi(\bxi,\lambda) = 0$.  By what we have computed already, we know that $\wbxi_0 = (\xi'_{01},\xi'_{02})$ and is uniquely determined. It follows
from~\Refb{eq: xi12} that it suffices to determine the coefficients in the formal power series for $\wxi_2(\lambda)$ since these
uniquely determine the coefficients in the formal power series for $\wxi_1(\lambda)$. Proceeding inductively, suppose we have
uniquely determined $\wbxi_0,\cdots,\wbxi_{p-1}$, where $p \ge 1$. It follows from ~\Refb{eq: xi12} that
$ \wxi_{p1} = K_p(\wxi_{02},\cdots,\wxi_{p2})$,
where $K_p(\wxi_{02},\cdots,\wxi_{p2}) = \widetilde{K}_p(\wxi_{02},\cdots,\wxi_{p-1\,2}) + \alpha_{m0} \wxi_{p2}$.
This gives one linear equation relating $\wxi_{p1}$ and $\wxi_{p2}$. We get a second linear equation by observing that at $\lambda = 0$,
$
\frac{\partial^p H^1_1}{\partial \lambda^p} = -p!\pi (\wxi_{p1} + (k-1) \wxi_{p2}).
$
The two linear equations we have for $\wxi_{p1}$ and $\wxi_{p2}$ are consistent (see Remark~\ref{rem: A1_A2}) and so $(\wxi_{p1}, \wxi_{p2})$ 
are uniquely determined, completing the inductive step.

Our arguments show there is a unique formal power series solution $\bxi(\lambda)=\bxi_0 + \lambda\wbxi(\lambda)$ to $\Psi(\bxi,\lambda) = 0$. Since
$\Psi$ is real analytic on a neighbourhood of $(\bxi_0,0)$, it follows by Artin's implicit function theorem that the formal power series $\bxi(\lambda)$ converges to
the required unique  real analytic solution to $\Psi(\bxi,\lambda) = 0$ on $[0,\lambda_0]$, where $\lambda_0 > 0$. 

\subsection{Solutions of $\Phi_\lambda$ with isotropy $\Delta S_{k-1}$}\label{sec: ximap}
For $k \ge 3$, there are two critical points of $\cal{F}$ with isotropy $\Delta S_{k-1}$ which define local minima for $\cal{F}|M(k,k)^{\Delta_{k-1}}$. 
We refer to these critical points as being of \emph{types I and II}. Critical points of type II appear in~\cite[Example 1]{SafranShamir2018} and are identified as spurious minima of
$\cal{F}$ for $k\in [6,20]$. In~\cite{ArjevaniField2020b} it is shown that For all $k \ge 6$, critical points of type I and II define spurious minima.

We focus on the initial point $\bxi_0\in \Sigma_0$ of a path $\bxi(\lambda)$ connecting to
a type II critical point $\bxi_1\in \Sigma_1$ and describe the consistency equations that determine $\bxi_0$. We give few
details on the construction of the path since
the method is already described by the analysis of the type A solution and the many technical details needed for type II contribute little new
to the analysis.
\subsubsection{Basic notation and computations} \label{sec: basic}
If $\mathfrak{t} = (\rho,\nu,\vareps) \in \real^3$, define $\WW^\mathfrak{t} \in \mathbb{P}_{k,k}^K$
as in Section~\ref{sec:methods} and recall that $\Phi(\WW^\mathfrak{t},0) = \is{0}$ for all $\mathfrak{t}\in\real^3$.

\subsubsection{Norm, inner product, angle definitions, and computations for $\WW^\mathfrak{t}$} \mbox{  }\\
We follow similar conventions to those used for isotropy $\Delta S_{k}$. 
\begin{align*}
&\text{(1) For $i < k$: } 
\|\ww^{\mathfrak{t},i}\| =  \sqrt{(1+\rho)^2 + (k-2) \vareps^2 + (\frac{\nu}{k-1})^2} \defoo \tau_0 \\
&\text{(2) For $i = k$: } 
\|\ww^{\mathfrak{t},k}\| =  \sqrt{(k-1)(\rho+ (k-2)\vareps)^2 + (1+\nu)^2} \defoo \kappa_0 \\
&\text{(3) For $i, j < k$, $i \ne j$:} \\
&\hspace*{0.25in}\ip{\ww^{\mathfrak{t},i}}{\ww^{\mathfrak{t},j}} = \frac{\nu^2}{(k-1)^2} + 2(1+\rho)\vareps + (k-3)\vareps^2\defoo A \\
&\text{(4) For $i < k$:}  \\
&\hspace*{0.25in}\ip{\ww^{\mathfrak{t},i}}{\ww^{\mathfrak{t},k}} =-\left[ \rho(1+\rho) + \frac{\nu(1+\nu)}{k-1} +\vareps (k-2)(1+2\rho) + \vareps^2 (k-2)^2\right]\\
& \hspace*{1.0in}					\defoo  A_k 
\end{align*}
\begin{align*}
&\text{(5) For $i, j < k$, $i \ne j$: }
 \ip{\ww^{\mathfrak{t},i}}{\vv^j} = \vareps 
&\text{(6) For $i< k$: }&
\ip{\ww^{\mathfrak{t},i}}{\vv^k} = -\frac{\nu}{k-1} \\
&\text{(7) For $j < k$: } 
\ip{\ww^{\mathfrak{t},j}}{\vv^j} = -[\rho + (k-2)\vareps] 
&\text{(8) For $i< k$: }&
\ip{\ww^{\mathfrak{t},i}}{\vv^i} =  1+\rho \\
&\text{(9) For $i= k$: }
 \ip{\ww^{\mathfrak{t},k}}{\vv^k} =  1+\nu &
\end{align*}
\subsubsection{Angle Definitions I}
We follow the conventions of Sections~\ref{sec: cp_speq}, \ref{sec: dek}. For example,
$\Lambda_0$ will denote the angle between $\ww^{\mathfrak{t},i}$ and $\ww^{\mathfrak{t},k}$, $i < k$.
\begin{enumerate}
\item 
$\Theta_0 =  \cos^{-1}\left( \frac{A}{\tau_0^2}\right)$, where $i,j < k, i \ne j$.
\item 
$\Lambda_0
 =  \cos^{-1} \left(\frac{A_k}{\tau_0\kappa_0}\right)$, $i < k$.
\end{enumerate}
\subsubsection{Angle Definitions II}
For clarity, we use ``$0$'' as a superscript rather than subscript to denote the vale of an $\alpha$-angle at $\lambda=0$. 
\begin{enumerate}
\item 
$\alpha^0_{ij}=  \cos^{-1}\left( \frac{\ip{\ww^{\mathfrak{t},i}}{\vv^j}}{\tau_0}\right)
 =  \cos^{-1}\left(\frac{\vareps}{\tau_0}\right)$.
\item 
$\alpha^0_{ik}=  \cos^{-1}\left( \frac{\ip{\ww^{\mathfrak{t},i}}{\vv^k}}{\tau_0}\right)
 =  \cos^{-1}\left(-\frac{\nu}{(k-1)\tau_0}\right)$.
\item 
$\alpha^0_{ii}= \cos^{-1}\left( \frac{\ip{\ww^{\mathfrak{t},i}}{\vv^i}}{\tau_0}\right)
 =  \cos^{-1}\left(\frac{1 +\rho}{\tau_0}\right)$.
\item 
$\alpha^0_{kj}= \cos^{-1}\left( \frac{\ip{\ww^{\mathfrak{t},k}}{\vv^j}}{\kappa_0}\right)
 =  \cos^{-1}\left(-\frac{\rho + (k-2)\vareps}{\kappa_0}\right)$.
\item 
$\alpha^0_{kk}= \cos^{-1}\left( \frac{\ip{\ww^{\mathfrak{t},k}}{\vv^k}}{\kappa_0}\right)
 =  \cos^{-1}\left(\frac{1 +\nu}{\kappa_0}\right)$.
\end{enumerate}
We seek real analytic solutions to $\Phi(\WW(\lambda),\lambda) = 0$ of the form 
\[
\WW(\lambda) =\Xi(\bxi(\lambda)) = \Xi(\bxi_0) + \lambda\Xi(\wbxi(\lambda)), \; \lambda \in [0,1],
\]
where  $\Xi(\bxi_0) = \WW(0)$ and $\wbxi(\lambda) = \lambda^{-1}(\bxi(\lambda) - \bxi_0)$.  
As described in Section~\ref{main-section} and Example~\ref{eq:XI_def}, 
$\Xi(\boldsymbol{\xi})$ has rows $\Xi^1(\bxi),\cdots,\Xi^k(\bxi)$, where 
\[
\Xi^1(\bxi)= [\xi_1,\xi_2,\cdots,\xi_2,\xi_3],\cdots, \Xi^k(\bxi)= [\xi_4,\xi_4,\cdots,\xi_4,\xi_5] 
\]
\subsection{The equations for $\mathfrak{t}$ uniquely determining $\bxi_0$}\label{sec: eqs}
Following Section~\ref{sec:methods}, if $\WW^{\mathfrak{t}} \in \mathbb{P}^K_{k,k}$, and $\WW(\lambda)=\WW^{\mathfrak{t}} + $ we  set
$\widehat{\Phi}(\WW(\lambda),\lambda)  = \widehat{\is{G}}_\lambda \in M(k,k)^{\Delta S_{k-1}} $, where the rows of  $\widehat{\is{G}}_\lambda$ are given by~\Refb{eq: Ghat}.
Define $\Psi:\real^5 \times \real \arr \real^5$ by 
\[
\Psi(\bxi,\lambda) = (\psi_1,\cdots,\psi_5)(\bxi,\lambda) = \Xi^{-1}\widehat{\Phi}(\WW(\lambda),\lambda)
\]
Set  $\boldsymbol{\varphi}^i = \widehat{\is{g}}_{0}^i$, $i\in\is{k}$.
Computing we find that
\begin{enumerate}
\item
If $ i < k$, 
\begin{equation}
\label{eq: cons1}
\begin{split}
\boldsymbol{\varphi}^i  = &(k-2)\left(\sin(\Theta_0)-\frac{\sin(\alpha_{ij}^0)}{\tau_0}\right)\ww^{\mathfrak{t},i} - \Theta_0 \sum_{j=1, j \ne i}^{k-1} \ww^{\mathfrak{t},j}  + \\
&\left(\frac{\kappa_0}{\tau_0} \sin(\Lambda_0) -\frac{\sin(\alpha^0_{ik})}{\tau_0}\right)\ww^{\mathfrak{t},i} - \Lambda_0 \ww^{\mathfrak{t},k} - \frac{\sin(\alpha^0_{ii})}{\tau_0}\ww^{\mathfrak{t},i}+\\
& \sum_{j=1, j\ne i}^{k-1} \alpha_{ij}^0 \vv^j + \alpha^0_{ik} \vv^k + \alpha^0_{ii} \vv^i+
 \pi \Xi(\wbxi_0)^\Sigma 
\end{split}
\end{equation}
\item If $i = k$, 
\begin{equation}
\label{eq: cons2}
\begin{split}
\boldsymbol{\varphi}^k  = & \sum_{j=1}^{k-1} \left[\frac{\tau_0}{\kappa_0} \sin(\Lambda_0) \ww^{\mathfrak{t},k} - \Lambda_0 \ww^{\mathfrak{t},j} \right] + \pi \Xi(\wbxi_0)^\Sigma - \\
& \sum_{j=1}^{k-1} \left[\frac{\sin(\alpha^0_{kj})}{\kappa_0}\ww^{\mathfrak{t},k} -\alpha^0_{kj} \vv^j \right] - \left[\frac{\sin(\alpha^0_{kk})}{\kappa_0}\ww^{\mathfrak{t},k} -\alpha^0_{kk} \vv^k\right].
\end{split}
\end{equation}
\end{enumerate}
\subsection{Consistency equations}\label{sec: con}
A solution to $\boldsymbol{\varphi}^i = 0$, $i \in \is{k}$, determines the initial point $\bxi_0$ of the path from  $\mathbb{P}_{k,k}^{\Delta S_{k-1}}$ 
to the associated critical point of $\Phi_1$.
Since (\ref{eq: cons1},\ref{eq: cons2}) share the common term $\pi \Xi(\wbxi_0)^\Sigma$, we have the following \emph{consistency 
equations} defined on $\mathbb{P}_{k,k}^{\Delta S_{k-1}}$.
\begin{equation}\label{eq: con1}
\boldsymbol{\varphi}^\ell =  \boldsymbol{\varphi}^m, \; \ell,m\in \is{k}.
\end{equation}
The consistency equations determine $\mathfrak{t}$ and hence $\bxi_0 \in \mathbb{P}_{k,k}^{\Delta S_{k-1}}$.
Since $\widehat{\is{G}}_0$ is fixed by $\Delta S_{k-1}$, $\boldsymbol{\varphi}^i = (i,j)^c \boldsymbol{\varphi}^j$, 
$\varphi_{ik} = \varphi_{jk}$, $i,j \in \is{k-1}$, and
$\varphi_{kj} = \varphi_{k\ell}$,  $j,\ell < k$.
It follows that \Refb{eq: con1} may be reduced to exactly three scalar equations. For example,
\begin{equation}\label{eq: consist3}
\varphi_{11} = \varphi_{12} = \varphi_{k1},\; \varphi_{1k} = \varphi_{kk},
\end{equation}
where $\psi_1(\bxi_0,0) = \varphi_{11}$, $\psi_2(\bxi_0,0) = \varphi_{12}$, $\psi_3(\bxi_0,0) = \varphi_{k1}$,
$\psi_4(\bxi_0,0) = \varphi_{1k}$, $\psi_5(\bxi_0,0) = \varphi_{kk}$.
\begin{rem}
Noting Remark~\ref{rem: con_grad_eq}, \Refb{eq: consist3} follows from Section~\ref{sec: cpequations} (case $p = 1$, with $\xi_1 = 1+\rho,\, \xi_2 = \eps, \xi_5 = 1+ \nu$ and
$\bxi_0\in\Sigma_0$). 
 \rend
\end{rem}

It is helpful to identify certain terms in $\boldsymbol{\varphi}^1,\boldsymbol{\varphi}^k$. 
Define
\begin{eqnarray*}
P&=&(k-2)\left[\sin(\Theta_0)-\frac{\sin(\alpha_{ij}^0)}{\tau_0}\right] + \frac{\kappa_0 \sin(\Lambda_0)-\sin(\alpha^0_{ik})-\sin(\alpha^0_{ii})}{\tau_0} \\
Q&=&(k-1)\left[\frac{\tau_0 \sin(\Lambda_0) - \sin(\alpha^0_{kj})}{\kappa_0}\right] - \frac{\sin(\alpha^0_{kk})}{\kappa_0} \\
\boldsymbol{\alpha}^1 & = & (\alpha^0_{ii},\alpha^0_{ij},\alpha^0_{ij},\ldots,\alpha^0_{ij},\alpha^0_{ik})\\
\boldsymbol{\alpha}^k & = & (\alpha^0_{kj},\alpha^0_{kj},\alpha^0_{kj},\ldots,\alpha^0_{kj},\alpha^0_{kk})
\end{eqnarray*}
The equality $\boldsymbol{\varphi}^1 = \boldsymbol{\varphi}^k$ may be written 
\begin{eqnarray*}
P \ww^{\mathfrak{t},1} - \left[\Theta_0\sum_{j=2}^{k-1}\ww^{\mathfrak{t},j} + \Lambda_0 \ww^{\mathfrak{t},k} \right] + \boldsymbol{\alpha}^1
& = & Q \ww^{\mathfrak{t},k} - \Lambda_0 \sum_{j=1}^{k-1} \ww^{\mathfrak{t},j} + \boldsymbol{\alpha}^k
\end{eqnarray*}
Hence, we derive expressions for $\varphi_{11} = \varphi_{12}$, $\varphi_{11} = \varphi_{k1}$, and $\varphi_{1k} = \varphi_{kk}$:
\begin{align*}
 (P+\Theta_0)(\trho - \vareps)&= \alpha^0_{ij} - \alpha^0_{ii} \\ 
 P\trho + (Q+ 2\Lambda_0)(\rho+(k-2)\vareps) + \Lambda_0 - (k-2)\vareps\Theta_0 
&=\alpha^0_{kj} - \alpha^0_{ii} \\  
 (P-(k-2)\Theta_0)\left(\frac{-\nu}{k-1}\right) - (2\nu+1)\Lambda_0 - Q(1+\nu)
&=\alpha^0_{kk} - \alpha^0_{ik} 
\end{align*}
where $\trho = 1+\rho$. 
\begin{rem}We may rewrite the equations in terms of $\xi_1,\xi_2, \xi_5$ (see Section~\ref{sec: cpequations}, case $p=1$), 
eliminating $\xi_3,\xi_4$ using $\bxi_0\in\Sigma_0$.\rend
\end{rem}
\subsection{Numerics I: computing $\mathfrak{t}$}\label{sec: numI}
We consider small values of $k$ (for large $k$, see Section~\ref{sec: asym}).
In~\cite[Example 1]{SafranShamir2018}, numerical data for the case $k= 6$ 
indicates the presence of a local minimum for $\cal{F}$ in the fixed point space $M(6,6)^{\Delta S_5}$. Methods (op.~cit.) were based on SGD, with Xavier initialization
in $M(6,6)$ (not $M(6,6)^{\Delta S_5}$) and covered the range $6 \le k \le 20$.  Randomly initializing in $M(6,6)^{\Delta S_5}$,
gradient descent converges with approximately equal probability to one of four minima: either $\VV$ or 
\[
\is{A} = \left[\begin{matrix}
-0.66 & 0.33 &  \ldots & 0.33 \\
0.33 & -0.66 &  \ldots & 0.33 \\
\ldots & \ldots & \ldots & \ldots \\     
0.33 & 0.33 &  \ldots & -0.66  
\end{matrix}\right], \hspace*{0.9in}(\text{type A})
\] 
\begin{align*}
\is{B}_1& =& \left[\begin{matrix}
-0.59 & 0.39 & \ldots&0.39 & 0.01 \\
0.39 & -0.59 & \ldots&0.39 & 0.01\\
\ldots & \ldots & \ldots & \ldots \\     
0.39 & 0.39 &\ldots & -0.59 & 0.01 \\
0.02 & 0.02 & \ldots & 0.02& 1.07  
\end{matrix}\right],\hspace*{0.5in}& (\text{type I}) \\
\is{B}_2&=&\left[\begin{matrix}
0.99 & -0.05 &  \ldots& -0.05 & 0.31 \\
-0.05 & 0.98 &  \ldots&-0.05 & 0.31 \\
\ldots & \ldots & \ldots & \ldots & \ldots\\     
-0.05 & -0.05  & \ldots &-0.05 & 0.31 \\
0.22 & 0.22 &  \ldots & 0.22 & -0.60  
\end{matrix}\right], \hspace*{0.5in}&(\text{type II})
\end{align*}
We have $\Gamma_\is{A} = \Gamma_\VV = \Delta S_6$ and $\Gamma_{\is{B}_1} = \Gamma_{\is{B}_2} = \Delta S_5$.  
These minima for $\cal{F} |M(6,6)^{\Delta S_{5}}$ are all local minima of $\cal{F}$ on $M(6,6)$. 

Using the entries of $\is{A}, \is{B}_1, \is{B}_2$ as approximations for $\bxi_0=\mathfrak{t}$,  
we solve the consistency equations for $k = 6$ using Newton-Raphson.
Regarding $k$ as a real parameter, we can use this value of $\mathfrak{t}$ to compute $\mathfrak{t}$
for other values of $k$. 
We show the results, to $8$ significant figures, for $k = 6$ and $1000$ in Table~\ref{inftable1}.
The $k = 1000$ values used a $k$-increment of $\pm 0.1$, starting at $k = 6$, and 
$50$ iterations of Newton-Raphson for each step. 

\begin{scriptsize}
\begin{table}[h]
\begin{tabular}{|c||c||c|c|c|}

\hline
Solution  & $k$ & $1+\rho$                       & $1+\nu$              &  $\vareps$     \\
          &    &                              &                    &              \\  \hline \hline
\emph{type A}     & $6$ &  $-0.66063967$   &$0.66063967$          & $0.33212793$              \\ \hline
\emph{type I}  & $6$ &  $-0.58622786$ &  $1.067795110115$          & $0.39200518$             \\ \hline
\emph{type II}  & $6$ &  $0.98254382$ & $-0.58566032$          &$-0.054141651$              \\ \hline
\emph{type A}     & $1000$ &$-0.99799996$ & $-0.99799996$      & $1.99999996\times 10^{-3}$     \\ \hline
\emph{type I}  & $1000$ &$-0.99799546$ & $1+1.591580519\times 10^{-3}$ & $2.00334518\times 10^{-3}$ \\ \hline
\emph{type II}  & $1000$ &$ 1+2.43361217\times 10^{-6}$ & $-0.9947270019$ & $-1.305602504\times 10^{-6}$  \\ \hline
\end{tabular}
\vspace*{0.01in}
\caption{
Values of $\mathfrak{t} = (\rho, \nu,\vareps)$ associated to the critical points of types A, I and II for $k = 6, 1000$}
\label{inftable1}
\end{table}
\end{scriptsize}

\subsection{Construction of the curve $\bxi(\lambda)$}\label{sec: ordertwo}
We follow the method used for isotropy $\Delta S_k$ and compute the terms of order $\lambda^2$ in the power series expansion of $\Phi(\WW(\lambda),\lambda)$ at $\lambda = 0$.. 
This is an elementary, but lengthy, computation and the results are given in Appendix~\ref{sec: hotI}.

Solving the consistency equations uniquely determines $\bxi_0$. The components $\Xi_{j}(\wbxi_0)^\Sigma$, $j \in\is{k}$, are
uniquely determined by the requirement that $\lambda^{-1}\Phi(\WW(\lambda),\lambda)$ vanishes at $\lambda = 0$ 
(that is, $\boldsymbol{\varphi}^\ell = 0$, $\ell \in \is{k}$). Consequently, once $\bxi_0$ is determined, $\Phi(\bxi_0 + \lambda \wbxi(\lambda),\lambda)$ 
is divisible by $\lambda^2$. Setting
\[
\is{H}(\wbxi,\lambda) = (\is{h}^1(\wbxi,\lambda),\ldots,  \is{h}^k(\wbxi,\lambda)) = \lambda^{-2} \Phi(\Xi(\bxi_0 + \lambda \wbxi(\lambda)),\lambda),
\]
we may express $\widehat{\is{h}}^1(\wbxi) = \is{h}^1(\wbxi,0)$ and $\widehat{\is{h}}^k(\wbxi) = \is{h}^k(\wbxi,0)$ in terms of $\bxi_0$ and the variable $\wbxi$. 
Explicit formulas for $\widehat{\is{h}}^1(\wbxi)$ and $\widehat{\is{h}}^k(\wbxi)$ are given at the end of Appendix~\ref{sec: hotI}. 
\subsubsection{Construction of the initial part of the solution curves $\bxi(\lambda)$}
The differences $\zeta_1= {h}_{11}-{h}_{12}$, $\zeta_2 =  {h}_{11}- {h}_{k1}$, $\zeta_3 = h_{1k} -h_{kk}$ do not depend on the
common term $\pi\Xi(\wbxip)^\Sigma$.  Define $\boldsymbol{\zeta} : \real^5 \times \real \arr \real^3$ by
$\boldsymbol{\zeta}(\wbxi,\lambda) = (\zeta_1,\zeta_2,\zeta_3)(\wbxi,\lambda)$. Let $J$ denote the $3 \times 5$ 
Jacobian matrix $\left[\frac{\partial \zeta_j}{\partial \wxi_i}(\wbxi_0,0)\right]$.
If we let $J^\star$ denote the $3 \times 3$ submatrix defined by columns $2,3$ and $4$, then a numerical check verifies that 
$J^\star$ nonsingular, $k \ge 3$, and that $|J^\star| \uparrow \infty$ as $k \arr \infty$. A formal proof can be 
given using the results of Section~\ref{sec: asym}.
It follows from the implicit function theorem that there exist
analytic functions $F_2, F_3, F_4$ defined on a neighbourhood $U$ of 
$(\wxi_{10},\wxi_{50},0)\in \real^3$, such that if we set $\wxi_\ell = F_\ell(\wxi_1,\wxi_2,\lambda)$, $\ell = 2,3,4$, then
\begin{eqnarray*}
\wxi_{\ell 0}&=&F_\ell(\wxi_{10},\wxi_{50},0),\; \ell = 2,3,4\\
0 &=&\boldsymbol{\zeta}(\wxi_1,F_2(\wxi_1,\wxi_5,\lambda),F_3(\wxi_1,\wxi_5,\lambda),F_4(\wxi_1,\wxi_5,\lambda),\wxi_5,\lambda),  
\end{eqnarray*}
for $(\wxi_1,\wxi_5,\lambda) \in U$.

Following the same argument used for $\Delta S_k$, we construct a unique formal 
power series solution to $\boldsymbol{\zeta} = 0$ and use 
Artin's theorem to prove convergence of the formal power series.

\subsection{Numerics II}\label{sec: numII}
In Table~\ref{inftable4}, we show the computation of $\bxi(1) \in \Sigma_1$ for 
$k=6$ and types A, I, and II. The results for type II agree with those in Safran \& Shamir~\cite[Example 1]{SafranShamir2018}
to 4 decimal places---the precision used in~\cite{SafranShamir2018}. Note that $\|\Phi(\bxi(1),1)\|$ is the gradient norm (Euclidean norm on $M(6,6)$). 

\begin{scriptsize}
\begin{table}[h]
\begin{tabular}{|c||c|c|c|c|c|c|}
\hline
Isotropy        & $\xi_1(1)$  & $\xi_2(1)$  & $\xi_3(1)$    & $\xi_4(1)$    & $\xi_5(1)$   & $\|\Phi(\bxi(1),1)\|$ \\ 
type            &               &                 &                &                &      &        \\ \hline \hline
\emph{type A}    &$-0.663397$ &$0.330710$   & $0.330710$  & $0.330710$   &$-0.663397$  & $2.61\times 10^{-18}$         \\ \hline
\emph{type I}  &$-0.587730$ &$0.391154$   &$-0.0137989$  & $0.0167703$ & $1.0683956$  & $1.18 \times  10^{-18}$       \\ \hline
\emph{type II}  &$0.986704$  &$-0.0504134$ &$0.308001$   & $0.224516 $   & $-0.601512$ & $1.97 \times 10^{-18}$     \\ \hline
\end{tabular}
\vspace*{0.01in}
\caption{
Values of $\bxi(1)$ and error estimate $\|\Phi(\bxi(1),1)\|$ for $k=6$ and types A, I, II.}
\label{inftable4}
\end{table}
\end{scriptsize}
For type II critical points, 
\begin{align*}
&|\xi_1(1) - (1+\rho)|, |\xi_2(1) - \vareps|  \approx 0.004,\quad |\xi_5(1) - (1+ \nu)| \approx 0.06\\
&|\xi_3 - (-\nu/5)| \approx  0.009, \quad |\xi_4 - (-\rho - 4\eps)| \approx 0.009
\end{align*}
The approximation to the components of $\bxi(1)$ (in $M(6,6)^{\Delta S_5}$) given
by $1+\rho, \vareps, -\nu/5, -(\rho+4\eps)$, and $1+\nu$ is quite good. This is not unexpected since numerics indicate that
$|\xi_i'(0)|< 4.1 \times 10^{-3}|$, $i \in \is{5}$.  For large values of $k$, we refer to Section~\ref{sec: asym}. 
Practically speaking, to go from
$\bxi(0)$ to $\bxi(1)$ requires few iterations of Newton-Raphson. For k = 6, more than three iterations gives no increase in accuracy.

\subsubsection{Numerical methods}
Previously, we indicated the method of computation for $\mathfrak{t}$. As part of that computation, two affine linear equations are derived for the
derivative $\bxi'_0$. The next stage of the computation obtains three linear equations in $\bxi'_0$, using the second order conditions of
Section~\ref{sec: ordertwo}. Expressions for $\xi'_1(0), \xi'_5(0)$ in terms of the remaining unknowns are obtained from the two affine linear equations and
substituted in the three linear equations which are then solved using an explicit computation of the inverse matrix. The continuation of the solution to the
path $\bxi(\lambda)$ is obtained by incrementing $\lambda$ from $\lambda_{\text init} > 0$ to $\lambda = 1$ (larger values of $\lambda$ can be allowed). In the fastest case,
we initialize at $\bxi_0$ (determined by $\mathfrak{t}$) and solve directly for $\bxi(1)$ using Newton-Raphson and Cramer's rule. This works very well
for a wide range of values of $k$. To compute the path,  
we increase $\lambda$ in steps of $\lambda_{\text inc}$ where $\lambda_{\text inc}$ is either $0.1$, $0.01$, or $0.001$. We initialize 
at $\bxi_0 + \lambda_{\text inc} \bxi'_0$ and use Newton-Raphson at each step to find the zero of
$\Phi(\bxi(\lambda_n),\lambda_n)$, where $n > 0$ and $\lambda_1 = \lambda_{\text inc}$.   
For $k \in [4,20000]$, the critical point $\Phi(\bxi(1),1)$ obtained numerically appears to be \emph{independent} of the continuation method: 
the fastest method---directly computing $\Phi(\bxi(1),1)$ using the initialization $\mathfrak{t}$---gives exactly the same results as those obtained using
small increments of $\lambda$.  For this range of values of $k$, $\|\Phi(\bxi(1),1)\| < 10^{-14}$, with errors of order $10^{-18}$ or 
less for small values of $k$. 

\begin{rem}
A program written in C, using long double precision, 
was used to do the computations shown in this section. 
The program is available by email request to either author (related programs in Python are also 
available). Access to data sets of values 
of $\mathfrak{t}$, $\bxi'(0)$, $\bxi(1)$ and $\Phi(\bxi(1),1)$ and
critical points and values of types A, I, and II for $3 \le k \le 20000$ may be downloaded from the authors websites. \rend
\end{rem}
\subsection{Critical points with isotropy $\Delta (S_2 \times S_{k-2})$}\label{sec: extra}
All examples presented so far have had critical points in $M(k,k)^{\Delta S_{k-1}}$. We conclude the section with
a brief description of the family of \emph{type M} critical points which are defined for $k \geq 5$ and have isotropy $\Delta (S_{k-2} \times S_2)$.
Since $\Delta (S_{k-2}\times S_2)\not\supset\Delta S_{k-1}$, this
family does not lie in $M(k,k)^{\Delta S_{k-1}}$.  

Set $K = \Delta (S_{k-2} \times S_2)$. We have $\text{dim}(M(k,k)^K) = 6$. The parametrization $ \Xi: \real^6 \arr F$ is given in
Section~\ref{ex: deltak-1ex} and we recall that 
\[
\Xi(\bxi) =  \left[\begin{matrix} 
A_{k-2}(\xi_1,\xi_2) & \xi_3\is{1}_{k-2,2} \\
\xi_4\is{1}_{2,k-2} & A_{2}(\xi_5,\xi_6) 
\end{matrix}\right],
\]
We note the column sums
\begin{eqnarray}
\label{eq: rowsums1n}
\Xi(\boldsymbol{\xi})^\Sigma_i&=&\xi_1 + (k-3)\xi_2 + 2 \xi_4, \;  i \le k- 2\\
\label{eq: rowsums2n}
\Xi(\boldsymbol{\xi})^\Sigma_{i}&=&(k-2) \xi_3 + \xi_5 + \xi_6,\; i \ge k-1. 
\end{eqnarray}
Following the same strategy used for families of type II, we find solutions $\rho, \eps, \eta,\nu$
of the associated \emph{four} consistency equations. In this case, $1+\rho, 1+\nu$ correspond to $\xi_1,\xi_6$ respectively and
$\eps, \eta$ correspond to $\xi_2,\xi_5$ respectively. Set $\zeta_3 = -(\nu + \eta)/(k-2)$, and $\zeta_{4} = -(\rho + (k-3)\eps)/2$,
so that the column sums (\ref{eq: rowsums1n},\ref{eq: rowsums2n}) are 1 where $\zeta_3$ corresponds to $\xi_3$ and $\zeta_4$ to $\xi_4$.

Having computed $\rho,\cdots,\nu$, Newton-Raphson is used to compute the critical point $\isk{c}$. The results are
shown in Table~\ref{table: errors} for $k=10^4$ together with the approximation $\isk{c}_0$ given by $1+\rho,\eps,\cdots,1+\nu$. 
\begin{scriptsize}
\begin{table}[h]
\begin{tabular}{|c||c|c|c|c|c|c|}
\hline
        & $\xi_1$  & $\xi_2$  & $\xi_3$    & $\xi_4$    & $\xi_5$   & $\xi_6$ \\ \hline \hline
$\isk{c}_0$   &$1.000503 $ &$-2.567\times 10^{-8}$   &$1.999\times 10^{-4}$  & $1.283\times 10^{-4}$ & $ 1.929\times 10^{-4}$  & $-0.999$       \\ \hline
$\isk{c}$  &$1.000503 $ &$-2.567\times 10^{-8}$  &$1.999\times 10^{-4}$  & $1.283\times 10^{-4}$ & $ 1.929\times 10^{-4}$  & $-0.999$       \\ \hline
$|c_i^0-c_i|$  &$5.6\times 10^{-11}$ &$1.8 \times 10^{-12}$   &$1.2\times 10^{-8}$  & $7.6 \times 10^{-9}$ & $1.9 \times 10^{-8}$  & $4.4 \times 10^{-8}$       \\ \hline
\end{tabular}
\vspace*{0.05in}
\caption{Critical point and approximation given by $\isk{c}_0 = (1+\rho,\cdots,1+\nu)$ for $k = 10^4$. The components of $\isk{c}_0$, $\isk{c}$ are only given
to 3 significant figures. Higher precision was used for estimating $|c_i^0-c_i|$. Both $\cal{F}(\isk{c})$ and $\cal{F}(\isk{c}_0)$ are approximately
$0.59 \times 10^{-4}$.}
\label{table: errors}
\end{table}
\end{scriptsize}

\begin{rem}
Critical points of type M appear in the data sets of ~\cite{SafranShamir2018} as spurious minima for $9 \le k \le 20$. If 
$k = 10^4$, then $\cal{F}(\isk{c}) \approx 5.922\times 10^{-5}$ and, combined with objective value data for all $k\in [9,20000]$ 
strongly suggests that the decay of $\cal{F}(\isk{c})$ is approximately $0.6 k^{-1}$.
All of this is consistent with the observation that spurious minimum values are often close to the global minimum. Similar families
exist with isotropy $\Delta (S_{k-p} \times S_{p})$ for $p > 2$~\cite{ArjevaniField2020c}. Provided $p/k, k^{-1}$ are sufficiently small, the decay rate of
$\cal{F}(\isk{c})$ appears to be $O(k^{-1})$. The expectation is that these families also give spurious minima.
\end{rem}

\section{Asymptotics in $k$ for critical points types \emph{A, I} and \emph{II}}\label{sec: asym}

\subsection{Introduction}
Assume $d = k$. In this section, we derive infinite series in $1/\sqrt{k}$ for critical points of types  A, I and II. Our methods are
general and apply to critical points with maximal isotropy $\Delta(S_p \times S_q)$, $k > p \gg k/2 \gg q = k-p$. One simplification for the results presented here is that
as $k\arr \infty$, $\ww^i \arr \pm \vv^i$. We also have the estimate
$\|\WW\|= \sqrt{k}(1 + ak^{-1} + O(k^{-\frac{3}{2}}))$ where $a = 0$ (resp.~$2\sqrt{2}$)  for type II (resp.~types A and I). 
This is not obvious but follows easily from our results. 
For families of critical points with $\Delta(S_{k-p} \times S_p)$-isotropy,  where $p  \ll k/2$ is fixed, 
$\ww^i$ will converge, but not necessarily to $\pm \vv^i$, if $i > k-p $.

We illustrate the approach by first discussing type II critical points. Suppose $\WW \in M(k,k)^{\Delta S_{k-1}}$ is of type II. 
Let $\Xi: \real^5 \arr M(k,k)^{\Delta S_{k-1}}$ be the parametrization of $M(k,k)^{\Delta S_{k-1}}$ defined in Section~\ref{ex: deltak-1ex}
and recall that $\Xi^{-1}(\WW)  = (w_{11}, w_{12},w_{1k},w_{k1}, w_{kk})$.
We seek power series for $\Xi^{-1}(\WW)$ of the form
\[
\xi_1 = 1 + \sum_{n=2}^\infty c_n k^{-\frac{n}{2}},\quad
\xi_2 = \sum_{n=2}^\infty e_n k^{-\frac{n}{2}},\quad \xi_5 = -1 + \sum_{n=2}^\infty d_n k^{-\frac{n}{2}} \]
\[
\xi_3 = \sum_{n=2}^\infty f_n k^{-\frac{n}{2}}\quad
\xi_4 = \sum_{n=2}^\infty g_n k^{-\frac{n}{2}}
\]
Numerical investigation of the type II solutions reveals that if the power series expansions exist then
$c_2 = c_3 =e_2 = e_3 =0$.  We assume this here but note that the vanishing of these coefficients can be proved directly. 
Observe also that the constant terms $\pm 1$ (resp.~$0$) for $\xi_1$, $\xi_5$ (resp.~$\xi_2$, $\xi_3$, $\xi_4$) imply that as $k \arr \infty$,  $\ww^i \arr \vv^i$, $i < k$,
and $\ww^k \arr -\vv^k$.

The first non-constant term in each series is an \emph{integer} power of $k^{-1}$. The presence of the powers of $k^{-\frac{1}{2}}$ 
occurs because of the angle terms. In particular (for type II critical points) the angle between $\vv^k$ and $\ww^k$ has series expansion
starting $\pi + e_4k^{-\frac{1}{2}} + \cdots$.  
Again, this can be verified by direct analysis of the equations and is confirmed by numerics.  

For type I critical points, the picture is similar but with some differences. First, the series for $\xi_1$ now starts with $-1$ and $c_2 \ne 0$.
The series for $\xi_2$ also has $e_2 \ne 0$ and $\xi_5$ now has constant term $+1$ (as for type II, $d_2 \ne 0$). As a consequence $\ww^i \arr -\vv^i$, $i < k$,
$\ww^k \arr\vv^k$. Type A is similar, with $\ww^i \arr -\vv^i$ for all
$i \in \is{k}$.

We indicate two related approaches to the derivation of these series and 
illustrate with reference to critical points of type A. Following Section~\ref{sec: dek}, let $\tau = \|\ww^i\|$, $i \in \is{k}$, $\alpha$ (resp.~$\beta$) 
be the angle between $\ww^i$ and $\vv^j$, $i \ne j$ (resp.~$\vv^i$), and
$\Theta$ be the angle between $\ww^i$ and $\ww^j$, $i \ne j$.
In the direct approach, we solve the equation $\grad{\cal{F}}(\WW) = 0$ for the critical point on the fixed point
space $M(k,k)^{\Delta S_k} \approx \real^2$.  In terms of the isomorphism $\Xi:\real^2 \arr M(k,k)^{\Delta S_k}$ (Section~\ref{ex: deltak-1ex}),
and using the results of Section~\ref{sec: cpequations}, we derive the pair of  equations 
\begin{equation}
\begin{split}
\label{eq: series1}
\left((k-1)\big(\sin(\Theta) -\frac{\sin(\alpha)}{\tau}\big) -\frac{\sin(\beta)}{\tau}\right)\xi_{a} =\mbox{     \hspace*{0.2in}        }\\
\hspace*{0.7in}\Theta(\sum_{j\ne i} w_{ja}) - (1-\delta_{1a}) \alpha  - \delta_{1a} \beta + \pi\Omega, \; a \in \is{2},
\end{split}
\end{equation}
where $\xi_a = w_{1a}$,  and $\Omega = 1 - \Xi_{1}(\bxi)^\Sigma= 1 - \xi_1 - (k-1)\xi_2$, for all $j \in \is{k}$.
Next, we compute the initial terms of (formal) power series in $k^{-\frac{1}{2}}$ for $\tau, \alpha,\beta$ and $\Theta$ using the formal series for $\xi_1, \xi_2$. Starting with largest terms 
in~\Refb{eq: series1} (here constant terms), equate coefficients so as to determine $c_2,c_3,e_2,e_3$. 
We find that $c_2 = e_2 = 2$, $c_3 = e_3 = 0$. Set $1/\sqrt{k} = s$, replace $\xi_1$ by $-1 + 2 s^2 + s^4\overline{\xi}_1(s)$, $\xi_2$ by
$2s^2 + s^4 \overline{\xi}_2(s)$, substitute in the equations and cancel 
the factors of $s^2$ to derive maps $F_i(\overline{\xi}_1,\overline{\xi}_1,s)$ defined on a neighbourhood of $(c_4,e_4,0)$ in $\real^2 \times \real$. 
As part of this, the values of $c_4,e_4$ are determined.
The Jacobian of $F = (F_1,F_2)$ is then shown to be non-singular at $(c_4,e_4,0)$ and it follows by the implicit function theorem that
we have analytic functions $\overline{\xi}_i(s)$, $i = 1,2$ defined on a neighbourhood $U$ of $s = 0$ such that $F(\overline{\xi}_1(s),\overline{\xi}_2(s),s) = 0$, $s \in U$.
Since the functions $\overline{\xi}_i$ are analytic, they have convergent power series representations on a neighbourhood $U'$ of $0$.  With some effort, it is
possible to estimate the radius of convergence of the series at $s = 0$~\cite[\S 1.3]{Krantz1992}. 
In practice, the series appears to converge for values of $s$ corresponding to relatively small (perhaps all) values of $k \ge 2$ (type A).
We give the full argument for type A later in the section; the arguments for types I and II are similar and not given in detail.

We sketch an alternative approach, based on the consistency equations, which gives good estimates, simplifies the initial computations, and
provides information on the path based approach described previously.  We illustrate the method for type A critical points. 
Starting with the consistency equation~\Refb{eq: consk}, and taking $\xi_1 = 1 + \rho$, $\xi_1 + (k-1)\xi_2 = 1$, we derive an equation  for $\xi_1$
\begin{equation*}
\left((k-1)\big(\sin(\Theta) -\frac{\sin(\alpha)}{\tau}\big) -\frac{\sin(\beta)}{\tau} + \Theta\right)\frac{1-\xi_1}{k-1} + \beta - \alpha = 0. 
\end{equation*}
Computing the initial coefficients of the series for $\xi_1$, we find that $\xi_1 = -1 + 2k^{-1} + 0 k^{-\frac{3}{2}} + O(k^{-2})$.
Now $\xi_2 = (1 - \xi_1)/(k-1) = 2 k^{-1} + 0 k^{-\frac{3}{2}} + O(k^{-2})$ and $\xi_1,\xi_2$ give the correct first two
non-constant terms for the type A critical point series solution. In practice, determining the initial terms of the series for the critical point is most important step for
finding the infinite series representation.  These terms can always be obtained
by first solving the consistency equations.  A consequence is that both the constant term (for diagonal)
entries and initial non-constant term for the path joining $\bxi_0$ to the associated critical point, are constant along the path.  For types A and II critical points the first
two non-constant terms are constant along the path (we discuss the situation for type I later). 
All of this explains the small derivatives with respect to $\lambda$ of $\bxi(\lambda)$ and why the
solutions obtained by the consistency equations are good approximations to the associated critical point. 
As we shall see, the estimate provided by the 
solution of the consistency equations,
is generally better than that provided by taking the approximation given by the first two non-constant terms in the infinite series for the critical point.

\subsection{Critical points of type  \emph{II}}\label{sec: typeII} 
\begin{thm}\label{thm: initial_terms}
For critical points of type II, we have the convergent series for the components of the critical point 
\[
\xi_1 = 1 + \sum_{n=4}^\infty c_n k^{-\frac{n}{2}},\quad
\xi_2 = \sum_{n=4}^\infty e_n k^{-\frac{n}{2}},\quad \xi_5 = -1 + \sum_{n=2}^\infty d_n k^{-\frac{n}{2}} \]
\[
\xi_3 = \sum_{n=2}^\infty f_n k^{-\frac{n}{2}},\quad
\xi_4 = \sum_{n=2}^\infty g_n k^{-\frac{n}{2}}
\]
where
\[
\begin{matrix}
\vspace*{0.1in}
c_4& =&\frac{8}{\pi}&d_2& =&2 +8 \frac{\pi +1}{\pi^2}& e_4& =&-\frac{4}{\pi} \\
f_2 &= & 2 &g_2 & = & \frac{4}{\pi}& & & \\
&&&&&&&&\\
\vspace*{0.1in}
c_5& =&-\frac{320\pi}{3\pi^4 (\pi-2)} &d_3& = &\frac{64 \pi -768}{3\pi^4(\pi-2)}&e_5& =&-\frac{32}{\pi^3} \\
f_3 & = & 0&g_3 & = &\frac{32}{\pi^3} &&& 
\end{matrix}
\]
\end{thm}
\proof We use the second method to find solutions $c_2, \ldots, e_5$
of the consistency equations and then use these to determine $f_2$, $f_3$, $g_2$, $g_3$ as described above.
The estimates on angles and norms needed for the computations are given in Appendix~\ref{sec: hot8}.
Using the estimates, and following the notation of Section~\ref{sec: con},
we may equate coefficients of $k^{-1}$ in the equations $\varphi_{11}=\varphi_{12}$, $\varphi_{11} = \varphi_{k1}$, $\varphi_{1k} = \varphi_{kk}$ to obtain
\begin{eqnarray*}
0 & = & 2 + c_4 - d_2 + \frac{e_4^2}{2} \\
0 & = & 4+c_4-d_2 +e_4\frac{\pi}{2} + \frac{e_4^2}{2} \\
0 & = & \pi+4 - \frac{\pi d_2}{2} + e_4 +c_4 = 0   
\end{eqnarray*}
From the first two equations, it follows that $e_4 = -\frac{4}{\pi}$,
Solving for $c_4,d_2$, we find $c_4 = \frac{8}{\pi}$ and $d_2 = 2 +\frac{8}{\pi}+ \frac{8}{\pi^2}$.

The coefficients $e_5,c_5,d_3$ are found by equating coefficients of $k^{-\frac{3}{2}}$. 
\begin{eqnarray*}
0 & = & e_4e_5-d_3 + c_5 \\
0 & = & e_4^2+ \frac{\pi e_5}{2} \\
0 & = & c_5+e_5- \frac{2e_4^3}{3} - \frac{d_3\pi}{2}
\end{eqnarray*}
Solving the equations, we find that
\begin{eqnarray*}
c_5&=&-\frac{320\pi}{3\pi^4 (\pi-2)} \approx -3.013\\
d_3&=&\frac{64 \pi -768}{3\pi^4(\pi-2)}\approx -1.699\\
e_5&=&-\frac{32}{\pi^3} \approx -1.032,
\end{eqnarray*}
The coefficients $f_2,f_3$ (resp.~$g_2,g_3$) are found by setting $1/\sqrt{k} = s$ 
and substituting for $\xi_1,\xi_2,\xi_5$ in $\xi_1 + (s^{-2}-2)\xi_2 + \xi_3 - 1 = O(s^4)$ (resp.~$\xi_5 + (s^{-2}-2)\xi_4 -1 = O(s^2)$).

We briefly describe the method for constructing the power series in $1/\sqrt{k}$ for the critical points (see the analysis of type A for more detail).
Set $s =1/\sqrt{k}$ and look for solutions of the form   $\xi_1 = 1+c_4s^4 + c_5 s^5 + s^6 \overline{\xi}_1(s)$,
$\xi_2 = e_4s^4 + e_5 s^5 + s^6 \overline{\xi}_2(s)$, $\xi_3 = f_2s^2 + f_3 s^3 + s^4 \overline{\xi}_3(s)$, $\xi_3 = g_2s^2 + g_3 s^3 + s^4 \overline{\xi}_4(s)$, and
$\xi_1 = -1+d_2s^2 + d_3 s^3 + s^4 \overline{\xi}_5(s)$. After substitution in the equations for the critical points, we
derive an equation $L(\overline{\xi}_1,\cdots,\overline{\xi}_5) = \mathcal{C} + O(s)$, where $L: \real^5 \arr \real^5$ is a linear isomorphism
and $\mathcal{C} \in \real^5$ is a constant.
The result follows by the implicit function theorem---we may also find $c_6,e_6,f_4,g_4$ and $d_4$ (these coefficients are different from the consistency equation solutions).
\qed

\subsubsection{Numerics for type II critical points}
In Table~\ref{table: compII}, we compare the components of the critical point $\mathfrak{c}$ with the approximation $\mathfrak{c}^a$ to the critical point given by taking the first three terms 
in the series given by Theorem~\ref{thm: initial_terms} (the first term will be the constant term, even if that is zero). We also include the approximation 
$\mathfrak{c}^s$ given by the solution of the consistency equations.
Interestingly, the consistency equation approximation $\mathfrak{c}^s$ outperforms the approximation $\mathfrak{c}^a$ given by the first three terms in
the series for the components of the critical point.

\begin{scriptsize}
\begin{table}[h]
\hspace*{-0.5in}
\begin{tabular}{|c||c|c|c|c|c|c|}

\hline
Comp.          & $\xi_1$       & $\xi_2$ &$\xi_3$  & $\xi_4$ & $\xi_5$         \\ \hline \hline
$\isk{c}^a$          &$1+2.51634\times 10^{-8}$         &$-1.2836 \times 10^{-8}$   & $2.00000\times 10^{-4} $  & $1.28356\times 10^{-4}$  & $-1+5.3400\times 10^{-4}$     \\ \hline
$\isk{c}^s$          &$1+2.51456\times 10^{-8}$         &$-1.2835 \times 10^{-8}$   & $1.99966\times 10^{-4} $  & $1.28302 \times 10^{-4}$  & $-1+5.3370\times 10^{-4}$     \\ \hline
$\isk{c}$            &$1+2.51446\times10^{-8}$         &$-1.2834 \times 10^{-8}$   & $1.99954\times 10^{-4} $  & $ 1.28295\times 10^{-4}$  & $-1+5.3365\times 10^{-4}$     \\ \hline
$|c_i^a-c_i|$       &$\approx 2\times 10^{-11}$         &$\approx 1\times 10^{-12}$   & $\approx4.19 \times 10^{-8} $  & $\approx 5 \times 10^{-8}$  & $\approx 3 \times 10^{-7}$     \\ \hline
$|c_i^s-c_i|$       &$\approx 1\times 10^{-12}$         &$\approx 8 \times 10^{-13}$   & $\approx 4.19 \times 10^{-8} $  & $\approx 7 \times 10^{-9}$  & $\approx 5 \times 10^{-8}$     \\ \hline
\end{tabular}
\vspace*{0.05in}
\caption{$k = 10^4$. Numerically computed comparison of type II critical point $\mathfrak{c}$, the approximation $\mathfrak{c}^a$ given by Theorem~\ref{thm: initial_terms} and
$\isk{c}^s$ given by the consistency equations.}
\label{table: compII}
\end{table}
\end{scriptsize}

\subsection{Critical points of type  A} 
\begin{prop}\label{prop: initial_termsA}
For critical points of type A, we have the convergent series for the components of the critical point
\[
\xi_1 = -1 + \sum_{n=2}^\infty c_n k^{-\frac{n}{2}},\quad
\xi_2 = \sum_{n=2}^\infty e_n k^{-\frac{n}{2}}
\]
where
\[
\begin{matrix}
c_2& =& 2&e_2& =&2 \\
c_3 &= & 0 &e_3 & = & 0 \\
c_4&=&\frac{8}{\pi}-4 &e_4&=&\frac{4}{\pi}-2
\end{matrix}
\]
\end{prop}
\proof We follow the direct method. First we need estimates for $\tau = \|\ww^i\|$ and the angles $\alpha,\beta,\Theta$. Substituting the series in the expressions for norms and angles,
we find
\begin{eqnarray*}
\tau^2 & = & 1+ (4-2c_2)k^{-1} - 2c_3k^{-\frac{3}{2}} +O(k^{-2}) \\
\tau & = & 1 + (2-c_2)k^{-1} - c_3k^{-\frac{3}{2}} +O(k^{-2})\\
\tau^{-1} & = & 1  -(2-c_2)k^{-1} + c_3k^{-\frac{3}{2}}  +O(k^{-2}) \\
\langle \ww^i,\ww^j\rangle/\tau^2 & = & (2c_2-4)k^{-2} +O(k^{-\frac{5}{2}}) \\
\langle \ww^i,\vv^j\rangle/\tau & = & e_2k^{-1} + e_3k^{-\frac{3}{2}} +O(k^{-2}),\; i \ne j \\
\langle \ww^i,\vv^i\rangle/\tau & = & -1 + 2k^{-1} + O(k^{-2})
\end{eqnarray*}
It follows straightforwardly that
\begin{enumerate}
\item $\Theta = \frac{\pi}{2}  -(2c_2-4)k^{-2} + O(k^{-\frac{5}{2}}) $.
\item $\sin(\Theta) = 1 + O(k^{-4})$.
\item $\cos(\alpha) =   e_2k^{-1} + e_3k^{-\frac{3}{2}} + O(k^{-2})$. 
\item $\alpha =  \frac{\pi}{2}  - e_2k^{-1}- e_3k^{-\frac{3}{2}} - O(k^{-2})$.
\item $\sin(\alpha) = 1 - \frac{e_2^2}{2}k^{-2} + O(k^{-\frac{5}{2}})$.
\item $\cos(\beta) = -1 +2k^{-1} + O(k^{-\frac{5}{2}})$.
\item $\sin(\beta) = 2k^{-\frac{1}{2}} + O(k^{-\frac{3}{2}})$.
\item $\beta = \pi - 2k^{-\frac{1}{2}} +O(k^{-\frac{3}{2}})$.
\end{enumerate}
Next   substitute in equations~\Refb{eq: series1} with $a = 1,2$ and $w_{1a} = \xi_a$ and compare constant terms.
It follows from the $a = 2$ equation that $e_2 = 2$ (the only constant term is on the right hand side of the equation). 
Taking $e_2 = 2$ and looking at the constant terms in the $a=1$ equation we find that $c_2 = 2$.  Examining 
terms in $k^{-\frac{1}{2}}$ in both equations,
we find that $c_3 = e_3 = 0$ (terms in $k^{-\frac{1}{2}}$ involving $\beta, \sin(\beta)$ cancel). 

It remains to prove that we have convergent power series solutions. Set $s = 1/\sqrt{k}$, and
define new variables $\overline{\xi}_i=\overline{\xi}_i(s)$, $i = 1,2$, where
$\xi_1 = -1 + 2s^2 +  s^4\overline{\xi}_1(s)$, $\xi_2 =2 s^2 + s^4\overline{\xi}_2(s)$ and 
$\overline{\xi}_1(0) = c_4$, $\overline{\xi}_2(0) = e_4$. We redo the previous estimates in terms of $s$ and
$\overline{\xi}_i$.
\begin{eqnarray*}
\tau^2&=&1+ s^4 (4 \overline{\xi}_2-2 \overline{\xi}_1) + s^6(4(\overline{\xi}_1-\overline{\xi}_2)+\overline{\xi}_2^2) + s^8(\overline{\xi}^2_1-\overline{\xi}^2_2)\\
\tau^{-1}&=&1- s^4(2 \overline{\xi}_2-\overline{\xi}_1) + \sum_{n=3}^\infty s^{2n}F_n(\overline{\xi}_1,\overline{\xi}_2)\\
\cos(\Theta)&=& 2\overline{\xi}_2s^4 + \sum_{n=3}^\infty s^{2n}C_n(\overline{\xi}_1,\overline{\xi}_2)\\
\sin(\Theta)& =& 1 - 2\overline{\xi}_2^2 s^8 + \sum_{n=5}^\infty s^{2n}S_n(\overline{\xi}_1,\overline{\xi}_2) \\
\Theta&=&\frac{\pi}{2} - 2\overline{\xi}_2 s^4 + \sum_{n=3}^\infty s^{2n}T_n(\overline{\xi}_1,\overline{\xi}_2)\\
\cos(\alpha) & = & 2s^2 + s^4\overline{\xi}_2  + \sum_{n=3}^\infty s^{2n}U_n(\overline{\xi}_1,\overline{\xi}_2) \\
\sin(\alpha) & = &   1 - 2s^4 + 2s^6\overline{\xi}_2 +\sum_{n=4}^\infty s^{2n}V_n(\overline{\xi}_1,\overline{\xi}_2)\\
\alpha & = &  \frac{\pi}{2} - 2s^2 - s^4\overline{\xi}_2 +  \sum_{n=3}^\infty s^{2n}W_n(\overline{\xi}_1,\overline{\xi}_2) \\ 
\end{eqnarray*}
\begin{eqnarray*}
\cos(\beta) & = & -1 + 2s^2 + s^4(3\overline{\xi}_2-\overline{\xi}_1) +  \sum_{n=3}^\infty s^{2n}X_n(\overline{\xi}_1,\overline{\xi}_2) \\
\sin(\beta) & = & 2s - \frac{s^3}{4}(2-3\overline{\xi}_2+\overline{\xi}_1) +  \sum_{n=2}^\infty s^{2n+1}Y_n(\overline{\xi}_1,\overline{\xi}_2)\\
\beta & = &  \pi - 2 s  - s^3\big(\frac{5}{6} + \frac{3\overline{\xi}_2}{4} -\frac{\overline{\xi}_1}{4}\big)+  \sum_{n=2}^\infty s^{2n+1}Z_n(\overline{\xi}_1,\overline{\xi}_2)
\end{eqnarray*}
where $F_n,\cdots,Z_n$ are real analytic functions in two variables. It is easy to verify that given $R > 0$, there exists $r > 0$ such that
the infinite series defined above are convergent for $|s| < r$ if $\|(\overline{\xi}_1,\overline{\xi}_2)\| \le R$.

Substitute $k = s^{-2}$ in~\Refb{eq: series1} with $a = 1,2$. Taking $a =1$, we have
\begin{equation*}
\begin{split}
\left[(s^{-2} - 1)\big(\sin(\Theta) - \frac{\sin(\alpha)}{\tau}\big) - \frac{\sin(\beta)}{\tau}\right](-1 +2 s^2 + s^4\overline{\xi}_1) =\qquad \\
\Theta\big((s^{-2}-1)(2s^2 + s^4\overline{\xi}_2) - \beta - \pi(-2 + 2s^2+ s^4\overline{\xi}_1 + (s^{-2} - 1)(2s^2+s^4\overline{\xi}_2))
\end{split}
\end{equation*}
Using our expressions for the angle and norm terms we find that the only terms involving $s$ are those involving $\beta,\, \sin(\beta)$ and these cancel. 
On the other hand if we equate the coefficients of $s^2$, we find that
\begin{equation}\label{eq: xi1}
\overline{\xi}_1 + \overline{\xi}_2(\frac{\pi}{2}-2) = 2 - \pi +O(s)
\end{equation}
We similarly seek terms in $s^2$ of the equation for $a = 2$. Here the left hand side makes no contribution and we
find  
\begin{equation}\label{eq: xi2}
\overline{\xi}_2  = \frac{4}{\pi}-2+ O(s)
\end{equation}
The equations (\ref{eq: xi1},\ref{eq: xi2}) are derived from~\Refb{eq: series1} by cancelling terms of order $s$ and constants in ~\Refb{eq: series1}
and then dividing by $s^2$. Taking $s = 0$, we see that $e_4 = \frac{4}{\pi}-2$, $c_4 =2e_4$ and the Jacobian of the
equations defined by dividing~\Refb{eq: series1} by $s^2$ is $1$ at $s = 0$, $(\overline{\xi}_1,  \overline{\xi}_2) = (\frac{4}{\pi}-2, \frac{8}{\pi}-4)$.
Applying the real analytic version of the implicit function theorem gives the required infinite series representation of the solutions. 
\subsubsection{Numerics for type A}
\begin{scriptsize}
\begin{table}[h]
\hspace*{-0.5in}
\begin{tabular}{|c||c|c|c|c|c|c|}

\hline
Comp.          & $\xi_1$       & $\xi_2$          \\ \hline \hline
$\isk{c}^a$          &$-1+2\times 10^{-4}$         &$2 \times 10^{-4}$        \\ \hline
$\isk{c}^{a+}$          &$-1+1.9998546\times 10^{-4}$         &$1.999927 \times 10^{-4}$        \\ \hline
$\isk{c}^s$          &$-1+1.9997999\times 10^{-4}$         &$2 \times 10^{-4}$        \\ \hline
$\isk{c}$            &$-1+1.9998000\times10^{-4}$         &$1.999930 \times 10^{-4}$        \\ \hline
$|c_i^a-c_i|$       &$\approx 2\times 10^{-8}$         &$\approx 7\times 10^{-9}$       \\ \hline
$|c_i^{a+}-c_i|$       &$\approx 6.5\times 10^{-10}$         &$\approx 2.6\times 10^{-10}$       \\ \hline
$|c_i^s-c_i|$       &$\approx 2\times 10^{-10}$         &$\approx 7 \times 10^{-9}$      \\ \hline
\end{tabular}
\vspace*{0.05in}
\caption{$k = 10^4$. Numerically computed comparison of type A critical point $\mathfrak{c}$, the approximations $\mathfrak{c}^a$, $\mathfrak{c}^{a+}$ given by Proposition~\ref{prop: initial_termsA}, and the
solution $\isk{c}^s$ of the consistency equations.}
\label{table: compA}
\end{table}
\end{scriptsize}

We compare the components of the critical point $\mathfrak{c}$ with the approximation $\mathfrak{c}^a$ (resp.~$\mathfrak{c}^{a+}$) 
to the critical point given by taking the first three (resp.~four) terms
in the series given by Proposition~\ref{prop: initial_termsA} (the first term will be the constant term, even if that is zero). We also include the approximation
$\mathfrak{c}^s$ given by the solution of the consistency equations.
The consistency equation approximation $\mathfrak{c}^s$ again outperforms the approximation $\mathfrak{c}^a$ given by the first three terms in
the series for the components of the critical point. However, $\mathfrak{c}^{a+}$ and $\mathfrak{c}^s$ give similar approximations with $\mathfrak{c}^{a+}$ 
outperforming $\mathfrak{c}^s$ on the approximation to $\xi_2$, as might be expected.
\subsection{Critical points of type  I} 
\begin{prop}\label{prop: initial_termsI}
For critical points of type \emph{I}, we have the convergent series for the components of the critical point
\[
\xi_1 = -1 + \sum_{n=2}^\infty c_n k^{-\frac{n}{2}},\quad
\xi_2 = \sum_{n=2}^\infty e_n k^{-\frac{n}{2}},\quad \xi_5 = 1 + \sum_{n=2}^\infty d_n k^{-\frac{n}{2}} \]
\[
\xi_3 = \sum_{n=2}^\infty f_n k^{-\frac{n}{2}},\quad
\xi_4 = \sum_{n=4}^\infty g_n k^{-\frac{n}{2}}
\]
where
\[
\hspace*{-0.5in}\begin{matrix}
c_2 = & 2              &d_2= &\frac{8(\pi - 1)}{\pi^2}& e_2   = & 2                             &f_2 =&0            &g_2 =& 2 - \frac{4}{\pi}  \\
c_3 = & 0              &d_3 =&-4.798751               & e_3   = & 0                             &f_3 =&0            &g_3 = &\frac{32}{\pi^2} \left(\frac{1}{\pi} - 1\right)                    \\
c_4 =&\frac{16}{\pi}-4 &&                             & e_4   = & \frac{8}{\pi}-2               &f_4= &\frac{16}{\pi^2} - \frac{12}{\pi} && \\
c_5 =&4.441691         &&                             & e_5 = & \frac{8(\pi^2+4(\pi-1))}{\pi^3} &f_5 =&6.205827     &&
\end{matrix}
\]
\end{prop}
\proof Brief details. Write $\xi_1(s) = -1 + 2s^2 + s^4\oxi(s)$, $\xi_2(s) = 2s^2 + s^4 \oxi_2(s)$, $\xi_3(s) = s^4\oxi_3(s)$,
 $\xi_4(s) = s^2 \oxi_4(s)$ and $\xi_5(s) = 1 + s^2\oxi_5(s)$, substitute in the equations for the critical points and,
after division by $s^2$, reduce to an equation $L(\oxi_1,\cdots,\oxi_5) = \cal{C} + O(s)$, where $L$ is linear and non-singular and
$\cal{C} \in \real^5$ is constant.
Following the same procedure used for type A critical points, we find the values of $\oxi_i(0)$, $i \in \is{5}$ and apply the
implicit function theorem to complete the proof.  The values of $c_5, d_3, e_5, f_5, g_3$ were computed by equating the  coefficients 
of $s^3$ to zero in the equations for the critical points. \qed

\subsubsection{Numerics for type \emph{I}}
\begin{scriptsize}
\begin{table}[h]
\hspace*{-0.5in}
\begin{tabular}{|c||c|c|c|c|c|c|}

\hline
Comp.                & $\xi_1$                      & $\xi_2$                     & $\xi_3$                     & $\xi_4$                    & $\xi_5$         \\ \hline \hline
$\isk{c}^a$          &$-0.999799988625$      &$2.00005940\times 10^{-4}$       &$-2.136\times 10^{-8}$     & $7.0466\times 10^{-5} $    &    $1.00016879$     \\ \hline
$\isk{c}^s$          &$-0.99979997459$      &$2.0001295047 \times 10^{-4}$      &$-1.689\times 10^{-8}$    & $7.0496\times 10^{-5}$     & $1.00016885$     \\ \hline
$\isk{c}$            &$-0.999799988626$      &$2.00005936 \times 10^{-4}$      &$-2.137\times 10^{-8} $    & $7.0494\times 10^{-5}$     & $1.00016885$     \\ \hline
$|c_i^a-c_i|$        &$\approx 4.5\times 10^{-12}$     &$\approx 4.1\times 10^{-12}$    &$\approx 2.7 \times 10^{-12} $ & $\approx 2.8 \times 10^{-8}$ & $5.8 \times 10^{-8}$     \\ \hline
$|c_i^s-c_i|$        &$\approx 1.4\times 10^{-8}$     &$\approx 7.0 \times 10^{-9}$   &$\approx  4.4 \times 10^{-9} $ & $\approx 2 \times 10^{-9}$ & $\approx 2.3 \times 10^{-9}$     \\ \hline
\end{tabular}
\vspace*{0.05in}
\caption{$k = 10^4$. Numerically computed comparison of type \emph{I} critical point $\mathfrak{c}$, the approximation $\mathfrak{c}^a$ given by Theorem~\ref{prop: initial_termsI} and the
solution $\isk{c}^s$ of the consistency equations.}
\label{table: compI}
\end{table}
\end{scriptsize}

We compare the components of the critical point $\mathfrak{c}$ with the approximation $\mathfrak{c}^a$ to the critical point given by taking the terms
given by Theorem~\ref{thm: initial_terms}.  We also include the approximation
$\mathfrak{c}^s$ given by the solution of the consistency equations. Note that the approximations given by the series for $\xi_1, \xi_2, \xi_3$ 
are far better than those given by the consistency equations; the reverse is the case for $\xi_4,\xi_5$, where fewer terms from the series are used.  
\begin{rem}
If we compare the coefficients given by Proposition~\ref{prop: initial_termsI} with those given by solving the consistency equations, we find that
$c_2,c_3,d_2,e_2,e_3,f_2,f_3,g_2$ are the same, the other coefficients differ. Even though there is only agreement of the first non-constant term for $\xi_4,\xi_5$, 
the approximations given by the consistency equations for these terms are notably better than those given by the series approximation. \rend
\end{rem}

\subsection{Decay of critical values at critical points of type \emph{II}}
Given $k \ge 6$, denote the critical point of type \emph{II} by $\mathfrak{c}_k \in M(k,k)^{\Delta S_{k-1}}$.
Using Theorem~\ref{thm: initial_terms}, we may write $\cal{F}(\mathfrak{c}_k)$ as an infinite series in $1/\sqrt{k}$: $\sum_{n=0}^\infty u_n k^{-\frac{n}{2}}$. 

Our main result gives a precise estimate on the decay of $\cal{F}(\mathfrak{c}_k)$.
\begin{thm}\label{thm: decayII}
(Notation and assumptions as above.)
\begin{eqnarray*}
\cal{F}(\isk{c}_k)&=&(\frac{e_4^2}{8} + \frac{1}{2} +\frac{e_4}{\pi})k^{-1} +O(k^{-\frac{3}{2}})\\
& = & (\frac{1}{2} -\frac{2}{\pi^2})k^{-1} +O(k^{-\frac{3}{2}})
\end{eqnarray*}
\end{thm}
We break the proof of the result into lemmas, several of which depend on the power series representation for $\mathfrak{c}_k$ given in Theorem~\ref{thm: initial_terms}.

Recall that 
\begin{eqnarray*}
\cal{F}(\WW)&=&\frac{1}{2}\sum_{i,j \in \is{k}} f(\ww^i,\ww^j) - \frac{1}{2}\sum_{i,j \in \is{k}} f(\ww^i,\vv^j) + \frac{1}{2}\sum_{i,j \in \is{k}} f(\vv^i,\vv^j)\\
f(\ww,\vv)&=&\frac{\|\ww\|\|\vv\|}{2\pi}\big(\sin(\theta_{\ww,\vv}) + (\pi-\theta_{\ww,\vv})\cos(\theta_{\ww,\vv}).
\end{eqnarray*}
Following our previous conventions, let $\Theta$ (resp.~$\Lambda$) denote the angles between $\ww^i$ and $\ww^j$ (resp.~$\ww^k$), $i,j < k$, and $\alpha_{\sigma \eta}$
denote the angle between $\ww^i$ and $\vv^j$ where we set $\eta = k$ (resp.~$\sigma = k$) if $j =k$ (resp.~$i=k$) and $\eta = j$, (resp.~$\sigma = i$) otherwise.
Define
\begin{eqnarray*}
\Psi_\Theta & = &  \sin(\Theta) + (\pi - \Theta)\cos(\Theta) \\
\Psi_\Lambda & = &  \sin(\Lambda) + (\pi - \Lambda)\cos(\Lambda) \\
\gamma_{\sigma \eta} & = & \sin(\alpha_{\sigma \eta}) + (\pi - \alpha_{\sigma \eta})\cos(\alpha_{\sigma \eta}),
\end{eqnarray*}
where the labelling for $\gamma_{\sigma \eta}$ follows the same convention as the labelling of the angles between $\ww^i$ and $\vv^j$. 
As usual, set $\|\ww^i\| = \tau$, $i < k$ and $\|\ww^k\| = \kappa$. 
Define
\[\begin{matrix}
E_1 & =  \frac{\tau^2}{4} & E_2 & =  \frac{\kappa^2}{4} & F_1 & =  \frac{\tau^2}{2\pi} \Psi_\Theta  & F_2 & =  \frac{\tau\kappa} {2\pi} \Psi_\Lambda\\
&&&&&&& \\
&&G_{i \eta}& =   \frac{\tau}{2\pi}\gamma_{i \eta}, & G_{k \eta}& =   \frac{\kappa}{2\pi}\gamma_{k \eta}&&
\end{matrix}
\]
\begin{lemma}
\begin{eqnarray*}
\frac{1}{2}\sum_{i,j \in \is{k}}f(\ww^i,\ww^j)&=&(k-1)E_1 + E_2 + (k-1)(k-2)F_1 + (k-1)F_2 \\
\sum_{i,j \in \is{k}}f(\ww^i,\vv^j)&=&(k-1)G_{i i} + (k-1)(k-2)G_{i j} + (k-1)G_{ik} + \\
&&\quad G_{kk} + (k-1)G_{kj} \\
\frac{1}{2}\sum_{i,j \in \is{k}}f(\vv^i,\vv^j)&=& \frac{k}{4} + \frac{k^2-k}{4\pi}
\end{eqnarray*}
\end{lemma}
\proof Elementary and omitted, \qed

Using the series representation of Theorem~\ref{thm: initial_terms} we find that
\begin{align*}
\tau^2 &=  1 + T_2 k^{-2} + T_{2.5}k^{-\frac{5}{2}} +T_3 k^{-3}, \quad  & \Psi_\Theta  =  1 + A_2 k^{-2} +A_{2.5}k^{-\frac{5}{2}} + A_3 k^{-3} \\
\tau\kappa & = 1 + K_1 k^{-1}+K_{1.5}k^{-\frac{3}{2}} + K_2 k^{-2}, \quad & \Psi_\Lambda = 1 + F_1 k^{-1}+F_{1.5}k^{-\frac{3}{2}} + F_2 k^{-2}
\end{align*}
where
\begin{enumerate}
\item $T_2 = 2(c_4+2)$, $T_{2.5} = 2c_5$, $T_3 = 2 c_6 + e_4^2 + 4 g_4$.
\item $A_2 = \frac{\pi}{2}(4+2e_4)$, $A_{2.5} =\pi e_5$, $A_3 =  \frac{\pi}{2}(4g_4 +  2 e_6 + e_4^2)$
\item $K_1 = \frac{e_4^2 - 2d_2}{2}$, $K_{1.5} = (e_4e_5-d_3)$, \\ $K_2 = c_4+2 + \frac{e_5^2-e_4^2+d_2e_4^2}{2} -f_4 e_4  -d_4 - \frac{e_4^4}{8} $
\item $F_1 = -\frac{\pi}{2}(2+e_4)$, $F_{1.5} = -\frac{\pi}{2}e_5$, $F_2 =  \frac{(e_4+2)^2}{2} + \frac{\pi}{2}(\frac{e_4^3}{2}-e_4d_2 + f_4 - g_4)$
\end{enumerate}

The next two lemmas are proved using straightforward substitution and computation. 
\begin{lemma}\label{lem: sqrt1}
(Notation and assumptions as above.) The coefficient of $k^{-\frac{1}{2}}$ in
\begin{enumerate}
\item $(k-1)E_1$ is $0$.
\item $E_2$ is $0$.
\item $(k-1)(k-2)F_1$ is $\frac{1}{4\pi}(T_{2.5} + A_{2.5})$.
\item $(k-1)F_2$ is $\frac{1}{2\pi}(K_{1.5}+F_{1.5})$
\end{enumerate}
In particular, the coefficient of $k^{-\frac{1}{2}}$ in $\frac{1}{2}\sum_{i,j\in \is{k}} f(\ww^i,\ww^j)$ is
\[
\frac{1}{2\pi}\left(c_5+e_4e_5-d_3\right).
\]
\end{lemma}
\begin{lemma}\label{lem: ww}
The coefficient of $k^{-1}$ in $\frac{1}{2}\sum_{i,j \in\is{k}}f(\ww^i,\ww^j)$ is
\begin{align*}
&\frac{T_2+2K_1}{4} +\frac{1}{4\pi}(T_3+A_3 - 3(T_2+A_2)+2(K_2 + F_2 + K_1F_1- K_1-F_1)) \\
&= (2 c_6 + 4 g_4 - 4c_4 - 4 + e_5^2 + d_2 e_4^2 - d_4-f_4 e_4 - e_4^4/4 + 4e_4 +2d_2)/4\pi +\\
& \qquad \frac{1}{4}\left(2c_4 - 2e_4 + e_6 + \frac{e_4^2}{2} +f_4+g_4\right) 
\end{align*}
\end{lemma}

Next we determine the coefficients of $k^{-\frac{1}{2}}$ and $k^{-1}$ in $\sum_{i,j \in \is{k}} f(\ww^i,\vv^j)$.
We have
\begin{align*}
\tau &=  1 + t_2 k^{-2} + t_{2.5}k^{-\frac{5}{2}} +t_3 k^{-3},\;\;  &\gamma_{ij} & =  1 + a_2 k^{-2}+ a_{2.5}k^{-\frac{5}{2}} + a_3 k^{-3} \\
\kappa & = 1 + m_1 k^{-1} + m_{1.5} k^{-\frac{3}{2}} +m_2 k^{-2},\;\;  & \gamma_{ik}&= 1 + p_1 k^{-1} + p_{1.5}k^{-\frac{3}{2}} + p_2 k^{-2} \\
\gamma_{kj}&=  1 + q_1 k^{-1} + q_{1.5}k^{-\frac{3}{2}} + q_2 k^{-2},\;\;&\gamma_{ii}& =  \pi + r_1 k^{-1} + r_2 k^{-2},\;\; \gamma_{kk} = O(k^{-\frac{3}{2}}) 
\end{align*}
where
\begin{enumerate}
\item $t_2 = (c_4+2)$, $t_{2.5}=c_5$, $t_3 =  c_6 + \frac{e_4^2}{2} + 2 g_4$.
\item $a_2 = \frac{\pi}{2}e_4$, $a_{2.5}= \frac{\pi}{2}e_5$, $a_3 =  \frac{\pi}{2} e_6$
\item $m_1 = \frac{e_4^2 - 2d_2}{2}$, $m_{1.5}=(e_4e_5-d_3)$, $m_2 = \frac{e_5^2-e_4^2+d_2e_4^2}{2} -f_4 e_4  -d_4 - \frac{e_4^4}{8} $
\item $p_1 = \pi$, $p_{1.5}=0$, $p_2 =  2+\frac{\pi}{2}g_4$
\item $q_1 = -\frac{\pi}{2}e_4$, $q_{1.5}=-\frac{\pi}{2}e_5$, $q_2 = \frac{e_4^2}{2} + \frac{\pi}{2} (f_4+\frac{e_4^3 - 2d_2e_4}{2})$.
\item $r_1 = 0$, $r_2 = -2\pi$.
\end{enumerate}

\begin{lemma}\label{lem: sqrt2}
The coefficient of $k^{-\frac{1}{2}}$ in
\begin{enumerate}
\item $(k-1) G_{ii}$ is $0$.
\item $(k-1)(k-2) G_{ij}$ is $\frac{1}{2\pi}(t_{2.5} + a_{2.5})$.
\item $(k-1)G_{ik}$ is $\frac{1}{2\pi}p_{1.5}$.
\item $G_{kk}$ is $0$.
\item $(k-1)G_{kj}$ is $\frac{1}{2\pi}(q_{1.5} + m_{1.5})$.
\end{enumerate}
In particular, the coefficient of $k^{-\frac{1}{2}}$ in $-\sum_{i,j\in \is{k}} f(\ww^i,\vv^j)$ is
\[
-\frac{1}{2\pi}\left(c_5+e_4e_5-d_3\right).
\]
\end{lemma}
\begin{lemma}\label{lem: zc}
The constant term and coefficient of $k^{-\frac{1}{2}}$ in the series expansion of $\cal{F}(\isk{c}_k)$ are zero.
\end{lemma}
\proof It follows from Lemmas~\ref{lem: sqrt1}, \ref{lem: sqrt2} that the coefficient of $k^{-\frac{1}{2}}$ is zero.
The proof that the constant term is zero is a straightforward computation and omitted. \qed

\begin{lemma}
The coefficient of $k^{-1}$ in
\begin{enumerate}
\item $(k-1) G_{ii}$ is $\frac{c_4}{2}$.
\item $(k-1)(k-2) G_{ij}$ is $\frac{1}{2\pi}(a_3+t_3 - 3(a_2+t_2))$.
\item $(k-1)G_{ik}$ is $\frac{1}{2\pi}(t_2+p_2 - p_1)$.
\item $G_{kk}$ is $0$.
\item $(k-1)G_{kj}$ is $\frac{1}{2\pi}(m_2+q_2 + m_1q_1- m_1 - q_1)$.
\end{enumerate}
\end{lemma}
\begin{lemma}\label{lem: wv}
The coefficient of $k^{-1}$ in $\sum_{i,j \in\is{k}}f(\ww^i,\vv^j)$ is
\begin{align*}
&\frac{\big(a_3+t_3 - 3(a_2+t_2)+ t_2+p_2 -p_1 + m_2+q_2+m_1q_1- m_1 - q_1 +\pi c_4\big)}{2\pi}\\
=& \frac{1}{4}( e_6 - 2e_4 + g_4 - 2 + f_4 + 2c_4) + \\
& \frac{1}{2\pi}(c_6 -2c_4 + 2 g_4 -2 + (e_5^2 + d_2e_4^2)/2-f_4 e_4 - d_4 - e_4^4/8 +   d_2)
\end{align*}
\end{lemma}

The next lemma completes the proof of Theorem~\ref{thm: decayII}.
\begin{lemma}\label{lem: fin}
The coefficient of $k^{-1}$ in $\cal{F}(\isk{c}_k)$ is
\[
\frac{e_4^2}{8} + \frac{1}{2} +\frac{e_4}{\pi} = \frac{1}{2} -\frac{2}{\pi^2}
\]
\end{lemma}
\proof To compute the coefficient of $k^{-1}$ in $\cal{F}(\isk{c}_k)$ it suffices to compute the
coefficient of $k^{-1}$ in $\frac{1}{2}\sum_{i,j\in \is{k}}f(\ww^i,\ww^j) - \sum_{i,j\in \is{k}}f(\ww^i,\vv^j)$.
Substituting the expressions given by Lemma~\ref{lem: ww}, \ref{lem: wv} gives the first expression. The equality follows
using the known value for $e_4$.  \qed
\begin{rems}
(1) It follows from Lemma~\ref{lem: fin} that the decay rate for $\cal{F}(\mathfrak{c}_k^s)$, where $\mathfrak{c}_k^s$ is the approximation to
$\mathfrak{c}_k$ given by the consistency equations is exactly the same as that for  $\cal{F}(\isk{c}_k)$. \\
(2) The decay rate does not depend on the higher order coefficients $e_5,c_5,d_3, f_4,g_4$.  \rend
\end{rems}
\subsection{Decay of critical values at critical points of types \emph{A} and \emph{I}}\label{sec: slodecay}
Given $k \ge 6$, denote the critical point of type \emph{A} by $\mathfrak{c}^A$  and of type \emph{I} by $\mathfrak{c}^I_k$.
Using Theorem~\ref{thm: initial_terms}, we may write $\cal{F}(\mathfrak{c}^{A}_k), \cal{F}(\mathfrak{c}^{I}_k)$ as infinite series in $1/\sqrt{k}$ (no positive powers of $\sqrt{k}$).  

\begin{prop}\label{prop: decayAI}
(Notation and assumptions as above.)
\[ 
\cal{F}(\isk{c}^A_k)=\cal{F}(\isk{c}^I_k)=\frac{1}{2} - \frac{1}{\pi} +O(k^{-\frac{1}{2}})
\]
\end{prop}
\proof The argument for type \emph{A} critical points is similar to that of Theorem~\ref{thm: decayII}, but much simpler.
Noting the critical point series for type \emph{I} are similar to those of type \emph{A}, the result for type \emph{I} may either be deduced from
the result for type \emph{A} or easily proved directly along the same lines as for type \emph{A}.
\qed

\section{Discussion}\label{sec: comm}
\subsection{Contributions to machine learning}
The student-teacher two-layer ReLU model is a central setting in machine learning that allows one to focus on the learning process and study it 
\emph{separately} from the expressiveness of the trained model. Using our analytic estimates and power series representations, we have derived 
estimates for the loss function for different families of local minima for arbitrarily large (but finite) numbers of neurons and inputs. 
Until now, the study of local minima for a large number of neurons and inputs in this model has only been done numerically~\cite{SafranShamir2018}. 
Moreover, although the Student-Teacher model has been studied for more than thirty years, starting in the statistical physics 
community with the thermodynamic limit~\cite{Therm1992} (see also  recent works \cite{Gendynam2019,HMmodel2020}), our work 
appears to be the first that exploits the symmetry of the model explicitly and effectively.

In sharp contrast to current approaches for studying shallow neural networks (thermodynamic limit \cite{Therm1992,Gendynam2019,HMmodel2020}, 
Mean-field \cite{MeanF2018}, optimal control \cite{OptCon2018}, Neural Tangent Kernel \cite{NTK2020} and Compositional Kernels~\cite{CompK2016}---the last 
two approaches apply to more general architectures), our symmetry based approach operates in a natural regime where the number of neurons and inputs is \emph{finite}, 
In particular, our methods do not require any sort of linearization of the optimization problem, but constitute direct assault on this highly nonconvex optimization landscape.

Once a power series representation is obtained for a given family of spurious minima, it is possible to obtain an analytic characterization of the Hessian spectrum 
and this, in turn, allows us to prove \cite{ArjevaniField2020b} that critical points of types A, I, and II define spurious minima for all
large enough $k$ (most probably, $k\ge 7$ for type I and $k \ge 6$ for types A and II).
More importantly, the  analytic characterization of the Hessian spectrum opens the possibility of rigorous analysis of
fundamental questions regarding the mysterious generalization
abilities of neural networks. For example, how does the loss at different families of spurious minima vary when increasing the number of neurons and inputs? Should one 
expect ill- or well-conditioned minima? Are global minima flatter than spurious minima?

\subsection{Contributions to bifurcation theory and dynamics}

The power series representations obtained are valid for real values of $k$ and this allows the study of bifurcation
in the \emph{real} parameter $k$. Viewed in this way, the type II critical point becomes a local minimum when $k \approx 5.58$. 
This raises the intriguing question of defining representations on spaces with non-integer dimension. Of course,
the fixed point space structures we use for families of critical points \emph{are} well-defined for non-integer values of $k$
(as are the Hessian computations given in~\cite{ArjevaniField2020b}).

The creation (or annihilation) of spurious minima in the symmetric setting offers interesting challenges in bifurcation. Most importantly, 
bifurcation does not originate from the global minima, which exist for all $k$. Rather the spurious minima arise from a non-local
bifurcation arising from the collision of multiple families of branches of saddle points.  Here the bifurcation theory of $S_k$, especially on the standard representation of
$S_k$~\cite[\S 16]{FieldRichardson1992}, plays a central role. In brief, the assumption of symmetry allows for the encoding of the highly complex changes in landscape entailed by the
creation of spurious minima. We refer to~\cite{ArjevaniField2021} for more details including minimal models for symmetry breaking on the standard representation of $S_k$.
Of course, this aspect of symmetry is often used in physics and applied mathematics as it can allow one to
\emph{encode} highly complex detail in a relatively tractable problem.  Decoding the information is attained through (controlled) symmetry breaking.

In the articles~\cite{BickField2016,BickField2017a,BickField2017b}, network dynamics is considered in terms of function and implicit optimization of function~\cite[\S 7]{BickField2016}, 
\cite[\S 6]{BickField2017b}. In the context of biological networks, we refer to bifurcation conditional on function and optimization as \emph{evolutionary bifurcation}. 
The student-teacher model considered in this paper provides a good ``toy'' model for evolutionary bifurcation in the sense that it has the property 
that there are many solutions (that is, local minima),
some of which are distinguished by being ``good'' solutions (decay to zero as $k \arr \infty$). For large $k$, the solutions seen by gradient descent will
typically be a spurious minimum with good decay properties. From a biological perspective, an optimal solution (a critical point giving the global minimum)
may well be a bad choice as environmental change may make the solution unviable (over-specialization).  Of course, this is a well-known 
phenomenon in many areas of engineering and science.

We conclude with brief comments and remarks about problems that directly follow from the work described in this paper.
It is likely that for sufficiently large $k$, spurious minima 
occur with isotropy $\Delta S_{k-p} \times \Delta S_p$ ($p \ll k/2$), and that these
critical points also admit power series representations.  
For example, the type M critical points that occur for $p = 2$ (Section~\ref{sec: asym}) which we conjecture define spurious minima for all $k\ge 9$ and 
have objective decaying to zero like $0.6 k^{-1}$ as $k \arr \infty$.

Since differentiable regularity constrains isotropy, it is natural to ask if critical points where the isotropy is not a subgroup of $\Delta S_k$
can be spurious minima? We have no examples.

There is the issue of bifurcation with respect to the parameter $\lambda$.  That is, does a curve $\bxi(\lambda)$ starting from a critical point of $\cal{F}_0$ ever 
undergo bifurcation \emph{within the fixed point space}? If this does not happen, then $\bxi$ can always be analytically continued to a critical point of
$\cal{F}$ provided that $\bxi(\lambda)$ is bounded away from $\partial \Omega_a$.

\section{Acknowledgments}
Part of this work was completed while YA was visiting the Simons Institute in 2019 for the \emph{Foundations of Deep Learning} program. We thank Haggai Maron,
Segol Nimrod, Ohad Shamir, Michal Shavit, and Daniel Soudry for helpful and insightful discussions. Specially thanks also to Christian Bick for helpful comments on earlier versions of the manuscript.

\appendix
\section{Terms of higher order in $\lambda$ along $\WW(\lambda)$}\label{sec: hotI}
In Section~\ref{sec: basic} the constants $\tau_0, \kappa_0, A, A_k$ are defined, all of which depend only on $\mathfrak{t}$.
For the next step, additional terms are needed depending on $\mathfrak{t}$ and $\wbxi$ or $\wbxi_0$. Define
\begin{eqnarray*}
N & = & (1+\rho)\wxi_{1} + (k-2)\vareps \wxi_{2} - \frac{\nu \wxi_{3}}{k-1}\\
N_k & = & -(k-1)(\rho+(k-2)\vareps)\wxi_{4} +  (1+\nu)\wxi_{5}\\
D & = & \vareps\wxi_{1} + (1+\rho+(k-3)\vareps)\wxi_{2}-\frac{\nu\wxi_{3}}{k-1} \\
D_k & = & -(\rho+(k-2)\vareps)(\wxi_{1} + (k-2)\wxi_{2}) +\\
& & (1+\rho + (k-2)\vareps)\wxi_{4}+(1+\nu)\wxi_{3} - \frac{\nu\wxi_{5}}{k-1} 
\end{eqnarray*}
In order to construct $\wbxi$, expressions are needed for norms and angles along $\WW(\lambda)$, up to terms of order $\lambda$.
In every case, expressions are truncations of a power series in $\lambda$ (all functions are real analytic).
\subsubsection{Norms \& inner products along $\WW(\lambda)$}
\begin{enumerate}
\item $\|\ww^i\|=\tau_0 + \frac{\lambda N}{\tau_0},\quad 1/\|\ww^i\|=\frac{1}{\tau_0} - \frac{\lambda N}{\tau_0^3}$, $i < k$.
\item $\|\ww^k\|= \kappa_0 + \frac{\lambda N_k}{\kappa_0},\quad 1/\|\ww^k\|= \frac{1}{\kappa_0} - \frac{\lambda N_k}{\kappa_0^3}$.
\item $\ip{\ww^i}{\ww^j} = A + 2\lambda D$, $i, j < k$, $i \ne j$. \\$\ip{\ww^i}{\ww^k} =A_k + \lambda D_k$, $i < k$.
\item
$\ip{\ww^i}{\vv^j}=\vareps + \lambda\wxi_{2}$, $i, j < k$, $i \ne j$\\
$\ip{\ww^i}{\vv^k}=-\frac{\nu}{k-1} + \lambda\wxi_{3}$, $i < k$ \\
$\ip{\ww^k}{\vv^j}=-[\rho + (k-2)\vareps] + \lambda\wxi_{4}$, $j < k$\\
$\ip{\ww^i}{\vv^i}= 1+\rho+\lambda \wxi_{1}$, $i < k$\\
$\ip{\ww^k}{\vv^k}= 1+\nu +  \lambda \wxi_{5}$.
\end{enumerate}

\subsection{Angles along $\WW(\lambda)$}\label{sec: hotII}
Repeated use is made of the $O(\lambda^2)$ approximation $-\frac{\lambda y}{\sin(x)}$ to $\cos^{-1}(x + \lambda y)- \cos^{-1}(x)$.  
\subsubsection{Terms involving $\Theta(\lambda)$,  $\Lambda(\lambda)$.}
Ignoring terms which are $O(\lambda^2)$, we have
\begin{enumerate}
\item $\Theta(\lambda)= \Theta_0 - \frac{2\lambda}{\tau_0^2\sin(\Theta_0)}D + 
 \frac{2A\lambda}{\tau_0^4\sin(\Theta_0)}N$, $i,j < k$, $i\ne j$.
\item $\Lambda(\lambda)= 
\Lambda^0 + \frac{A_k\lambda}{\tau_0\kappa_0^3\sin(\Lambda_0)}N_k+ 
\frac{A_k\lambda}{\tau_0^3\kappa_0\sin(\Lambda_0)}N - 
\frac{\lambda}{\tau_0\kappa_0\sin(\Lambda_0)}D_k$,  $i < k$.
\end{enumerate}
If we define the $\wbxi$-independent terms $R_\ell, S_\ell$, $\ell\in\is{5}$, by
\begin{align*}
\sum_{\ell = 1}^5 R_\ell\wxi_\ell & = -\frac{2}{\tau_0^2\sin(\Theta_0)}D + \frac{2A}{\tau^4\sin(\Theta_0)}N\\
\sum_{\ell = 1}^5 S_\ell\wxi_\ell & = \frac{A_k}{\tau_0\kappa_0^3\sin(\Lambda_0)}N_k+ 
\frac{A_k}{\tau_0^3\kappa_0\sin(\Lambda_0)}N - \frac{1}{\tau_0\kappa_0\sin(\Lambda_0)}D_k,
\end{align*}
then
\begin{align*}
\Theta(\lambda)&= \Theta_0 + \lambda\left(\sum_{\ell = 1}^5 R_\ell\wxi_{\ell}\right), \; i,j < k,\; i\ne j  \\
\Lambda(\lambda)&=  \Lambda_0 + \lambda\left(\sum_{\ell = 1}^5 S_\ell\wxi_{\ell}\right), \; i < k.
\end{align*}
where $R_4 = R_5 = 0$ and
\begin{align*}
R_1 & =  \frac{2}{\tau_0^2\sin(\Theta_0)}\left(\frac{(1+\rho)A}{\tau_0^2} - \vareps\right)\\
R_2 & =  \frac{2}{\tau_0^2\sin(\Theta_0)}\left(\frac{(k-2)\vareps A}{\tau_0^2} - (1+\rho + (k-3)\vareps)\right)\\
R_3 & =  \frac{2}{\tau_0^2\sin(\Theta_0)}\left(\frac{\nu}{k-1}\big(1 - \frac{A}{\tau_0^2}\big)\right)\\ 
S_1 & =  \frac{1}{\tau_0\kappa_0\sin(\Lambda_0)}\left(\frac{A_k(1+\rho)}{\tau_0^2} + (\rho+(k-2)\vareps)\right) \\
S_2 & =  \frac{1}{\tau_0\kappa_0\sin(\Lambda_0)}\left(\frac{A_k(k-2)\vareps}{\tau_0^2} + (k-2)(\rho+(k-2)\vareps)\right) \\
S_3 & =  -\frac{1}{\tau_0\kappa_0\sin(\Lambda_0)}\left(\frac{A_k\nu}{(k-1)\tau_0^2} + (1+\nu)\right) \\
S_4 & =  -\frac{1}{\tau_0\kappa_0\sin(\Lambda_0)}\left(\frac{A_k(k-1)(\rho+(k-2)\vareps)}{\kappa_0^2} + (1+\rho+(k-2)\vareps)\right) \\
S_5 & =  \frac{1}{\tau_0\kappa_0\sin(\Lambda_0)}\left(\frac{A_k(1+\nu)}{\kappa_0^2} + \frac{\nu}{(k-1)}\right) 
\end{align*}
Also needed are expressions for $\sin(\Theta(\lambda))$ and $\beta^{\pm 1}\sin(\Lambda(\lambda))$, where $\beta = \big(\frac{\kappa(\lambda)}{\tau(\lambda)}\big)$.
For this, it suffices to consider
$\sin(\Theta_0 +\lambda (\sum_{\ell = 1}^5 R_\ell\wxi_\ell))$ and $\beta(\lambda)^{\pm 1}\sin(\Lambda_0 +\lambda (\sum_{\ell = 1}^5 S_\ell\wxi_\ell))$.
For $\ell \in \is{5}$, define $J_\ell\in \real$ by
$J_\ell = \frac{A}{\tau_0^2} R_\ell$.
Ignoring $O(\lambda^2)$ terms, we find that
{\small
\begin{align*}
\sin(\Theta(\lambda)) & =  \sin(\Theta_0) + \lambda\left(\sum_{\ell = 1}^5 J_\ell\wxi_{\ell}\right)\\
\sin(\Lambda(\lambda))\frac{\tau(\lambda)}{\kappa(\lambda)} & =  \frac{\sin(\Lambda_0)\tau_0}{\kappa_0} + \lambda\left(\sum_{\ell = 1}^5 K^{kj}_\ell\wxi_{\ell}\right)\\
\sin(\Lambda(\lambda))\frac{\kappa(\lambda)}{\tau(\lambda)} & = \frac{\sin(\Lambda_0)\kappa_0}{\tau_0} + \lambda\left(\sum_{\ell = 1}^5 K^{ik}_\ell\wxi_{\ell}\right),
\end{align*}
where
\begin{align*}
K^{kj}_1 & = \frac{A_kS_1}{\kappa_0^2} + \frac{(1+\rho)\sin(\Lambda_0)}{\tau_0\kappa_0}, \quad
K^{kj}_2  = \frac{A_kS_2}{\kappa_0^2} + \frac{(k-2)\vareps\sin(\Lambda_0)}{\tau_0\kappa_0}, \\
K^{kj}_3 & = \frac{A_kS_3}{\kappa_0^2} - \frac{\nu\sin(\Lambda_0)}{(k-1)\tau_0\kappa_0}, \quad 
K^{kj}_4  = \frac{A_kS_4}{\kappa_0^2} + \frac{(k-1)(\rho+(k-2)\vareps)\tau_0\sin(\Lambda_0)}{\kappa_0^3}, \\
K^{kj}_5 & = \frac{A_kS_5}{\kappa_0^2} - \frac{(1+\nu)\tau_0\sin(\Lambda_0)}{\kappa_0^3}, \quad
K^{ik}_1  = \frac{A_kS_1}{\tau_0^2} - \frac{(1+\rho)\kappa_0\sin(\Lambda_0)}{\tau_0^3}, \\
K^{ik}_2 & = \frac{A_kS_2}{\tau_0^2} - \frac{(k-2)\vareps\kappa_0\sin(\Lambda_0)}{\tau_0^3}, \quad
K^{ik}_3  =  \frac{A_kS_3}{\tau_0^2} + \frac{\nu\kappa_0\sin(\Lambda_0)}{(k-1)\tau_0^3}.\\ 
K^{ik}_4 & = \frac{A_kS_4}{\tau_0^2}-\frac{(k-1)(\rho+(k-2)\vareps)\sin(\Lambda_0)}{\tau_0\kappa_0}, \quad
K^{ik}_5  =  \frac{A_kS_5}{\tau_0^2} +  \frac{(1+\nu)\sin(\Lambda_0)}{\tau_0\kappa_0}.
\end{align*}
} \normalsize

\subsubsection{Terms involving $\alpha(\lambda)$.}
Ignoring $O(\lambda^2)$ terms we have
\begin{align*}
\alpha_{ij}(\lambda)&=\alpha_{ij}^0 - \frac{\lambda}{\tau_0\sin(\alpha_{ij}^0)}\left(\wxi_{2} - 
 \frac{\vareps N}{\tau_0^2}\right),\;i,j < k,\; i\ne j \\
\alpha_{ik}(\lambda) &= \alpha_{ik}^0 - \frac{\lambda}{\tau_0\sin(\alpha_{ik}^0)}\left(\wxi_{3} +  
\frac{\nu N}{(k-1)\tau_0^2}\right),\; i < k\\
\alpha_{ii}(\lambda) &= \alpha_{ii}^0 - \frac{\lambda}{\tau_0\sin(\alpha_{ii}^0)}\left(\wxi_{1} -  
\frac{(1+\rho)N}{\tau_0^2}\right) ,\; i < k\\
\alpha_{kj}(\lambda) &= \alpha_{kj}^0 - \frac{\lambda}{\kappa_0\sin(\alpha_{kj}^0)} \left(\wxi_{4} +\frac{(\rho + (k-2)\vareps)N_k} {\kappa_0^2} \right),\;j < k\\
\alpha_{kk}(\lambda) &= \alpha_{kk}^0 - \frac{\lambda}{\kappa_0\sin(\alpha_{kk}^0)}\left(\wxi_{5} -  \frac{(1+\nu)N_k }{\kappa_0^2}\right)
\end{align*}
Finally, we need expressions for the quotient of $\sin(\alpha)$ by $\tau$ or $\kappa$.
For $\sigma \in \{i,k\}$ $\eta \in \{i,j,k\}$ and $(\sigma,\eta) \not= (j,j)$, we have
\begin{eqnarray*}
\alpha_{\sigma\eta}(\lambda)&=&\alpha_{\sigma\eta}^0 +\lambda\left( \sum_{\ell=1}^5 E^{\sigma\eta}_\ell \wxi_{\ell}\right) \\
\frac{\sin(\alpha_{\sigma\eta}(\lambda))}{\|\ww^\sigma\|}&=&\frac{\sin(\alpha_{\sigma\eta}^0)}{\|\ww^{\mathfrak{t},\sigma}\|} +\lambda\left( \sum_{\ell=1}^5 F^{\sigma\eta}_\ell \wxi_{\ell}\right),
\end{eqnarray*}
where
{\small
\begin{align*}
E^{ij}_1 & =  \frac{\vareps(1+\rho)}{\tau_0^3 \sin(\alpha_{ij}^0)},\; F^{ij}_1 = \frac{\vareps}{\tau_0} E^{ij}_1-\frac{(1+\rho)\sin(\alpha^0_{ij})}{\tau_0^3}\\
E^{ij}_2 & =  \frac{1}{\tau_0\sin(\alpha_{ij}^0)}\left[\frac{(k-2)\vareps^2}{\tau_0^2} - 1\right], \; F^{ij}_2 = \frac{\vareps}{\tau_0} E^{ij}_2 -\frac{(k-2)\vareps \sin(\alpha^0_{ij})}{\tau_0^3}\\
E^{ij}_3 & =  -\frac{\vareps \nu}{(k-1)\tau_0^3 \sin(\alpha_{ij}^0)},\; F^{ij}_3 = \frac{\vareps}{\tau_0} E^{ij}_3 + \frac{\nu\sin(\alpha^0_{ij})}{(k-1)\tau_0^3}\\
E^{ij}_4 & =  F^{ij}_4 =  E^{ij}_5 = F^{ij}_5= 0\\
E^{ik}_1 & =  -\frac{\nu(1+\rho)}{\tau_0^3 (k-1)\sin(\alpha_{ik}^0)},\; F^{ik}_1 = -\frac{\nu}{(k-1)\tau_0} E^{ik}_1 -\frac{(1+\rho)\sin(\alpha^0_{ik})}{\tau_0^3}\\
E^{ik}_2 & =  -\frac{(k-2)\vareps\nu}{\tau_0^3(k-1)\sin(\alpha_{ik}^0)}, \; F^{ik}_2 = -\frac{\nu}{(k-1)\tau_0} E^{ik}_2 -\frac{(k-2)\vareps \sin(\alpha^0_{ik})}{\tau_0^3}\\
E^{ik}_3 & =  \frac{1}{\tau_0 \sin(\alpha_{ij}^0)}\left[\frac{\nu^2}{(k-1)^2\tau_0^2}-1\right],\; F^{ik}_3 = -\frac{\nu}{(k-1)\tau_0} E^{ik}_3 + \frac{\nu\sin(\alpha^0_{ik})}{(k-1) \tau_0^3}\\
E^{ik}_4 & =  F^{ik}_4 =  E^{ik}_5 = F^{ik}_5= 0 \\
E^{ii}_1 & =  \frac{1}{\tau_0\sin(\alpha_{ii}^0)}\left[ \frac{(1+\rho)^2}{\tau_0^2} -1\right],\; F^{ii}_1 = \frac{(1+\rho)}{\tau_0} E^{ii}_1 - \frac{(1+\rho)\sin(\alpha^0_{ii})}{\tau_0^3}\\
E^{ii}_2 & =  \frac{(1+\rho)(k-2)\vareps}{\tau_0^3\sin(\alpha_{ii}^0)}, \; F^{ii}_2 = \frac{(1+\rho)}{\tau_0} E^{ii}_2 -\frac{(k-2)\vareps \sin(\alpha^0_{ii})}{\tau_0^3}\\
E^{ii}_3 & =  -\frac{(1+\rho)\nu}{(k-1)\tau_0^3 \sin(\alpha_{ii}^0)},\; F^{ii}_3 = \frac{(1+\rho)}{\tau_0} E^{ii}_3 + \frac{\nu\sin(\alpha^0_{ii})}{(k-1)\tau_0^3}\\
E^{ii}_4 & =  F^{ii}_4 =  E^{ii}_5 = F^{ii}_5= 0 
\end{align*}
\begin{align*}
E^{kj}_4 & =  \frac{1}{\kappa_0 \sin(\alpha_{kj}^0)}\left[\frac{(k-1)(\rho+(k-2)\vareps)^2}{\kappa_0^2}-1\right],\; F^{kj}_4 = -\frac{(\rho+(k-2)\vareps)}{\kappa_0} E^{kj}_4 + \\
& \hspace*{0.4in}\frac{(k-1)(\rho+ (k-2)\vareps)\sin(\alpha_{kj}^0)}{\kappa_0^3} \\
E^{kj}_5 & =  -\frac{(1+\nu)(\rho+(k-2)\vareps)}{\kappa_0^3\sin(\alpha_{kj}^0)},\;F^{kj}_5 = -\frac{(\rho+(k-2)\vareps)}{\kappa_0}E^{kj}_5 - \frac{(1+\nu)\sin(\alpha_{kj}^0)}{\kappa_0^3}\\
E^{kj}_i&= F^{kj}_i = 0,\; i \notin \{4,5\}. \\
E^{kk}_4 & =  -\frac{(k-1)(1+\nu)(\rho+(k-2)\vareps)}{\kappa_0^3 \sin(\alpha_{kk}^0)},\\
F^{kk}_4 & = \frac{(1+\nu)}{\kappa_0} E^{kk}_4+\frac{(k-1)(\rho+ (k-2)\vareps)\sin(\alpha_{kk}^0)}{\kappa_0^3}\\
E^{kk}_5 & =  \frac{1}{\kappa_0\sin(\alpha_{kk}^0)}\left[\frac{(1+\nu)^2}{\kappa_0^2}-1\right],\;F^{kk}_5 = \frac{(1+\nu)}{\kappa_0}E^{kk}_5- \frac{(1+\nu)\sin(\alpha_{kk}^0)}{\kappa_0^3}\\
E^{kk}_i&= F^{kk}_i = 0,\; i \notin \{4,5\}. 
\end{align*}
}\normalsize
\begin{rem}\label{rem: errors}
A comment on the accuracy of the consistency equations and the formulas listed above.
One check is given by the continuation of the curve $\bxi(\lambda)$ to $\lambda = 1$. This gives the critical points of $\cal{F}$ and is
consistent with the results in Safran \& Shamir~\cite{SafranShamir2018} (see Section~\ref{sec: numII}). 
A more sensitive and subtle test is given by
looking for solutions with $\Delta S_k$ symmetry. Here the angles $\alpha_{ij}$, $\alpha_{kj}, \alpha_{ik}$ should be equal, as should
$\alpha_{ii}, \alpha_{kk}$, and $\Theta_{ij}$, $\Theta_{ik}$. Any computations not respecting the symmetry indicate an error.
At this time, based on careful numerical checks, we believe the formulas given above are correct. \rend
\end{rem}
\subsection{Formulas for $\widehat{\is{h}}^k(\wbxi)$ and $\widehat{\is{h}}^1(\wbxi)$}
We have
{\small
\begin{eqnarray*}
\widehat{\is{h}}^k(\wbxi) & = & \left(\frac{(k-1)[\tau_0\sin(\Lambda_0) - \sin(\alpha_{kj}^0)]-\sin(\alpha_{kk}^0)}{\kappa_0} \right) \Xi^{k}(\wbxi) - \\
&& \Lambda^0 \sum_{j=1}^{k-1}\Xi^{j}(\wbxi) -\left(\sum_{\ell=1}^5 S_\ell \wxi_\ell\right)\big( \sum_{j=1}^{k-1}\ww^{\mathfrak{t},j} \big) + \\
&&\left((k-1)\left[\sum_{\ell=1}^5 K^{kj}_\ell\wxi_\ell - \sum_{\ell = 1}^5 F_\ell^{kj} \wxi_\ell\right] - \sum_{\ell = 1}^5 F_\ell^{kk} \wxi_\ell\right)\ww^{\mathfrak{t},k} + \\
&&  \left(\sum_{\ell = 1}^5 E_\ell^{kj} \wxi_\ell\right)\big(\sum_{j=1}^{k-1}\vv^j\big) +\left(\sum_{\ell = 1}^5 E_\ell^{kk} \wxi_\ell\right)\vv^k + 
\pi\Xi(\wbxip)^\Sigma 
\end{eqnarray*}
\begin{eqnarray*}
\widehat{\is{h}}^1(\wbxi) & = & \left((k-2)\sin(\Theta_0) + \frac{\kappa_0}{\tau_0}\sin(\Lambda_0)\right) \Xi^{1}(\wbxi) - \Theta_0 \sum_{j=2}^{k-1} \Xi^{j}(\wbxi) - \\ 
&& \Lambda_0\,\Xi^{k}(\wbxi) - \left(\frac{(k-2) \sin(\alpha^0_{ij}) + \sin(\alpha^0_{ik}) + \sin(\alpha^0_{ii})}{\tau_0}\right) \Xi^{1}(\wbxi)+ \\
&& \left((k-2)\sum_{\ell=1}^5 J_\ell \wxi_\ell + \sum_{\ell=1}^5 K_\ell^{ik}\wxi_\ell\right)\ww^{\mathfrak{t},1} -\\
&& \left((k-2)\sum_{\ell=1}^5 F^{ij}_\ell \wxi_\ell + \sum_{\ell=1}^5 F^{ik}_\ell\wxi_\ell +\sum_{\ell=1}^5  F^{ii}_\ell\wxi_\ell \right)\ww^{\mathfrak{t},1} -\\
&& \big(\sum_{\ell=1}^5 R_\ell \wxi_\ell\big)\big(\sum_{j=2}^{k-1} \ww^{\mathfrak{t},j}\big) -\big(\sum_{\ell=1}^5 S_\ell \wxi_\ell\big)\ww^{\mathfrak{t},k} + \left(\sum_{\ell = 1}^5 E_\ell^{ij} \wxi_\ell\right)\big(\sum_{j=2}^{k-1}\vv^j\big) +\\
&& \left(\sum_{\ell = 1}^5 E_\ell^{ik} \wxi_\ell\right)\vv^k  + \left(\sum_{\ell = 1}^5 E_\ell^{ii} \wxi_\ell\right)\vv^1 + \pi \Xi(\wbxip)^\Sigma 
\end{eqnarray*}
} \normalsize

\section{Computations \& Estimates, Type II}\label{sec: hot8}
If $\cos^{-1}(x) = \pi/2 -\beta$, then
$\beta = \sin^{-1}(x)$. It follows from the power series for $\sin^{-1}(x)$ (Example~\ref{ex: cpsexpl}), or directly, that
$\sin^{-1}(x) = x + x^3/3! + O(x^5)$.
Since $\cos^{-1}(1-x) = 2\sin^{-1}(\sqrt{x/2})$,
\[
\cos^{-1}(1-x)=\sqrt{2x} + \frac{x^{\frac{3}{2}}}{6\sqrt{2}} + O(x^{\frac{5}{2}}).
\]
In what follows, frequent use is made of the estimates
\[
(1+x)^{\frac{1}{2}} = 1 + \frac{x}{2} - \frac{x^2}{8} + O(x^3),\;\;(1+x)^{-\frac{1}{2}} = 1 - \frac{x}{2} + \frac{3x^2}{8} + O(x^3).
\]
\subsubsection{Computing the initial terms}
To simplify notation, set $\ww^{\mathfrak{t},i} = \ww^i$, $i < k$, and  $\ww^{\mathfrak{t},k} = \ww^k$.
We need to take account of the truncations
\[ 
\rho^{(5)}  =  c_{4} k^{-2} + c_5 k^{-\frac{5}{2}},\quad \nu^{(3)}  =  -2 + d_2 k^{-1} + d_3 k^{-\frac{3}{2}}, \quad \vareps^{(5)}  =  e_4 k^{-2} +  e_5 k^{-\frac{5}{2}}.
\]
Throughout what follows, the order of the remainder is only indicated when that is important for computations.
\begin{enumerate}
\item $(k-2)\vareps = e_4 k^{-1} + e_5 k^{-\frac{3}{2}}$.
\item $\rho+(k-2)\vareps = e_4k^{-1} +e_5k^{-\frac{3}{2}}$.
\item $(\rho+(k-2)\vareps)^2= e_4^2k^{-2} + 2e_4e_5 k^{-\frac{5}{2}}$.
\item $-\frac{\nu}{k-1} = 2k^{-1} + (2-d_2)k^{-2} - d_3k^{-\frac{5}{2}}$.
\end{enumerate}
\subsubsection{Norm estimates on $\tau = \|\ww^i\|$, $i < k$.}
\begin{enumerate}
\item
$\tau = 1 + (c_4+2)k^{-2} + c_5 k^{-\frac{5}{2}}$, \\
\item
$\tau^{-1} = 1 - (c_4 + 2)k^{-2} - c_5 k^{-\frac{5}{2}}$.
\end{enumerate}
\subsubsection{Norm estimates on $\tau_k= \|\ww^k\|$.}
\begin{eqnarray*}
\tau_k&= &1 + \frac{e_4^2 -2d_2}{2}k^{-1} +(e_4e_5-d_3)k^{-\frac{3}{2}},\\
\tau_k^{-1}&=& 1 - \frac{e_4^2 -2d_2}{2}k^{-1} -(e_4e_5-d_3)k^{-\frac{3}{2}}, \\
\tau_k^{-1}\tau & = & 1 - \frac{e_4^2-2d_2}{2}k^{-1} -(e_4e_5-d_3)k^{-\frac{3}{2}}, \\ 
\tau_k\tau^{-1} & = & 1 + \frac{e_4^2-2d_2}{2}k^{-1} +(e_4e_5-d_3)k^{-\frac{3}{2}},\\ 
(\tau_k\tau)^{-1} &=& 1 -  \frac{e_4^2-2d_2}{2}k^{-1} -(e_4e_5-d_3)k^{-\frac{3}{2}}. 
\end{eqnarray*}
\subsubsection{Estimates on angles and inner products}
\begin{enumerate}
\item $\langle \ww^i,\ww^j\rangle /\tau^2 =(2e_4 + 4)k^{-2}+ 2e_5 k^{-\frac{5}{2}}$.
\item $\Theta_{ij}^0 = \frac{\pi}{2} - (2e_4 + 4)k^{-2} -2e_5 k^{-\frac{5}{2}}$.
\item $\sin(\Theta_{ij}^0) = 1 + O(k^{-4})$
\item $\langle \ww^i,\ww^k\rangle /(\tau\tau_k) =-(e_4+2)k^{-1}-e_5k^{-\frac{3}{2}}$.
\item $\Theta_{ik}^0 = \frac{\pi}{2} +(e_4+2)k^{-1}+e_5k^{-\frac{3}{2}}$.
\item $\sin(\Theta_{ik}^0) = 1- \frac{(e_4 + 2)^2}{2k^2}- (e_4 + 2)e_5 k^{-\frac{5}{2}}$.
\item $\langle \ww^i,\vv^i\rangle /\tau =1 - \frac{2}{k^2}$.
\item $\alpha_{ii}^0 = 2k^{-1}+(\frac{e_4^2}{4} + 2-d_2)k^{-2}$.
\item $\sin(\alpha_{ii}^0) = 2k^{-1} + (\frac{e_4^2}{4} + 2-d_2)k^{-2}$.
\item $\langle \ww^i,\vv^j\rangle /\tau =\frac{e_4}{k^2} + e_5 k^{-\frac{5}{2}}$.
\item $\alpha_{ij}^0 = \frac{\pi}{2} - e_4k^{-2} - e_5k^{-\frac{5}{2}}$.
\item $\sin(\alpha_{ij}^0) = 1 + O(k^{-4})$.
\item $\langle \ww^i,\vv^k\rangle /\tau =2k^{-1} + (2-d_2)k^{-2}$.
\item $\alpha_{ik}^0 = \frac{\pi}{2} - \frac{2}{k} - (2-d_2)k^{-2}$.
\item $\sin(\alpha_{ik}^0) = 1- 2k^{-2}$.
\item $\langle \ww^k,\vv^k\rangle /\tau_k =-1 + \frac{e_4^2}{2k} + e_4e_5k^{-\frac{3}{2}}$.
\item $\alpha_{kk}^0 = \pi +\frac{e_4}{\sqrt{k}} +e_5k^{-1}$.
\item $\sin(\alpha_{kk}^0) = -\frac{e_4}{\sqrt{k}} -e_5 k^{-1}$.
\item $\langle \ww^k,\vv^j\rangle /\tau_k =-e_4k^{-1} -e_5k^{-\frac{3}{2}}$.
\item $\alpha_{kj}^0 = \frac{\pi}{2} + \frac{e_4}{k}+e_5 k^{-\frac{3}{2}}$.
\item $\sin(\alpha_{kj}^0) = 1-\frac{e_4^2}{2k^2}$.
\end{enumerate}
\subsubsection{Estimates on key terms in the consistency equations, Section~\ref{sec: con}}
\begin{enumerate}
\item $P =(c_4+\frac{e_4^2-2d_2}{2})k^{-1} + (c_5 +e_4e_5-d_3)k^{-\frac{3}{2}}$.
\item $Q =  e_4k^{-\frac{1}{2}}$.
\item $\alpha_{ij} - \alpha_{ii}=\frac{\pi}{2} - 2 k^{-1}$. 
\item $\alpha_{kj} - \alpha_{ii}= \frac{\pi}{2} + (e_4-2)k^{-1}+e_5 k^{-\frac{3}{2}}$.
\item $\alpha_{kk} - \alpha_{ik}=\frac{\pi}{2} +\frac{e_4}{\sqrt{k}} +(2+e_5)k^{-1}$.
\end{enumerate}

\end{document}